\documentclass[a4paper, 11pt]{article}
\pdfoutput=1  
\usepackage{etoolbox,verbatim}
\usepackage{wrapfig}
\usepackage{subfig}
\usepackage{soul}
\usepackage{tabularx}
\newtoggle{arxiv}
\toggletrue{arxiv} 

\newtoggle{customthms}
\toggletrue{customthms} 

\newtoggle{colt}
\togglefalse{colt}

\usepackage{xcolor}
\usepackage{tcolorbox}
\tcbuselibrary{skins,breakable}

\tcbset{
  aibox/.style={
      width=\linewidth,
      top=6pt,
      bottom=0pt,
      left=1.5pt,
      right=1.5pt,
      colframe=black,
      colbacktitle=black,
      enhanced,
      center,
      attach boxed title to top left={yshift=-0.1in,xshift=0.15in},
      boxed title style={boxrule=0pt,colframe=white,},
    }
}
\newtcolorbox{AIbox}[2][]{aibox,title=#2,#1}

\definecolor{primalcolor}{HTML}{960000}
\definecolor{contrarycolor}{HTML}{009696}
\definecolor{primaldarkcolor}{HTML}{660000}
\definecolor{contrarydarkcolor}{HTML}{006666}
\definecolor{irongray}{HTML}{6D6E71}
\definecolor{steelgray}{HTML}{F1F4F7}
\newcommand{\linkcolor}{contrarydarkcolor}
\newcommand{\urlcolor}{contrarydarkcolor}
\newcommand{\citecolor}{contrarydarkcolor}

\newcommand{\thmcolordark}{primaldarkcolor}

\usepackage[T1]{fontenc}

\usepackage{natbib}
\usepackage{amssymb,upgreek,bm}
\usepackage{mathtools}
\usepackage[english]{babel}
\usepackage{multirow,tcolorbox,tikz-cd,turnstile,xspace,xparse}
\usepackage{relsize,breakcites,appendix}
\usepackage{longtable,tablefootnote,booktabs,float,makecell}
\usepackage[normalem]{ulem}
\usepackage{tabu}
\usepackage[shortlabels]{enumitem}
\usepackage{footnote}
\usepackage{boxedminipage}
\usepackage{multirow,nicefrac}
\usepackage{verbatim} %
\usepackage{nicematrix}
\usepackage{subcaption}

\usepackage[colorlinks=true,linkcolor=\linkcolor,urlcolor=\urlcolor,citecolor=\citecolor,breaklinks]{hyperref}

\usepackage{prettyref}

\usepackage[capitalise,nameinlink]{cleveref}

\iftoggle{colt}{
    \DeclareRobustCommand{\qed}{
        \usepackage{thmtools}
        \ifmmode \mathqed
        \else
            \leavevmode\unskip\penalty9999 \hbox{}\nobreak\hfill
            \quad\hbox{\qedsymbol}%
        \fi
    }
}
{
    \usepackage{amsthm}

}
\usepackage{thm-restate}

\iftoggle{arxiv}
{
    \usepackage{mathrsfs}
            \usepackage[
  letterpaper,
  left=1in,
  right=1in,
  top=0.9in,
  bottom=0.9in,
  headheight=14pt,
  headsep=14pt,
  footskip=20pt
]{geometry}

\usepackage{fancyhdr}
\usepackage{lipsum} %

\pagestyle{fancy}
\fancyhf{} 
\fancyfoot[C]{\small \thepage}

    \usepackage[bitstream-charter]{mathdesign}
    
    \usepackage[scaled=0.92]{PTSans}

    \usepackage[noend]{algpseudocode}
    \usepackage{algorithm}

}
{
    \usepackage{mathrsfs}
}

\DeclareMathAlphabet{\mathbfsf}{\encodingdefault}{\sfdefault}{bx}{n}

\numberwithin{equation}{section}

\iftoggle{arxiv}
{
    }
{
    
}
\iftoggle{arxiv}
{
    
}
{
    
}

\usepackage{thmtools}

\Crefname{equation}{Eq.}{Eqs.}
\Crefname{assumption}{Assumption}{Assumptions}
\Crefname{condition}{Condition}{Conditions}
\Crefname{claim}{Claim}{Claims}
\Crefname{property}{Property}{Properties}
\Crefname{construction}{Construction}{Constructions}

\declaretheoremstyle[
    headformat=\normalfont\textcolor{\thmcolordark}{\bfseries\NAME\,\NUMBER}\NOTE,%
    notefont={\normalfont\textcolor{\thmcolordark}{\bfseries}}, 
    notebraces={}{},
    bodyfont=\normalfont\itshape,
    spaceabove = 6pt,
    spacebelow = 6pt,
    ]{coloredthmversion}

\declaretheoremstyle[
    headformat=\normalfont\textcolor{\thmcolordark}{\bfseries\NAME\,\NUMBER}\NOTE,%
    bodyfont=\normalfont\itshape,
    spaceabove = 6pt,
    spacebelow = 6pt,
    ]{coloredthm}

\declaretheoremstyle[
    headformat=\normalfont\textcolor{\thmcolordark}{\bfseries\NAME\,\NUMBER}\NOTE,%
    bodyfont=\normalfont,
    spaceabove = 6pt,
    spacebelow = 6pt,
    ]{coloreddef}

\iftoggle{customthms}
{
    \theoremstyle{coloredthmversion}
}
{}

\iftoggle{customthms}{
  \theoremstyle{coloredthm}
  \newtheorem{theorem}{Theorem}
  \newtheorem{lemma}{Lemma}[section]
  \newtheorem{corollary}{Corollary}[section]
  \newtheorem{proposition}[lemma]{Proposition}
}
{}

\newtheorem*{thminformal*}{Informal Theorem}

\iftoggle{customthms}{
    \theoremstyle{coloreddef}
    \newtheorem{definition}{Definition}[section]

    \newtheorem{property}{Property}[section]
}
{
  
}

\newtheorem{assumption}{Assumption}[section]
\newtheorem{condition}{Condition}[section]

\makeatletter
\newcommand{\neutralize}[1]{\expandafter\let\csname c@#1\endcsname\count@}
\makeatother

\iftoggle{customthms}{
    \newtheoremstyle{named}{}{}{\itshape}{}{\bfseries}{}{.5em}{\Cref{#3} {\normalfont (informal)} }{}
    \theoremstyle{named}
    
    \theoremstyle{plain}
}
{}

\newtheorem*{theorem*}{Theorem}
\newtheorem*{lemma*}{Lemma}
\newtheorem*{corollary*}{Corollary}
\newtheorem*{proposition*}{Proposition}
\newtheorem*{claim*}{Claim}
\newtheorem*{fact*}{Fact}
\newtheorem*{observation*}{Observation}
\newtheorem*{definition*}{Definition}
\newtheorem*{remark*}{Remark}
\newtheorem*{example*}{Example}

\def\ddefloop#1{\ifx\ddefloop#1\else\ddef{#1}\expandafter\ddefloop\fi}
\def\ddef#1{\expandafter\def\csname bb#1\endcsname{\ensuremath{\mathbb{#1}}}}
\ddefloop ABCDEFGHIJKLMNOPQRSTUVWXYZ\ddefloop

\def\ddefloop#1{\ifx\ddefloop#1\else\ddef{#1}\expandafter\ddefloop\fi}
\def\ddef#1{\expandafter\def\csname frak#1\endcsname{\ensuremath{\mathfrak{#1}}}}
\ddefloop ABCDEFGHIJKLMNOPQRSTUVWXYZ\ddefloop

\def\ddefloop#1{\ifx\ddefloop#1\else\ddef{#1}\expandafter\ddefloop\fi}
\def\ddef#1{\expandafter\def\csname fr#1\endcsname{\ensuremath{\mathfrak{#1}}}}
\ddefloop ABCDEFGHIJKLMNOPQRSTUVWXYZ\ddefloop

\def\ddefloop#1{\ifx\ddefloop#1\else\ddef{#1}\expandafter\ddefloop\fi}
\def\ddef#1{\expandafter\def\csname eul#1\endcsname{\ensuremath{\EuScript{#1}}}}
\ddefloop ABCDEFGHIJKLMNOPQRSTUVWXYZ\ddefloop

\def\ddefloop#1{\ifx\ddefloop#1\else\ddef{#1}\expandafter\ddefloop\fi}
\def\ddef#1{\expandafter\def\csname scr#1\endcsname{\ensuremath{\mathscr{#1}}}}
\ddefloop ABCDEFGHIJKLMNOPQRSTUVWXYZ\ddefloop

\def\ddefloop#1{\ifx\ddefloop#1\else\ddef{#1}\expandafter\ddefloop\fi}
\def\ddef#1{\expandafter\def\csname b#1\endcsname{\ensuremath{\mathbf{#1}}}}
\ddefloop ABCDEFGHIJKLMNOPQRSTUVWXYZ\ddefloop

\def\ddefloop#1{\ifx\ddefloop#1\else\ddef{#1}\expandafter\ddefloop\fi}
\def\ddef#1{\expandafter\def\csname bhat#1\endcsname{\ensuremath{\hat{\mathbf{#1}}}}}
\ddefloop ABCDEFGHIJKLMNOPQRSTUVWXYZ\ddefloop

\def\ddefloop#1{\ifx\ddefloop#1\else\ddef{#1}\expandafter\ddefloop\fi}
\def\ddef#1{\expandafter\def\csname btil#1\endcsname{\ensuremath{\tilde{\mathbf{#1}}}}}
\ddefloop ABCDEFGHIJKLMNOPQRSTUVWXYZ\ddefloop

\def\ddefloop#1{\ifx\ddefloop#1\else\ddef{#1}\expandafter\ddefloop\fi}
\def\ddef#1{\expandafter\def\csname bst#1\endcsname{\ensuremath{\mathbf{#1}^\star}}}
\ddefloop ABCDEFGHIJKLMNOPQRSTUVWXYZ\ddefloop

\def\ddefloop#1{\ifx\ddefloop#1\else\ddef{#1}\expandafter\ddefloop\fi}
\def\ddef#1{\expandafter\def\csname bst#1\endcsname{\ensuremath{\mathbf{#1}^\star}}}
\ddefloop abcdeghijklmnopqrstuvwxyz\ddefloop

\def\ddefloop#1{\ifx\ddefloop#1\else\ddef{#1}\expandafter\ddefloop\fi}
\def\ddef#1{\expandafter\def\csname bhat#1\endcsname{\ensuremath{\hat{\mathbf{#1}}}}}
\ddefloop abcdefghijklmnopqrstuvwxyz\ddefloop

\def\ddefloop#1{\ifx\ddefloop#1\else\ddef{#1}\expandafter\ddefloop\fi}
\def\ddef#1{\expandafter\def\csname b#1\endcsname{\ensuremath{\mathbf{#1}}}}
\ddefloop abcdeghijklnopqrstuvwxyz\ddefloop

\def\ddefloop#1{\ifx\ddefloop#1\else\ddef{#1}\expandafter\ddefloop\fi}
\def\ddef#1{\expandafter\def\csname barb#1\endcsname{\ensuremath{\bar{\mathbf{#1}}}}}
\ddefloop abcdefghijklmnopqrstuvwxyz\ddefloop

\def\ddef#1{\expandafter\def\csname c#1\endcsname{\ensuremath{\mathcal{#1}}}}
\ddefloop ABCDEFGHIJKLMNOPQRSTUVWXYZ\ddefloop
\def\ddef#1{\expandafter\def\csname h#1\endcsname{\ensuremath{\widehat{#1}}}}
\ddefloop ABCDEFGHIJKLMNOPQRSTUVWXYZ\ddefloop
\def\ddef#1{\expandafter\def\csname hc#1\endcsname{\ensuremath{\widehat{\mathcal{#1}}}}}
\ddefloop ABCDEFGHIJKLMNOPQRSTUVWXYZ\ddefloop
\def\ddef#1{\expandafter\def\csname t#1\endcsname{\ensuremath{\widetilde{#1}}}}
\ddefloop ABCDEFGHIJKLMNOPQRSTUVWXYZ\ddefloop
\def\ddef#1{\expandafter\def\csname tc#1\endcsname{\ensuremath{\widetilde{\mathcal{#1}}}}}
\ddefloop ABCDEFGHIJKLMNOPQRSTUVWXYZ\ddefloop

\DeclareMathOperator*{\argmin}{arg\,min}

\newcommand{\ballkr}[1][r]{\cB_{k}(r)}

\DeclareMathSymbol{\shortminus}{\mathbin}{AMSa}{"39}

\makeatletter
\newcommand\addtometadatalist[5][]{%
    \begingroup
    \if\relax#3\relax\def\sep{}\else\def\sep{#5}\fi
    \let\protect\@unexpandable@protect
    \xdef#3{\expandafter{#3}\sep #4[#1]{#2}}%
    \endgroup
}

\newcommand\metadatalist{}
\newcommand\metadataformat[2][]{{\small \textbf{#1:} #2}}
\newcommand\metadata[2][]{\addtometadatalist[#1]{#2}{\metadatalist}{\metadataformat}{\\}}
\makeatother

\newcommand{\paperwebsite}[1]{\metadata[Website]{\url{#1}}}
\newcommand{\papercode}[1]{\metadata[Code]{\url{#1}}}
\newcommand{\paperdocs}[1]{\metadata[Documentation]{\url{#1}}}
\newcommand{\paperblog}[1]{\metadata[Blog]{\url{#1}}}
\renewcommand\date[1]{\metadata[Date]{#1}}
\newcommand\correspondence[1]{\metadata[Correspondence]{#1}}

\usepackage{preamble/color-edits}

\addauthor{ms}{magenta}
\addauthor{cp}{blue}

\newcommand{\ignore}[1]{}

\iftoggle{arxiv}
{
\input{preamble/arxiv_title}
}
{}

\hypersetup{
    colorlinks=true,
    urlcolor=orange(sae/ece),
}

\usepackage{xspace}
\usepackage{bbm}
\usepackage{graphicx}
\usepackage{subfig} 
\usepackage{annotate-equations}
\usepackage{pifont}
\usepackage{colortbl}
\definecolor{ckgreen}{RGB}{30,130,60}
\definecolor{ckred}{RGB}{190,40,40}
\definecolor{ckorange}{RGB}{215,120,20}
\definecolor{oursbg}{RGB}{255,241,224}
\newcommand{\cmark}{\textcolor{ckgreen}{\ding{51}}}
\newcommand{\xmark}{\textcolor{ckred}{\ding{55}}}
\newcommand{\pmark}{\textcolor{ckorange}{$\boldsymbol{\sim}$}}

\definecolor{basepolicy}{HTML}{7A6E5D}
\definecolor{visualc}{HTML}{4F93D2}
\definecolor{tactilec}{HTML}{5EAA5E}
\definecolor{oursc}{HTML}{E98A4E}

\definecolor{dark_red}{RGB}{122, 0, 0}
\definecolor{coral}{RGB}{255, 119, 94}
\definecolor{pink_orange}{RGB}{255, 72, 126}
\definecolor{vibrant_pink}{RGB}{255, 0, 104}
\definecolor{pink_pink}{RGB}{255, 37, 153}
\definecolor{wine}{RGB}{204, 0, 102}

\definecolor{light_orange}{RGB}{255, 198, 107}
\definecolor{orange(sae/ece)}{rgb}{1.0, 0.49, 0.0}
\definecolor{dark_orange}{RGB}{216,92,0}

\definecolor{org-purp-0}{RGB}{165, 76, 0}
\definecolor{org-purp-1}{RGB}{250, 130, 28}
\definecolor{org-purp-2}{RGB}{226, 89, 68}
\definecolor{org-purp-3}{RGB}{206, 92, 124}
\definecolor{org-purp-4}{RGB}{116, 80, 146}
\definecolor{org-purp-5}{RGB}{110, 78, 157}

\definecolor{teal(sae/ece)}{rgb}{0, 0.47, 0.52}
\definecolor{aqua}{RGB}{52,172,139}
\definecolor{dark_aqua}{RGB}{35,115,93}
\definecolor{dark_green}{RGB}{0, 92, 34}
\definecolor{solid_green}{RGB}{39, 163, 50}

\definecolor{grape}{RGB}{112,48,160}
\definecolor{purple}{rgb}{0.74, 0.65, 1.0}
\definecolor{dark_purple}{rgb}{0.58, 0.0, 0.82}
\definecolor{periwinkle}{RGB}{191, 140, 230}

\definecolor{light_gray}{rgb}{0.9, 0.9, 0.9}
\definecolor{medium_gray}{rgb}{0.6, 0.6, 0.6} 
\definecolor{dark_gray}{rgb}{0.2, 0.2, 0.2} 

\definecolor{sky_blue}{RGB}{37, 166, 213}
\definecolor{light_blue}{rgb}{0.33, 0.80, 1}
\definecolor{dark_blue}{rgb}{0.098, 0.239, 0.52}
\definecolor{ocean}{RGB}{13, 121, 202}
\definecolor{light_ocean}{RGB}{18, 178, 235}
\definecolor{dark_ocean}{RGB}{10, 89, 148}
\definecolor{vibrant_blue}{RGB}{14, 120, 255}

\definecolor{dark_brown}{rgb}{0.3255, 0.004, 0.001}

\definecolor{PiColor}{HTML}{B4B2A9}
\definecolor{MemERColor}{HTML}{AFCBE8}
\definecolor{MemZeroColor}{HTML}{5B93D1}
\definecolor{RoboMMEColor}{HTML}{7CB8A7}
\definecolor{EventVLAColor}{HTML}{2E7D32}
\definecolor{HAMLETColor}{HTML}{C6BEEC}
\definecolor{MemoryVLAColor}{HTML}{7F77DD}
\definecolor{OursColor}{HTML}{EB6834}

\newcommand{\memer}{\textcolor{MemERColor}{\textbf{MemER}}\xspace}
\newcommand{\memzero}{\textcolor{MemZeroColor}{\textbf{Mem-0}}\xspace}
\newcommand{\robomme}{\textcolor{RoboMMEColor}{\textbf{Uniform Sampling}}\xspace}
\newcommand{\eventvla}{\textcolor{EventVLAColor}{\textbf{EventVLA}}\xspace}
\newcommand{\hamlet}{\textcolor{HAMLETColor}{\textbf{HAMLET}}\xspace}
\newcommand{\memoryvla}{\textcolor{MemoryVLAColor}{\textbf{MemoryVLA}}\xspace}

\newcommand{\ourscolored}{\textcolor{orange(sae/ece)}{\textbf{IMPRINT}}\xspace}

\newcommand{\para}[1]{\textbf{#1. }}

\newcounter{qnum}
\newcounter{tnum}
\newcommand{\ours}{\textcolor{black}{\textsc{Imprint}}\xspace}

\colorlet{actc}{solid_green}
\colorlet{memc}{ocean}

\newcommand{\hist}{\mathrm{H}}
\newcommand{\dataset}{\mathbb{D}}            %
\newcommand{\loss}{\mathrm{L}}

\newcommand{\mem}{m}                %
\newcommand{\memtok}{B}                      %
\newcommand{\memenc}{\mathrm{E}^{\mathrm{mem}}_{\phi}}  %

\newcommand{\tokenizer}{\mathrm{T}_\psi}
\newcommand{\token}{e}

\newcommand{\progress}{\rho}

\newcommand{\obs}{o}

\newcommand{\img}{I}
\newcommand{\proprio}{q}

\newcommand{\imgSpace}{\mathcal{I}}
\newcommand{\proprioSpace}{\mathcal{Q}}

\newcommand{\dec}{\mathcal{D}_\varphi}

\newcommand{\task}{\ell}

\newcommand{\state}{s}

\newcommand{\latent}{z}

\newcommand{\policy}{\pi_{\theta}}
\newcommand{\action}{a}
\newcommand{\acttraj}{\mathbf{\action}}
\newcommand{\actchunk}{a_{t:t+h}}

\newcommand{\actSpace}{\mathcal{A}}

\usepackage{amsmath}

\title{Memory Without Curation: Learning Multi-Stream Latent Memory for Robot Policies}

\author{
\small Yilin Wu$^{1,2,\dagger}$ \quad Rutav Shah$^{1,3}$ \quad Shuaiyi Huang$^{1}$ \quad Sumedh A. Sontakke$^{1}$ \quad Homanga Bharadhwaj$^{1,4}$\\
\small Andrea Bajcsy$^{2}$ \quad Max Simchowitz$^{1,2}$ \quad Guanya Shi$^{1,2}$ \\
\vspace{-.3em}
\rule{.38\textwidth}{.7pt}
\\
\footnotesize
$\dagger$ work done during the internship at Amazon FAR\\
\footnotesize
$^{1}$Amazon FAR\quad
$^{2}$Carnegie Mellon University  \quad
$^{3}$The University of Texas at Austin\quad
$^{4}$Johns Hopkins University\quad
}

\date{\today}
\correspondence{\href{mailto:yilinwu@andrew.cmu.edu}{yilinwu@andrew.cmu.edu}}
\paperwebsite{https://yilin-wu98.github.io/imprint_website/}

\begin{document}

\begin{tcolorbox}[
    colback=steelgray, colframe=irongray,
    boxrule=0pt,
    arc=2mm%
  ]
  \maketitle
   \vspace{-3em}
  \tcbline
  \begin{minipage}[t]{1.0\linewidth}
    \centering
    \includegraphics[width=\linewidth]{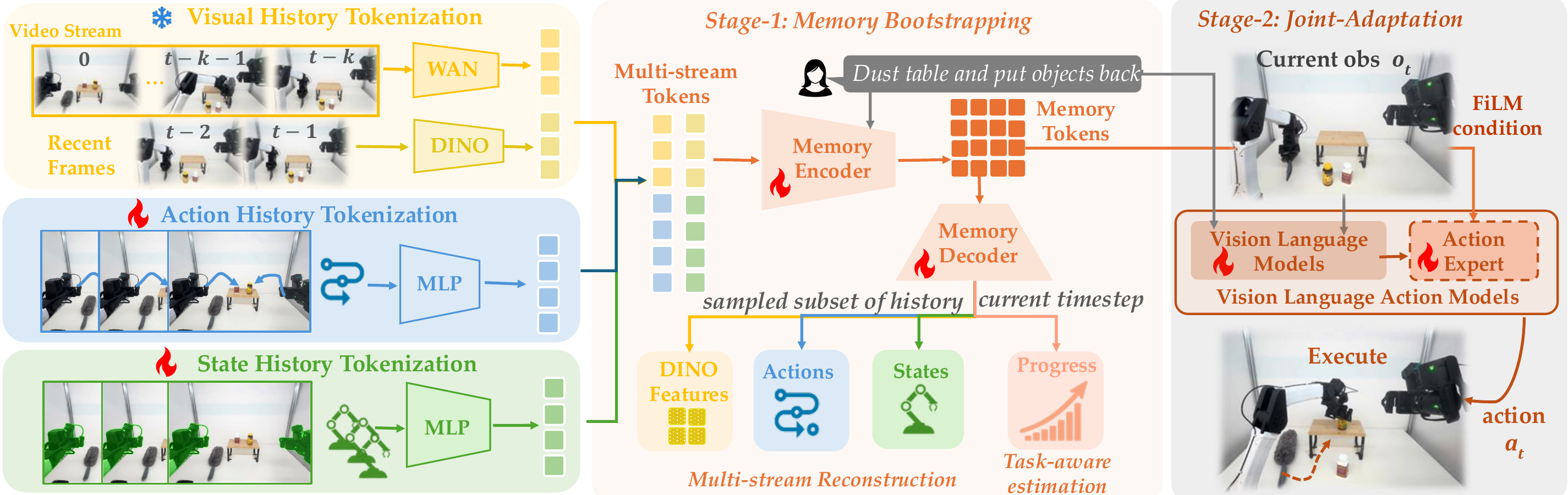}
    \captionof{figure}{%
        \footnotesize\emph{\textbf{Overview of \ours}}. Our latent memory pipeline learns to compress visual, action, and proprioceptive histories through multi-stream reconstruction and task-progress estimation during memory bootstrapping. It then jointly adapts with VLA for action generation.
    }
    \label{fig:method_overview}
\end{minipage}

  \tcbline
  \makeatletter
  \ifdefempty{\metadatalist}{}{\metadatalist\par}
  \makeatother
\end{tcolorbox}

\begin{abstract} It is widely appreciated that robotic policies require long-term memory to perform complex tasks. 
However, directly conditioning policies on all past observations leaves policy learning  susceptible to learning spurious shortcuts, and introduces large latency at policy deployment. As such, today's memory-augmented policies \emph{curate} what to remember, either by retaining only recent frames, selected keyframes, or textual summaries, or by learning latent representations of the visual stream alone. 
In this work, we propose an alternative approach \ours, a policy-agnostic latent memory framework that augments vision-language-action (VLA) models with multi-stream memory without explicitly constraining what the policy should remember. %
\ours compresses different streams into a latent memory that is updated incrementally at inference, enabling low latency. To avoid shortcut learning, \ours adopts a two-stage training procedure, first supervising memory to reconstruct multiple streams (observations, actions, proprioception) of past information, as well as estimate current task progress, followed by end-to-end finetuning, jointly with the policy, on action prediction. Across both simulation and real-world tasks demanding memory, \ours improves average success over the base VLA by at least \textbf{26\%}, outperforming popular alternatives, and maintaining \textbf{lowest inference latency} among memory-augmented VLAs. Through careful ablations, we demonstrate the utility of incorporating diverse information streams into robotic memory, validating the advantages of foregoing history curation. Project website: \url{https://yilin-wu98.github.io/imprint_website/}.

\end{abstract}
\thispagestyle{plain}

\section{Introduction}

\begin{wrapfigure}{r}{0.5\linewidth}
    \centering
    \vspace{-0.6cm}
    \includegraphics[width=\linewidth]{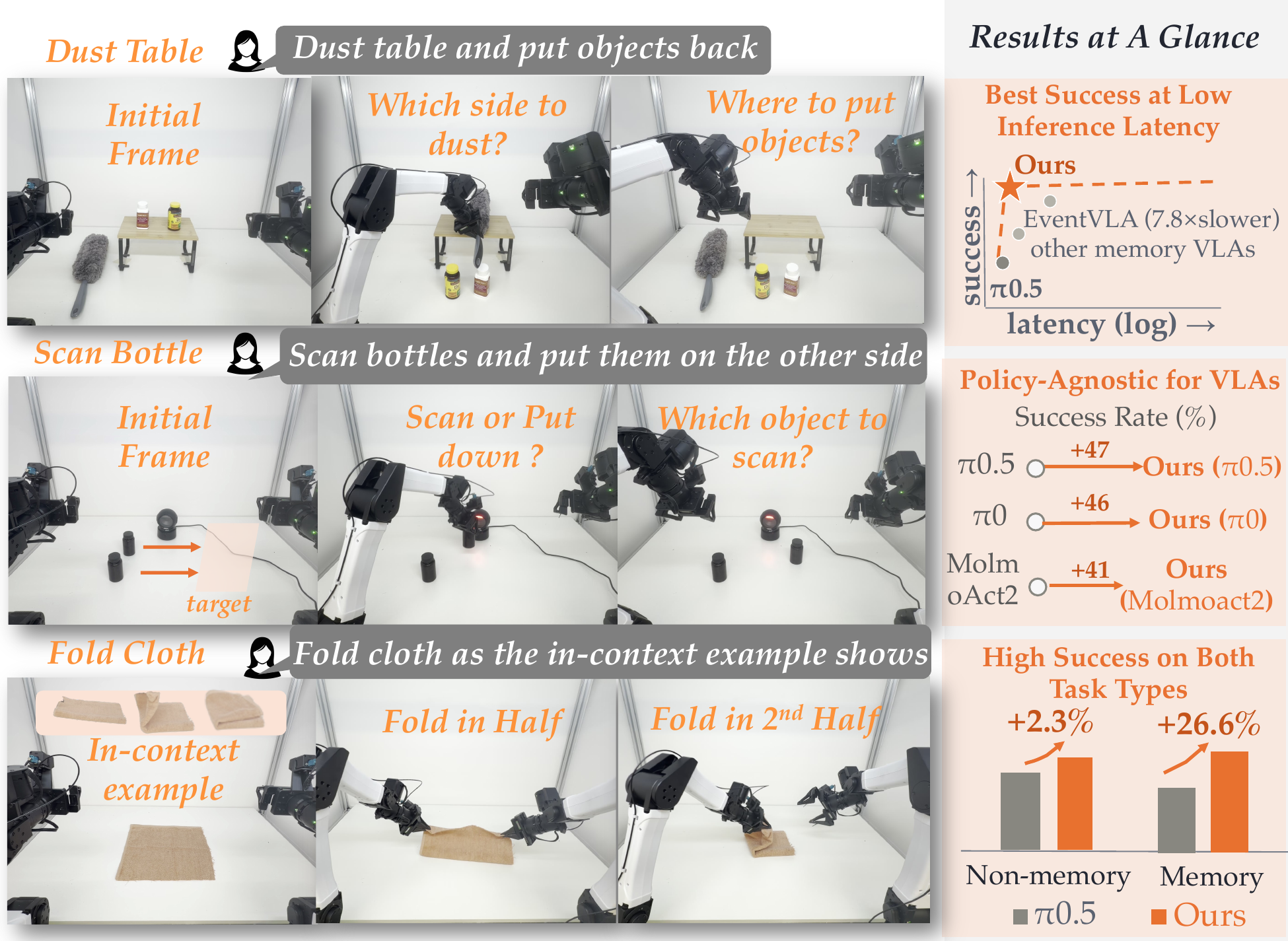}
    \caption{\textbf{Real World Tasks \& Result Highlight}}
    \label{fig:teaser_figure}
    \vspace{-0.4cm}
\end{wrapfigure}

Many manipulation tasks require information unavailable from current observations. For example, a robot dusting a table must move objects aside, sweep the surface, and put them back. It relies on visual, action, and proprioceptive histories to remember the objects' original locations and the direction of previous sweeping motion (Fig.~\ref{fig:teaser_figure}). Because many initial vision-language-action (VLA) models~\citep{black2024pi_0,bjorck2025gr00t,intelligence2025pi} act only on current observations, recent work has started augmenting them with memory to retain task history.

To keep inference efficient, most memory-augmented policies address the challenges of history retention by \emph{curating} what the policy should remember, either in explicit modalities like recent frames, selected keyframes, and textual summaries ~\citep{dai2026robomme,mark2026bpp,yang2026eventvla,zeng2026kemo,sridhar2026scaling,torne2026mem}, or by training latent representation on carefully selected objectives \citep{dashora2026workspacemodelslightweightrobotic}. However, these choices impose a predefined notion of relevance: sparse keyframes can omit intermediate motion, textual summaries abstract away sensorimotor details, and bounded windows exclude distant history. 
In cloth folding (Fig.~\ref{fig:teaser_figure}), visual history may capture the change of cloth shape while action and proprioceptive histories provide more information about precise grasping locations and folding trajectories, helping the robot reproduce the fold accurately. Thus, diverse memory requirements across tasks make it difficult to specify in advance exactly \emph{what needs to be remembered}, limiting the performance of those memory-curated methods.

In this paper, we propose \ours, an alternative framework that augments robotic foundation models \emph{\textbf{with a rich multi-stream memory that learns what the policy can remember without explicitly constraining it}}, while enjoying strong performance and low inference cost. 
To achieve this, \ours{} compresses history of multiple \emph{streams}, including visual observations, actions, and proprioceptive states, into a shared latent space of memory, which conditions action prediction for a policy.
Building such latent memory raises two core challenges. First, \emph{encoding}: how to compress heterogeneous and growing streams efficiently online? Second, \emph{learning}: how to retain useful information from each stream to predict actions?
Past approaches training memory representations from scratch jointly with the policy via action loss alone \citep{shi2026memoryvla} have struggled in Sec.~\ref{sec:exp_learning}, because such coupling can encourage learning spurious shortcuts in demonstrations~\cite{de2019causal}. Two-stage methods instead supervise memory representations before adapting to control but typically focus on a single visual stream~\citep{dashora2026workspacemodelslightweightrobotic,koo2026hamlet}.

In \ours, we overcome these challenges through key innovations in both our design of memory encoding and learning.  For encoding, our insight is that memory should reduce visual streams to a scale comparable to other streams, then fuse them efficiently over time. To this end, we use a pretrained video tokenizer~\citep{wan2025wan} to remove spatiotemporal redundancy from visual streams, bringing them close to the scale of other streams. From there, all modalities supervise a recurrent bottleneck~\citep{yang2025gated} which integrates all streams into the latent memory, updated incrementally so that efficiency is maintained as the episode grows.

For learning, our insight is that memory should be supervised directly, before adaptation to control, to retain both fine-grained multi-stream history and coarse task progress. Rather than focusing on perfect reconstruction, our goal is to ``prime'' the bottleneck so that end-to-end policy finetuning avoids learning spurious memory shortcuts. Supervising on past visual features, actions, and proprioceptive states encourages memory to preserve complementary scene and motion information across all three streams, while progress estimation helps distinguish task stages that look similar but require different actions.
Since neither objective requires a policy, we can have a policy-agnostic pipeline that integrates the learned memory into different VLA backbones through a lightweight adapter and jointly fine-tune memory representation with policy via action loss.

Our contributions are threefold.
First, we propose \ours{}, a policy-agnostic latent memory framework. It compresses multi-stream histories into latent memory efficiently at inference and trains with history reconstruction and progress estimation before fine-tuning it jointly with the policy. Second, we analyze how memory capacity, encoding, training objectives, and recipes affect downstream performance. Third, we evaluate \ours on RMBench~\citep{chen2026rmbench}, a customized 12-task RoboCasa benchmark~\citep{nasiriany2024robocasa,shah2026scaling} spanning memory and non-memory tasks, and three real-world tasks. \ours{} improves success over the base VLA by \textbf{47.4\% on RMBench, and 50\% in the real world}, while maintaining lowest average latency of \textbf{122.8 ms} per action chunk and highest success among the memory baselines included in our latency comparison.

\section{Related Work}

\para{Memory Benchmarks for Robotic Manipulation}
Memory has been studied in mobile manipulation, where robots must retain spatial knowledge across environments~\citep{zhang2025benchmarking,liu2025dynamem}. For tabletop manipulation, MIKASA-Robo, RMBench~\citep{chen2026rmbench}, and RoboMME~\citep{dai2026robomme} probe diverse memory tasks, largely in controlled visual environments with simple objects and limited manipulation difficulty. ReMemBench~\citep{shah2026scaling} and MemMimic~\citep{gao2026gated} introduce more complex manipulation, but offer limited scene and background diversity. These dedicated benchmarks primarily target memory-dependent tasks; MemMimic separately evaluates non-memory RoboMimic tasks outside its suite~\citep{gao2026gated}. More broadly, RoboDojo~\citep{chen2026robodojo} evaluates memory alongside broader manipulation capabilities, but evaluates background and scene randomization separately in its generalization tasks. Our RoboCasa-based benchmark~\citep{nasiriany2024robocasa} jointly spans memory and non-memory tasks across diverse objects and household environments, while varying manipulation complexity.

\para{Memory-Augmented Robot Policies}
Memory has been explored across different classes of robot policies. Recurrent states and extended observation histories help visuomotor policies handle partial observability and long-horizon dependencies~\citep{shah2026scaling,torne2025learning,gao2026gated, jiang2026robottt0}. Structured memory methods retain object-centric scene graphs for exploration and control~\citep{jiang2024roboexp,qian2026expanding,banwasi2026messymem}. 
For pretrained VLAs, existing approaches retain selected keyframes~\citep{mark2026bpp,yang2026eventvla,shah2026halo,sridhar2026scaling}, combine short-term video memory with long-term textual summaries, ~\citep{torne2026mem}, or learn latent memory representations~\citep{shi2026memoryvla,koo2026hamlet}. Beyond VLAs, MemoryWAM~\citep{yang2026memorywam} explores persistent memory in world-action models, while MaP-WAM~\citep{mapwam} converts episodic language and visual memories into plans for execution. Our work focuses on memory for pretrained VLAs, studying how compact latent representations of multi-stream histories should be directly supervised and jointly adapted with control.

\para{Learning Memory Representations}
Learning compact latent memory requires supervision for what information should survive compression. Reconstruction objectives preserve information from sequential experience, including visual content~\citep{wang2023videomae}, latent dynamics~\citep{hafner2019learning}, and past actions~\citep{torne2025learning}, while time-contrastive learning~\citep{sermanet2018time} and progress prediction~\citep{zhai2025vision} provide signals about task progression. For latent memory, HAMLET~\citep{koo2026hamlet} uses time-contrastive initialization on visual history alone, while Workspace Models~\citep{dashora2026workspacemodelslightweightrobotic} reconstruct only salient visual content. In contrast, we supervise memory on all three streams, reconstructing visual, action, and proprioceptive histories together with task progress. Rather than aiming for faithful reconstruction, this supervision \emph{primes} the bottleneck, so that joint fine-tuning with the policy is less likely to learn spurious memory shortcuts from demonstrations.

\section{Problem Formulation}
\label{sec:problem}

\para{Notation \& Setup}
At any real time $t$, the robot observes RGB camera images $\img_t \in \imgSpace$ from cameras and proprioceptive states $\proprio_t \in \proprioSpace$, which we concatenate as $\obs_t := [\img_t, \proprio_t]$. Let $\action \in \actSpace$ denote a low-level action (e.g., joint states), $\acttraj_t := \actchunk$ as an $h$-step action chunk, and $\task$ as the task description. In this paper, we focus on VLAs with parameters $\theta$ that acts only on the current observation, $\acttraj_t \sim \pi_{\theta}(\cdot \mid \obs_t, \task)$. We denote the history by $
\hist_t := \big(\obs_{\le t},\; \action_{<t}\big)$
which contains all past visual observations, proprioceptive states, and actions. $\dataset$ is a set of demonstrations to train policies.

\para{Memory-Augmented Policies}
Many manipulation tasks require information beyond current observations: two histories $\hist_t \neq \hist'_t$ can end at the same observation $\obs_t$ yet call for different actions,
$
\pi(\acttraj_t \mid \obs_t, \task, \hist_t) \neq \pi(\acttraj_t \mid \obs_t, \task, \hist'_t)$.
For example, while a robot dusts a table, the current scene no longer shows original locations or directions to sweep, but returning the objects and finishing the dusting require that. To handle such \emph{history-dependent} steps, a memory-augmented policy $\policy^\textrm{mem}$ usually summarize the history into a memory representation $m_t$ through a function $f$ and conditions on it to generate actions:
$m_t = f(\hist_t, \task),
\acttraj_t \sim \policy^\textrm{mem}(\cdot \mid \obs_t, \task, m_t).$

The central question is therefore how to design the memory function $f$ without restricting what it can represent, as explicit forms do. Since $\hist_t$ grows continuously and spans heterogeneous visual, action, and proprioceptive streams, retaining it in full is neither scalable nor necessary for control. Instead, $f$ should compress the growing history into a bounded representation that preserves the information needed for future decisions. We therefore consider the setting where $f$ is a learned latent memory encoder $\memenc$ that maps the multi-stream interaction history to a continuous representation:
$
\mem_t = \memenc(\hist_t, \task) \in \mathbb{R}^{\memtok \times d},$
where $\memtok$ does not grow with $t$. Our goal is a memory representation $\mem_t$ that compresses the multi-modal history context $\hist_t$ without changing the action distribution:
\begin{equation}
\pi(\acttraj_t \mid \obs_t, \task, \hist_t) = \pi(\acttraj_t \mid \obs_t, \task, \mem_t).
\label{eq:sufficiency}
\end{equation}
To approach this goal, the encoder and policy are learned from demonstrations by minimizing an objective $\loss$ that depends on the memory:
\begin{equation}
\min_{\theta, \phi}\; \mathbb{E}_{\dataset, t}\Big[\loss\big(\hist_t, \acttraj_t;\; \policy,\; \memenc\big)\Big].
\label{eq:general_objective}
\end{equation}

Achieving \eqref{eq:sufficiency} through \eqref{eq:general_objective} raises two challenges: how to design the encoder $\memenc$, and how to choose the objective $\loss$.

\para{Challenge 1: Encoding Heterogeneous Interaction Histories Online}
The history combines a high-dimensional video stream with low-dimensional proprioceptive states and actions, which differ in scale and structure.
The video stream is particularly hard to compress: both spatial and temporal dimensions have high redundancy, yet the small changes between them, such as object locations, matter for later decisions. Moreover, the history grows at every step while memory is queried at every control step, so re-encoding the full history adds latency that grows with the episode length.

\para{Challenge 2: Choosing the Memory Objective}
The standard choice for Eq.~\ref{eq:general_objective} is end-to-end optimization through action loss~\citep{shi2026memoryvla}, where the memory is learned only through gradients backpropagated from the policy:
\begin{equation}
\loss_{\textrm{act}}(\theta, \phi) = \mathbb{E}_{\dataset, t}\Big[-\log \policy^\textrm{mem}\big(\acttraj_t \mid \obs_t, \task, \memenc(\hist_t, \task)\big)\Big].
\label{eq:action_objective}
\end{equation}
This objective supervises memory only indirectly through action prediction, coupling what the bottleneck should retain with how the policy uses it for control. In particular, it provides no explicit supervision that memory preserves information from each heterogeneous history stream.

\section{\ours: Interaction Memory with Progress- and Reconstruction-\\Informed Training}
\label{sec:method}
We address the challenges in Sec.~\ref{sec:problem} with efficient encoding and two-stage learning. We first compress visual, action, and proprioceptive histories into a fixed-size latent memory (Sec.~\ref{sec:encoding}). We then bootstrap memory with history reconstruction and progress estimation (Sec.~\ref{sec:stage1}), before jointly adapting the memory and VLA with action supervision for control (Sec.~\ref{sec:stage2}).

\subsection{Multi-stream Memory Encoding}
\label{sec:encoding}

The first challenge lies in the efficiency to compress a growing, heterogeneous interaction history into latent memory. We address this with two levels of compression: within stream and across streams over time. Because video dominates the history in both dimensionality and redundancy, we first compress it with a frozen causal video tokenizer~\citep{wan2025wan,agarwal2026cosmos}. The memory encoder then integrates the resulting visual tokens with actions and proprioceptive states over time into a fixed number of latent memory tokens as shown in left panel of Fig.~\ref{fig:method_overview}.

\para{Multi-Stream Memory Tokenization}
We represent history through four token streams: \emph{historical video}, \emph{recent images}, \emph{executed actions}, and \emph{proprioceptive states}. Let $\img_{\leq n_t}$ denote the previous observed frames available at time $t$. With $L$ frames per chunk, $K_t=\lfloor n_t/L \rfloor$ chunks are complete. A frozen causal video tokenizer $\tokenizer^{\text{vid}}$~\citep{wan2025wan} divides these frames $\img_{\leq K_tL}$ into chunks and then tokenize them into a small set per camera. The remaining frames $\img_{K_tL+1:n_t}$ form the recent-image stream, tokenized by off-the-shelf DINOv3 encoder $\tokenizer^v$~\citep{simeoni2025dinov3} to construct visual features $\latent_k^v$. Separate Multi-Layer Perceptrons (MLPs) embed previously executed actions $\action_{< t}$ and proprioceptive states $q_{\leq t}$ into the other two streams.

\para{Memory Bottleneck}
Let $\token^{\mathrm{vid}}_t$, $\token^{\mathrm{recent}}_t$, $\token^{\mathrm{act}}_t$, and $\token^{\mathrm{state}}_t$ denote the outputs of the four tokenization streams, and let $\token_t = (
\token^{\mathrm{vid}}_t,
\token^{\mathrm{recent}}_t,
\token^{\mathrm{act}}_t,
\token^{\mathrm{state}}_t)$ denote their concatenation.We introduce $\memtok$ learned query embeddings 
that read out
information from the four token streams conditioned on the
task description $\task$, following~\citep{jaegle2021perceiver}. The memory encoder computes
$\mem_t = \memenc\left(\token_t;
\task
\right)
\in \mathbb{R}^{\memtok\times d}$. The learned queries are shared across episodes, and together with language features of $\task$, compress the history into
$\memtok$ memory tokens. We consider two bottlenecks with the
same supervision (Sec.~\ref{sec:supervision}). Writing the four
token streams collectively as $\token_t$, they compute
\begin{align}
\text{Full Attention:}\quad
\mem_t &= \memenc\left(
\token_t; \task
\right), \\
\text{Recurrent:}\quad
(\mem_t, h_t) &= \memenc\left(
\token_t; h_{t-1}, \task
\right).
\end{align}

i) \emph{Full Attention bottleneck.}
Following Perceiver~\citep{jaegle2021perceiver} and Workspace Models~\citep{dashora2026workspacemodelslightweightrobotic}, the queries
cross-attend to all history tokens, including the recent window.
Each query revisits the full history, so storage and per-query
cost grow linearly with history length. 

ii) \emph{Recurrent bottleneck.}
We use gated delta (GDN) blocks~\citep{yang2025gated} to process each chunk $e_t$ by updating fixed-size
matrix states, collectively denoted by $h_t$. The queries read
these states and then attend to the recent window, keeping
storage and per-query cost bounded independently of history
length. Training uses the parallel formulation of the gated
delta blocks.

\subsection{Decoupled Memory Supervision}
\label{sec:supervision}
Optimizing Eq.~\ref{eq:general_objective} requires learning both what information memory should retain and how the policy should use it for control. With action loss alone, the memory encoder is guided only indirectly through action prediction, providing no explicit requirement that it preserve information from each history stream. We therefore decouple these two roles: first learn memory with direct, policy-independent supervision, and then jointly adapt the memory and policy with action loss as in Fig.~\ref{fig:method_overview}.

\textbf{Memory Objective \& Training Pipeline.}
We first learn latent memory with two complementary self-supervised objectives that force it to capture both information from interaction histories and global task context. \emph{History reconstruction} requires $\mem_t$ to reconstruct past visual features, actions, and proprioceptive states, to retain information from each stream. However, reconstruction alone does not explicitly capture the task stage. We therefore use \emph{progress estimation}, to encourage $\mem_t$ to predict task progress and thereby encode global temporal context. Both objectives are policy-independent and annotation-free, implemented with a decoder $\dec$ that operates on $\mem_t$.

Once memory representations are learned, we adapt them for control by jointly training the memory encoder and policy via the action-prediction objective, leading to the two-stage training procedure:
\begin{equation}
\begin{aligned}
\text{Stage 1:}\quad & \bar{\phi},\, \bar{\varphi} = \argmin_{\phi,\,\varphi}\;
\mathbb{E}_{\dataset, t}\Big[\loss_{\text{mem}}\Big(
\dec\big(\memenc(\hist_t, \task)\big),\; \mathbf{y}_t\Big)\Big], \\
\text{Stage 2:}\quad & \theta^\star,\, \phi^\star = \argmin_{\theta,\,\phi}\;
\mathbb{E}_{\dataset, t}\Big[-\log
\policy\big(\acttraj_t \mid \obs_t, \task, \memenc(\hist_t, \task)\big)\Big], \quad \phi^{(0)} = \bar{\phi}
\end{aligned}
\label{eq:two_stage}
\end{equation}
where $\mathbf{y}_t$ represents the reconstruction targets at sampled past steps and the progress $\rho_t$, and $\loss_{\text{mem}}$ is the prediction error defined in \eqref{eq:stage1}. Stage~1 trains the encoder $\memenc$ and decoder $\dec$ without the policy. Stage~2 discards the decoder and initialize encoder $\memenc$ from parameters $\bar{\phi}$ and finetunes the VLA policy $\policy$ jointly with the memory encoder, so that the action objective adapts latent memory toward decision-relevant content and teaches the policy to read it.

\subsubsection{Stage 1: Memory Pre-training}
\label{sec:stage1}

To use for multi-stream memory, the decoder $\dec$ reconstructs $K$ historical steps sampled uniformly at each timestep $t$~\citep{jaegle2021perceiver}. For each sampled step $k \le t$, it forms one query per visual patch and one summary query, all using the embedding for $k$ from a learned timestep table shared across episodes. These queries attend to each other within the step and cross-attend to $\mem_t$. The patch queries predict that step's visual features $\hat{z}^v_k$, while the summary query predicts its action $\hat{\action}_k$ and state $\hat{\proprio}_k$. A separate global query predicts progress $\hat{\rho}_t$. The loss is:
\begin{equation}
\loss_{\text{mem}} =
\underbrace{\frac{1}{K}\sum_{k}\Big[
\lambda_v \big\| \hat{z}^v_k - z^v_k\big\|^2
+ \lambda_a \big\|\hat{\action}_k - \action_k\big\|^2
+ \lambda_q \big\|\hat{\proprio}_k - \proprio_k\big\|^2 \Big]}_{\text{history reconstruction}}
+ \underbrace{\lambda_\rho \big(\hat{\rho}_t - \rho_t\big)^2}_{\text{progress estimation}},
\label{eq:stage1}
\end{equation}
where $z^v_k = \tokenizer^v(\img_k)$ are frozen, normalized patch features and $\rho_t \in [0,1]$ is the normalized progress of the demonstration (i.e. current time-step over total episode length).   %
The first three terms supervise history reconstruction at individual past steps, and the last term supervises progress estimation over the entire history, thereby integrating all relevant streams. These objectives aim to learn useful memory representations rather than achieve perfect compression. Together, they \emph{prime} memory tokens with complementary scene, motion, and task-progress information for downstream policy fine-tuning, helping reduce reliance on spurious memory shortcuts in the demonstrations.

\subsubsection{Stage 2: Policy-Agnostic Memory Adaptation}
\label{sec:stage2}
\para{Memory Interface}
Stage~1 produces a policy-independent memory encoder that can integrate
with different VLAs through a lightweight action-expert adapter,
following~\citep{dai2026robomme} (Fig.~\ref{fig:method_overview}). At each layer, action tokens $x$
cross-attend to $\mem_t$ and receive FiLM modulation~\citep{perez2018film}:
$\tilde{x} = \mathrm{FiLM}\big(x,\; \mathrm{CrossAttn}(x, \mem_t)\big).$
This preserves the VLA's native inputs, context length, and conditioning,
while better preserving pretrained representations than appending memory
to the VLM context~\citep{dai2026robomme}. Attending to a fixed number of
memory tokens adds little cost, independent of history length.
We instantiate this adapter in different VLA backbones including $\pi_0$~\citep{black2024pi_0},
$\pi_{0.5}$~\citep{intelligence2025pi}, and
MolmoAct2~\citep{fang2026molmoact2} (App.~\ref{app:vla_integration}).

\para{Joint Adaptation}
Following Stage~2 of \eqref{eq:two_stage}, we discard the decoder and initialize the policy from $\theta_0$ and the encoder from $\bar{\phi}$. The adapter is initialized near zero, so training starts close to the original VLA. We then optimize the action objective \eqref{eq:action_objective} with the VLA's native loss (e.g., flow matching), updating the encoder, adapter, and VLA. Because the learned latent memory already carries critical information from the history, the policy can easily rely on that to generate correct actions, and the action gradient in turn further refines memory for control.

\section{Experiments}
\label{sec:experiments}

We evaluate whether \ours improves manipulation tasks with different memory requirements while keeping memory compact and efficient (full details in App.~\ref{app:experimental_protocols} to~\ref{app:failure_analysis}).
We first compare \ours against prior memory methods in simulation and real world, on task success and inference latency in Sec.~\ref{sec:exp_overall_eval}. We then analyze \emph{memory encoding}, how memory capacity and recurrent compression affect memory representation and performance in Sec.~\ref{sec:exp_encoding}; and \emph{memory learning}, how memory supervision and adaptation affect control, and its flexibility to different VLAs in Sec.~\ref{sec:exp_learning}.

\subsection{Experimental Setup}
\label{sec:exp_setup}

\para{Benchmarks and Tasks}
We evaluate \ours on two simulation benchmarks and three real-world manipulation tasks. In simulation, we use all 9 memory-dependent tasks from RMBench~\citep{chen2026rmbench} and a customized 12-task benchmark built on RoboCasa~\citep{nasiriany2024robocasa} and ReMemBench~\citep{shah2026scaling}, containing 6 memory and 6 non-memory tasks with diverse objects and scenes. This setup evaluates both the benefits of memory and its impact on tasks without memory necessity.
In the real world, we evaluate three memory-dependent tasks on a bimanual YAM platform: \emph{Dust Table}, \emph{Scan Bottle}, and \emph{Fold Cloth}, covering spatial memory of object locations, task-progress tracking of scanning, and learning from in-context robot demonstrations of varied folding styles. Task details are provided in Appendix~\ref{app:rmbench_tasks},~\ref{app:robocasa_tasks},~\ref{app:realworld_tasks}.

\para{Baselines}
We compare \ourscolored against three VLAs, $\pi_0$, $\pi_{0.5}$, MolmoAct2 without memory and three categories of memory-augmented VLAs from prior work: sampling visual frames with heuristics \robomme and \eventvla ~\citep{dai2026robomme, yang2026eventvla}, hierarchical memory methods with textual summary and keyframes \memzero and \memer ~\citep{chen2026rmbench,sridhar2026scaling}, and latent-memory methods \hamlet and \memoryvla ~\citep{koo2026hamlet,shi2026memoryvla}. We also conduct controlled ablations of encoder, memory capacity, training objectives, and adaptation with VLAs. Details of baselines and ablations are in Table.~\ref{tab:memory_design_comparison}, App.~\ref{app:implementation},~\ref{app:additional_results}.

\para{Training and Evaluation}
Unless otherwise specified, we use $\pi_{0.5}$ as base policy and $\memtok=256$ memory tokens with $d=1024$. We report results over 100 trials per task in RmBench and 50 trials per task in RoboCasa.
Following ~\citep{lbmtri2025}, we plot the smoothed distribution of task success symmetrically, with width indicating density and a horizontal line marking the mean. Implementation details, full breakdown of per-task results, additional ablations, and detailed failure analysis are in App.~\ref{app:implementation}, ~\ref{app:additional_results}, \ref{app:additional_experiments}, and~\ref{app:failure_analysis}.

\subsection{\ours Outperforms Existing Memory-Augmented VLAs with Better Task Performance and more Efficient Inference}
\label{sec:exp_overall_eval}
We first evaluate the overall effectiveness of \ours across two simulation benchmarks and real-world manipulation tasks (App.~\ref{app:results_overall}). We compare against the base VLA and existing heuristic, hierarchical and latent memory approaches, measuring both task success and inference latency.
\begin{wrapfigure}{r}{0.6\linewidth}
    \centering
    \vspace{-0.2cm}
    \includegraphics[width=\linewidth]{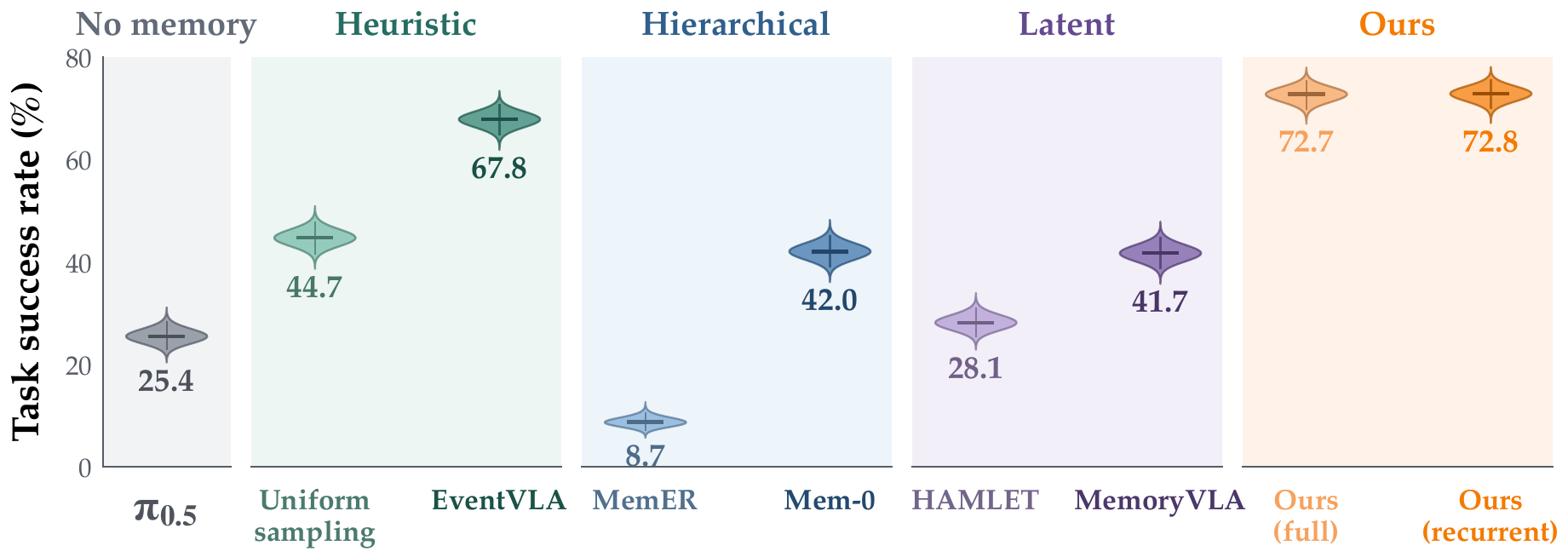}
    \caption{\textbf{\small{Average Task Success on Rmbench}}}
    \label{fig:rmbench_results}
    \vspace{-0.4cm}
\end{wrapfigure}

\para{Simulation Results}
Figure~\ref{fig:rmbench_results} compares \ours with heuristic, hierarchical, and latent memory approaches on 9 memory-dependent RMBench tasks. Heuristic selection or sampling of sparse visual frames helps over the no-memory $\pi_{0.5}$ baseline from 25.4\% to 67.8\% but fails on tasks that need denser information (e.g., press button). Hierarchical methods achieve at most 42.0\% success, caused by the loss of information in the reduced subset of memories and large action jump near the subtask boundaries. Prior latent memories achieve at most 41.7\%, but either operate primarily on visual history or lack direct supervision to preserve information from each history stream. In contrast, recurrent \ours supervises a multi-stream latent memory to retain visual, action, and proprioceptive history while tracking task progress, reaching 72.8\% success. 

\begin{wrapfigure}{r}{0.6\linewidth}
    \centering
    \vspace{-0.4cm}\includegraphics[width=\linewidth]{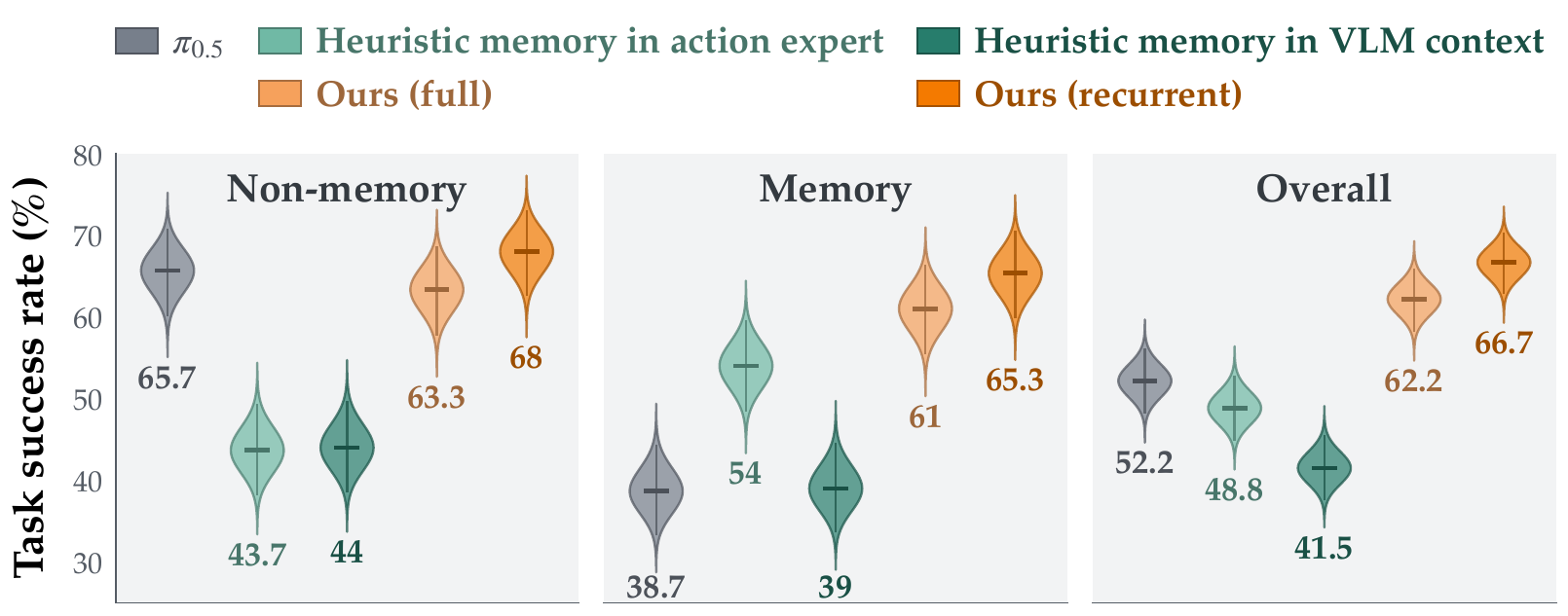}
    \caption{\textbf{\small{Average Task Performance For Memory and Non-Memory Tasks in Robocasa}}}
    \label{fig:robocasa_results}
    \vspace{-0.3cm}
\end{wrapfigure}

Figure~\ref{fig:robocasa_results} examines whether \ours helps with tasks that do not require history. Heuristic memory can improve memory-dependent performance about 15.3\% but substantially degrades non-memory tasks by 22\%, indicating that simply exposing the VLA to additional history can interfere with behaviors already solvable from the current observation. In contrast, recurrent \ours raises memory task success to 65.3\% while slightly improving non-memory task performance to 68.0\%, achieving 66.7\% overall. This suggests that compressing history into a learned latent representation allows the policy to exploit memory when needed without harming non-memory tasks.

\begin{wrapfigure}{r}{0.6\linewidth}
    \centering
    \vspace{-0.6cm}
    \includegraphics[width=\linewidth]{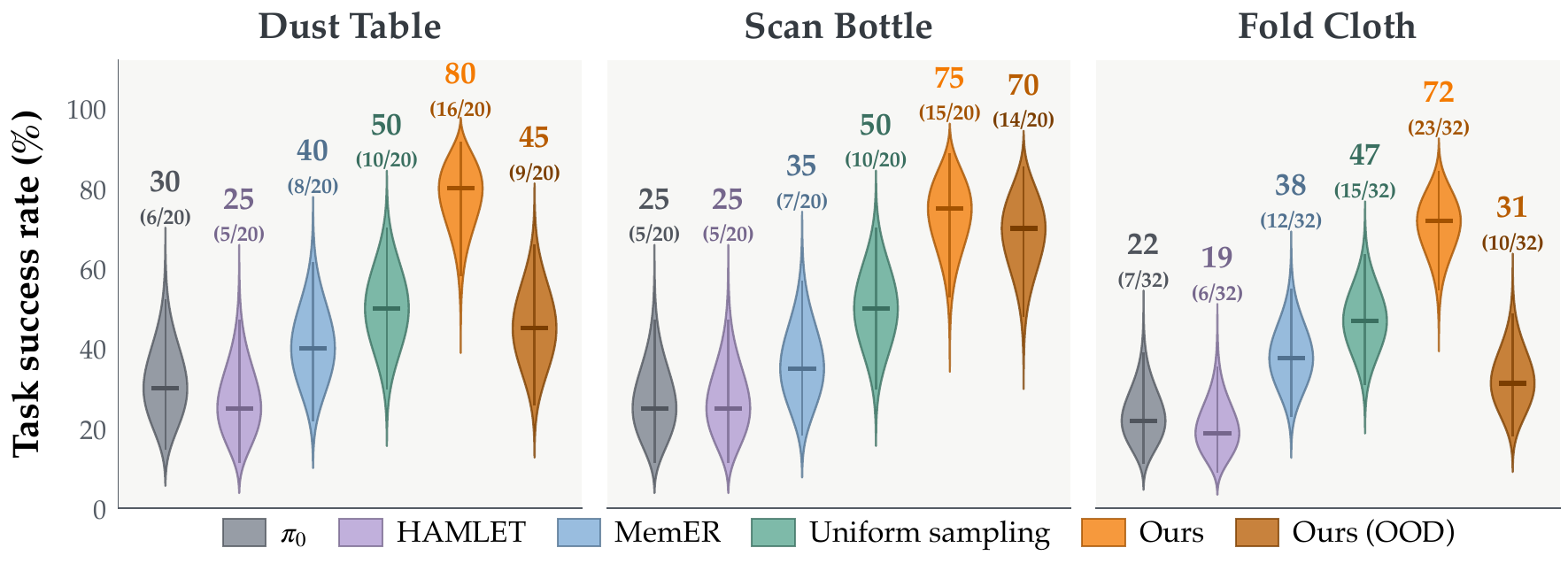}
    \caption{\textbf{Real World Task Performance Across 3 Tasks}}
\label{fig:real_world_results}
    \vspace{-0.7cm}
\end{wrapfigure}
\para{Real-World Results}
Figure~\ref{fig:real_world_results} shows the gains of \ours transfer to real-world tasks. \ours achieves 80\%, 75\%, and 72\% success across three tasks, consistently outperforming all baselines.
Compared with the strongest baseline (uniform sampling), \ours improves average success by 26.7\%.
The consistent gains over all memory baselines suggest that directly learning a compact multi-stream memory is also useful under real-world tasks that requires spatial memory,  progress estimation of scanning, and in-context learning of folding styles.

\para{Real-World OOD Evaluation}
We further evaluate \ours on \textit{unseen} objects (App.~\ref{app:real_world_eval}), achieving success rates of 45\%, 70\%, and 31\% across three tasks (Fig.~\ref{fig:real_world_results}). These results provide encouraging initial evidence that the learned memory can transfer beyond the training configurations. While generalization remains challenging for longer-horizon, precision-sensitive tasks such as \emph{Dust Table} and \emph{Fold Cloth}, this OOD evaluation highlights the potential of \ours to handle novel objects.

\begin{wrapfigure}{r}{0.5 \linewidth}
    \centering
    \vspace{-0.2cm}
    \includegraphics[width=\linewidth]{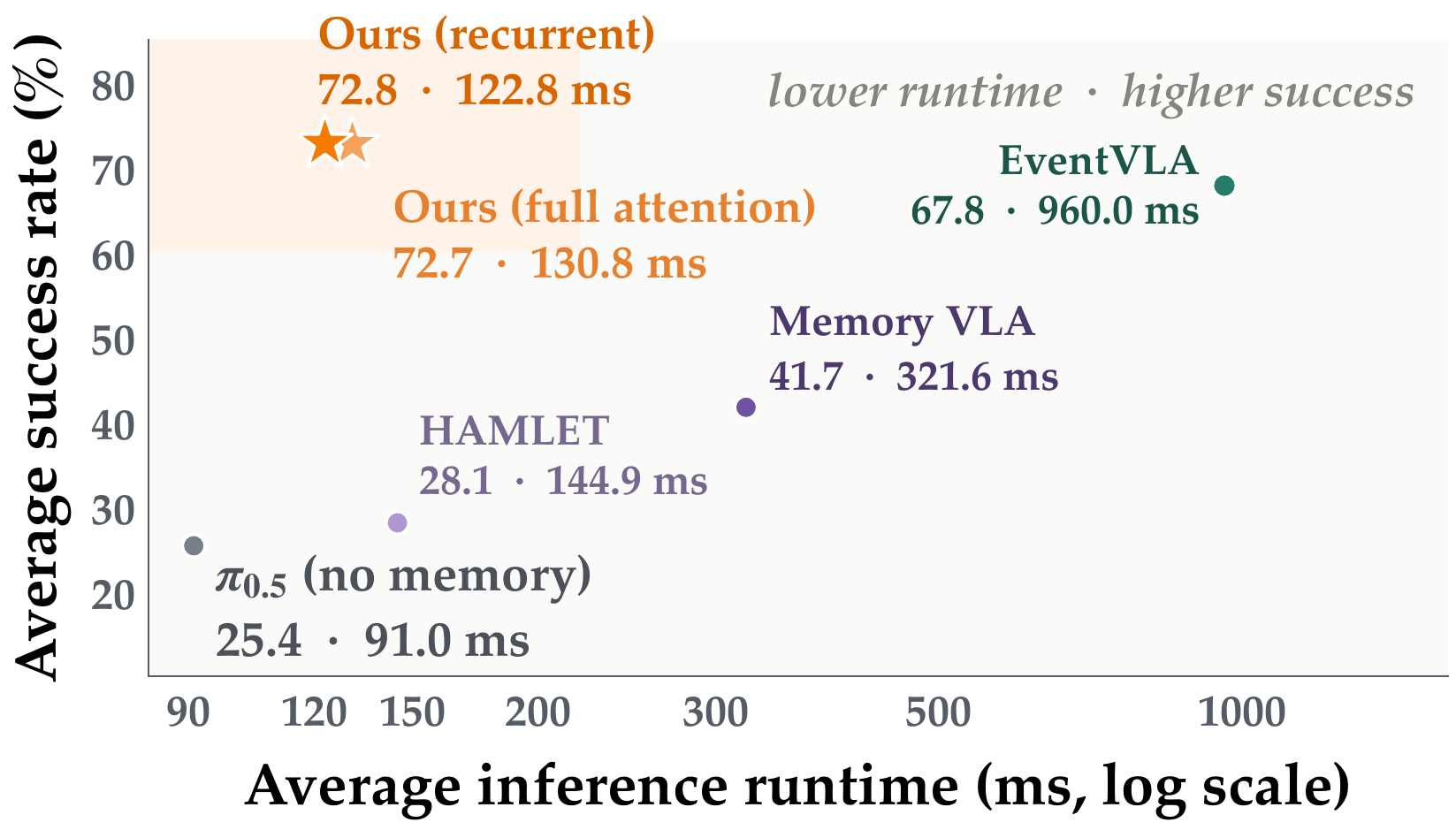}
    \caption{\textbf{Runtime vs Success on RMBench}}
    \label{fig:inference_cost}
    \vspace{-0.2cm}
\end{wrapfigure}

\para{Success and Inference Efficiency}
Figure~\ref{fig:inference_cost} compares task success and average inference latency per 50-step action chunk under L40S GPU. Prior memory methods incur either limited gains or substantial latency: HAMLET and MemoryVLA reach 28.1\% and 41.7\% success at 144 and 321\,ms, while EventVLA reaches 67.8\% but requires 960\,ms due to long context to Qwen backbone. Hierarchical methods are even slower with several seconds from high-level VLMs. In contrast, recurrent \ours achieves the highest success (72.8\%) at 122.8\,ms, closest to the no-memory $\pi_{0.5}$ runtime and dominating all evaluated memory baselines in both success and latency. Full inference time analysis is in App.~\ref{app:timing_protocol}.

\subsection{Richer Multi-Stream Inputs, Greater Capacity, and Recurrent Compression Enable More Effective Memory Encoding}
\label{sec:exp_encoding}
We next study the encoding challenge from Sec.~\ref{sec:problem}: compressing a growing interaction history into a continuous latent representation. We examine three properties of the memory bottleneck: its \emph{input}, which evaluates whether multi-stream information is necessary for the task; its \emph{capacity}, which controls how much historical information can be retained, and its \emph{update mechanism}, which determines how efficiently history is accumulated over time. Detailed results are in App.~\ref{app:results_encoding}.

\begin{figure}
    \centering
    \includegraphics[width=\linewidth]{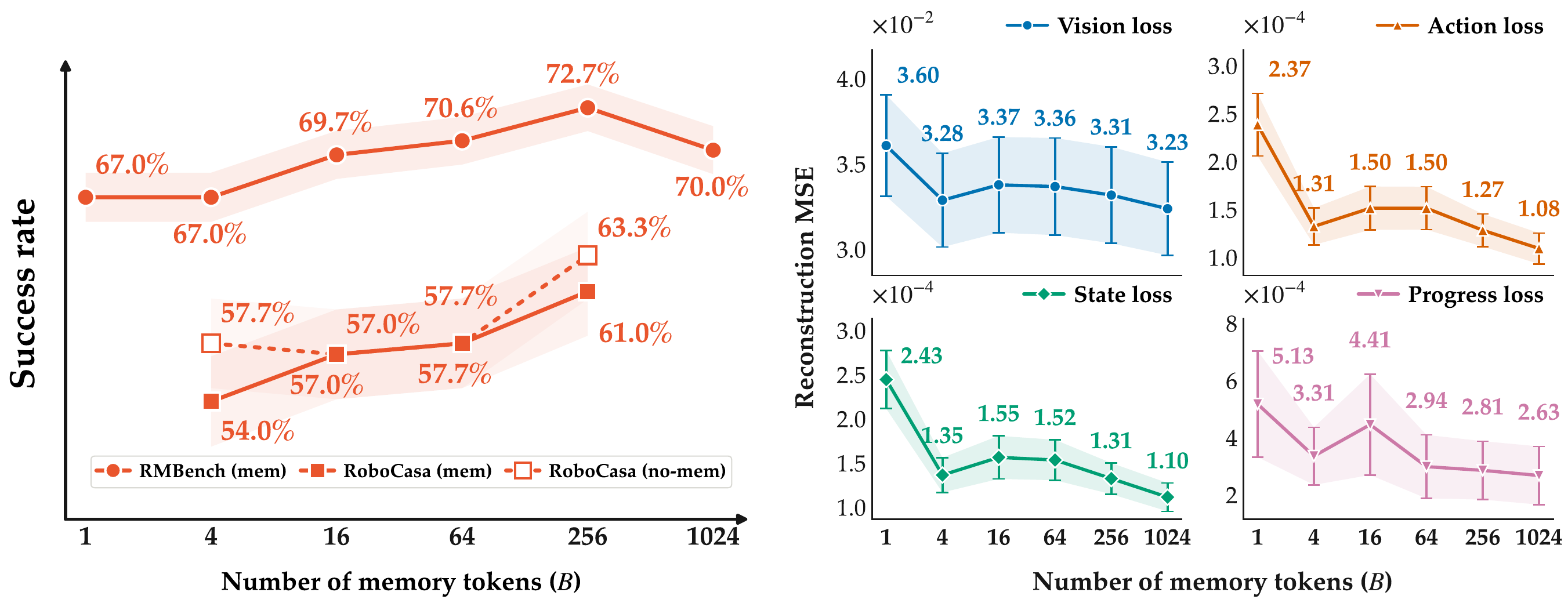}
    \caption{\textbf{Ablations on Memory Capacity.} We evaluate \ours trained with different memory tokens against task performance (left) and supervision loss evaluated on the validation set (right). }
    \label{fig:memory_capacity}
\end{figure}

\para{Multi-Stream History Provides Complementary Information}
We ablate the different input streams of histories supplied to memory while holding the full attention encoder and two-stage training recipe fixed. In Fig.~\ref{fig:training_ablation} (right), Vision alone achieves 60.6\% success; adding action history improves performance to 68.4\%, while the visual-and-state variant reaches 60.9\%. \ours which combines all streams of histories performs best at 72.7\%, showing that those streams provide complementary information for memory-dependent control and specific curation of memory to certain stream, e.g., vision, can decrease the task performance.

\para{More Capacity Preserves More Useful Context}
In Fig.~\ref{fig:memory_capacity}, with the full-attention encoder and training recipe fixed, increasing memory capacity from 1 to 256 tokens raises RMBench success from 67.0\% to 72.7\%, and increasing it from 4 to 256 tokens raises RoboCasa memory-task success from 54.0\% to 61.0\%. Non-memory success stays near 57\% through 64 tokens before reaching 63.3\% at 256. However, 1024 tokens yields the lowest reconstruction losses but lowers RMBench success to 70.0\%, and increasing from 4 to 16 tokens improves success despite higher reconstruction errors, suggesting that reconstruction serves to \emph{prime} useful memory representations for policy adaptation rather than to achieve perfect compression.

\para{Recurrent Compression Improves Encoding Efficiency}
Beyond capacity, the memory encoder must encode a growing history efficiently. At 256-token capacity, our recurrent encoder matches or improves the full attention variant's performance while reducing average inference latency from 130.8 ms to 122.8 ms per action chunk. Both achieve approximately 72.8\% success on RMBench (Fig.~\ref{fig:rmbench_results}), while recurrent encoding improves overall RoboCasa success from 62.2\% to 66.7\% (Fig.~\ref{fig:robocasa_results}).

\subsection{Memory Pretraining and Joint Adaptation Improve Task Performance}
\label{sec:exp_learning}

\begin{wrapfigure}{r}{0.7\linewidth}
    \centering
    \vspace{-0.4cm}
    \includegraphics[width=\linewidth]{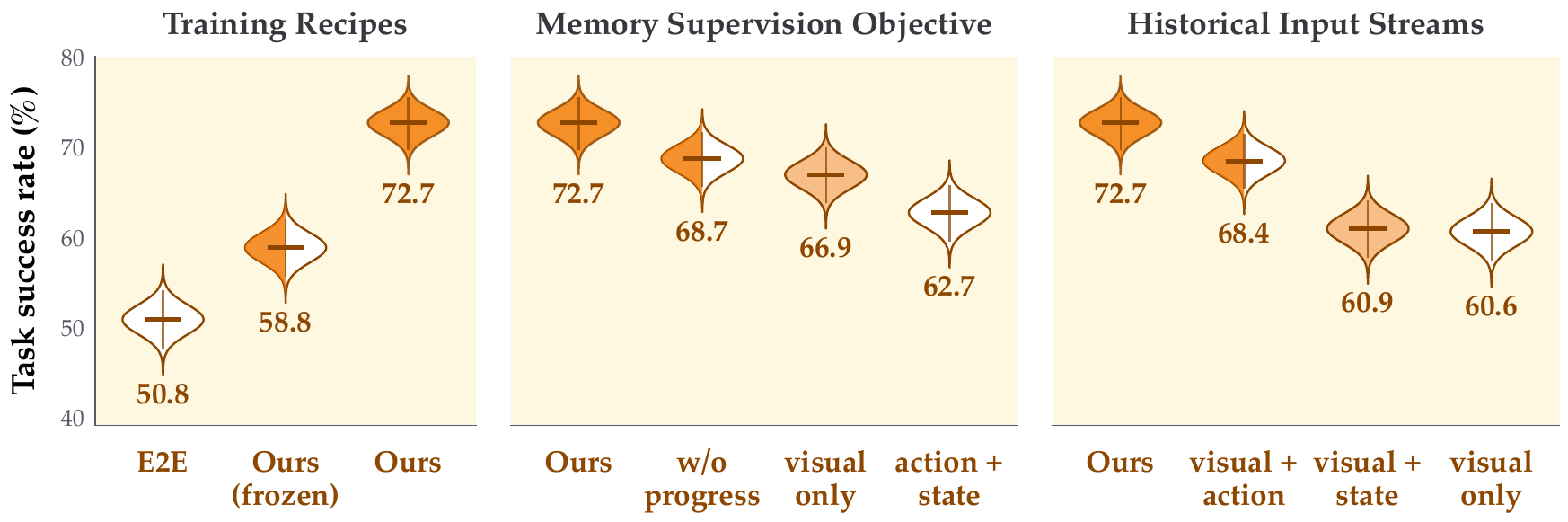}
    \caption{\textbf{Ablations of Memory Learning.} We ablate different input streams (right) and varied supervision objectives (middle) and training recipes (left) of \ours on RMBench.}
    \label{fig:training_ablation}
    \vspace{-0.6cm}
\end{wrapfigure}

\para{Pretraining and Joint Adaptation Are Both Important}
We compare learning memory end-to-end via action supervision (E2E) in Eq.~\ref{eq:action_objective}, frozen pretrained memory without adaptation, and pretraining followed by joint memory--VLA adaptation. In Fig.~\ref{fig:training_ablation} (left), success improves from 50.8\% (E2E) to 58.8\% with bootstrapping alone, and further to 72.7\% with joint adaptation. This shows that both initial memory learning and subsequent adaptation for action prediction are important.

\begin{figure}[t]
    \centering
    \includegraphics[width=\linewidth]{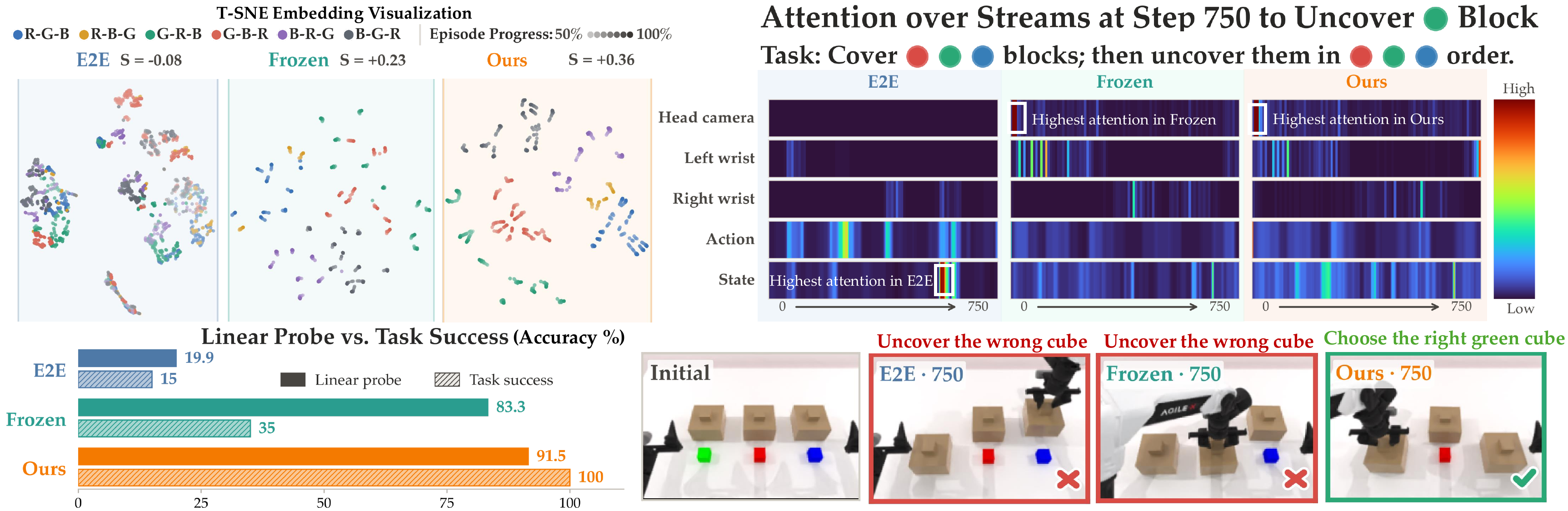}
    \caption{\textbf{Analysis of Memory Bottleneck.}
    We compare E2E, frozen memory encoder, and our adapted memory with t-SNE embeddings, linear probes, and attention visualization on \emph{Cover Block}.}
    \label{fig:memory_bottleneck_analysis}
    \vspace{-0.6cm}
\end{figure}

\para{Two-stage Training Produces More Informative Memory}
Figure~\ref{fig:memory_bottleneck_analysis} examines what memory retains and its impact on task success on \emph{Cover Block} (App.~\ref{app:representation_analysis}). Memory pretraining improves linear-probe accuracy of classifying the color order by 63.4\%, reaching 91.5\% after joint adaptation, while \ours improves T-SNE structure to $S=0.36$ with better class separation. Attention from both frozen and \ours focus on the initial frames of head camera with color order revealed, whereas E2E focus on state history. However, even with right attention, only joint adaptation enables \ours to uncover blocks in the correct order, showing pretraining improves retention of critical information while adaptation optimizes it for action prediction.

\para{Reconstruction and Progress Estimation are Complementary Objectives}
We next ablate different supervision objectives with the fixed adaptation in Fig.~\ref{fig:training_ablation} (middle). The combination of visual, action, and state reconstruction with progress estimation obtains 72.7\% success. Removing progress estimation reduces success to 68.7\%, while using only visual reconstruction or only action and state reconstruction further decreases success to 66.9\% and 62.7\%, respectively. Memory thus benefits from both history retention and progress estimation.

\begin{wrapfigure}{r}{0.45\linewidth}
    \centering
    \vspace{-0.5cm}
    \includegraphics[width=\linewidth]{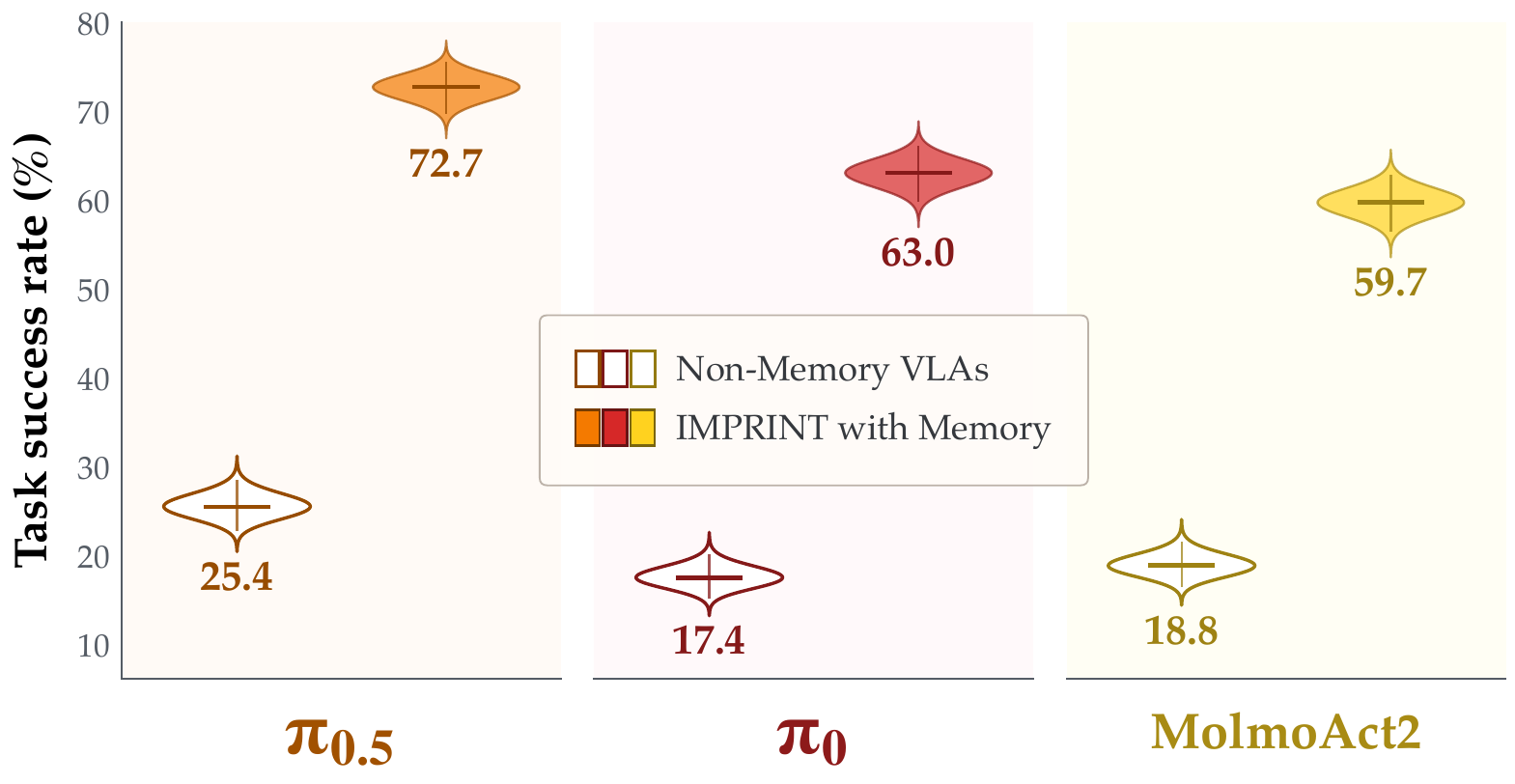}
    \caption{\textbf{\ours Improves Different VLA Backbones with Memory.}}
    \label{fig:flexible_vla}
    \vspace{-0.5cm}
\end{wrapfigure}

\para{\ours is Policy-agnostic across VLAs}
Because Stage~1 learning is independent of the downstream policy, we apply the same memory pipeline and adaptation procedure to three different VLA backbones. In Fig.~\ref{fig:flexible_vla}, memory augmentation improves success from 25.4\% to 72.7\% for $\pi_{0.5}$, from 17.4\% to 63.0\% for $\pi_0$, and from 18.8\% to 59.7\% for MolmoAct2. The consistent gains across different base policies show that the proposed memory bootstrapping and joint-adaptation recipe is not tied to a particular VLA. Details are in App.~\ref{app:vla_integration} and Table~\ref{tab:backbone_per_task}.

\section{Conclusion and Limitations}
We presented \ours{}, a latent memory framework that augments VLAs with multi-stream memory without constraining what the policy should remember. \ours{} compresses visual, action, and proprioceptive histories into a latent memory, supervised with reconstruction and progress estimation before joint fine-tuning with VLAs. Across simulation, real-world, and three VLA backbones, \ours{} outperforms base VLAs and other memory approaches with modest inference overhead.

\para{Limitations}
Latent memory is less interpretable than explicit memory, making failure analysis more challenging. Extending the memory to include failures for in-context learning could further broaden \ours applicability. Currently, \ours{} learns memory representations solely from teleop. demonstrations; incorporating more scalable data sources is an exciting future direction.

\section{Acknowledgments}

We thank Yihui Gao and Zhanyi Sun for valuable discussions on memory learning during the early stages of brainstorming for this work. We also thank Zhenyu Jiang and Huihan Liu for sharing insights into real-world robotic tasks and the importance of memory. We are grateful to Chaoyi Pan for assistance with the research infrastructure, and to Haoru Xue and Tonghe Zhang for helpful discussions on in-context learning.

\newpage
\newpage
\bibliographystyle{plainnat}
\bibliography{reference}
\newpage

\newpage
\tableofcontents

\newpage
\appendix

\section{Implementation Details}
\label{app:implementation}

This section describes the implementation of \textsc{IMPRINT},
the baseline methods we compare to, and the configurations used in our ablations.

\subsection{IMPRINT Implementation}
\label{app:imprint_implementation}

\textsc{IMPRINT} is our proposed two-stage latent memory learning pipeline introduced
in Sec.~\ref{sec:method}: multi-stream interaction history is first bootstrapped with
reconstruction and task-progress estimation, and the resulting memory is
then jointly adapted with a VLA for control. Following the notation in
Sec.~\ref{sec:problem},  $B$ denotes the number of memory tokens and
$m_t\in\mathbb{R}^{B\times d}$ denotes the memory at decision step $t$. The
default description of \ours uses the recurrent bottleneck; the full-attention
bottleneck is a matched encoder variant trained with the same objectives and
memory interface. The frozen video tokenizer performs within-stream
compression, while the memory bottleneck integrates information across
streams and over time.

\subsubsection{Multi-stream Input Processing and Tokenization}
\label{app:input_processing}

\paragraph{Observations.}
For RMBench, each observation contains three synchronized RGB views in the
fixed order head camera, left wrist camera, and
right wrist camera. The simulator renders each view at $320\times240$
pixels. Before visual feature extraction, we resize each image to
$256\times256$. We downsample the full history by 2 for memory construction. For customized RoboCasa, each observation contains three synchronized RGB views, ordered for memory
  construction as agentview left,
  agentview right, and
wristview. Each view is rendered and stored at
  $256\times256$ pixels. 
  For real world tasks,
  each observation contains three synchronized RGB views in the fixed order
  head camera, left wrist camera, and
  right wrist camera, corresponding to the head camera and the two
  wrist cameras. All cameras capture images at $640\times480$ pixels. We decode
  the recorded videos in RGB format and resize each image to $256\times256$
  using bilinear interpolation with antialiasing before DINOv3 feature
  extraction. 

\paragraph{Consolidated visual history.}
We encode finalized history chunks with the frozen
\texttt{Wan2.2-5B} video VAE~\citep{wan2025wan}. Its temporal factor is
four retained frames; because memory retains every second simulator frame,
one finalized chunk spans eight raw simulator frames. The VAE produces
48-dimensional latent features. We average-pool each camera's latent spatial
grid to $2\times2$ tokens, yielding $3\times4=12$ tokens per finalized
multiview chunk. A chunk enters consolidated memory only after all four
retained frames are available.

\paragraph{Recent visual history.}
Frames following the last finalized video chunk ($<=3$) are encoded by the frozen
\texttt{dinov3} image encoder. At
$256\times256$ resolution, ViT-B/16 produces a $16\times16$ patch grid after
removing prefix tokens. We average-pool this grid to $8\times8$, producing
64 tokens of width 768 per view and 192 tokens per synchronized multiview
frame. The same DINOv3 tensor is used as the recent visual input to the memory
encoder and as the Stage--1 visual reconstruction target.

In some ablations, we used a separate SigLIP/PaliGemma representation from $\pi_{0.5}$ visual encoder and use  for the reconstruction target from the VLM prefix, both with width
2048. We treat this as a distinct input profile.

\paragraph{Robot signals and causal alignment.}
RMBench uses a 14-dimensional absolute joint action/state representation,
which is zero-padded to 16 dimensions for the memory model. The main
configuration uses raw action and raw proprioceptive vectors rather than
first differences. For a sampled store variant with retained frame indices
$f_0,\ldots,f_n$, the memory action stream is
\begin{equation}
  a^{\mathrm{mem}}_0=\mathbf{0},\qquad
  a^{\mathrm{mem}}_j=a^{\mathrm{raw}}_{f_{j-1}}\quad (j>0).
  \label{eq:appendix_causal_action}
\end{equation}
Thus, the observation at retained index $f_j$ is paired only with an action
that has already been executed. Importantly, this shift is constructed
separately for every temporally sampled variant; indexing a canonical shifted
sequence at skipped frame indices is not equivalent. The Stage--1 decoder
target for query $j$ remains the current, unshifted
$a^{\mathrm{raw}}_{f_j}$.

\paragraph{Language.}
The memory encoder is conditioned on one 2048-dimensional language embedding of the task instruction $\task$ per
episode from $\pi_{0.5}$ language encoder. Stage--1
and Stage--2 training use this stored vector.

\subsubsection{Memory Encoder Architectures}
\label{app:memory_encoders}

\paragraph{Full-attention bottleneck.}
Separate projections map the consolidated video, recent visual, action, and proprioceptive streams to a shared memory width $d_m$. The two visual streams use linear layers, and actions and proprioception use two-layer MLPs. The projected tokens are concatenated along the token dimension. Modality is identified only through the separate projections. $B$ learned queries, modulated by the task language embedding through FiLM~\citep{perez2018film}, then pass through two pre-normalized Perceiver blocks~\citep{jaegle2021perceiver}. Each block applies cross-attention from the queries to the history tokens with padding masked, self-attention among the queries, and a residual $\tanh$-activated linear update. The output $m_t \in \mathbb{R}^{B \times d_m}$ is a fixed interface shared by all policy backbones. Table~\ref{tab:encoder_hparams} lists the hyperparameters.

\begin{table}[h]
\centering
\caption{Full-attention memory encoder hyperparameters.}
\label{tab:encoder_hparams}
\begin{tabular}{ll}
\toprule
Setting & Value \\
\midrule
Memory width $d$ & 1024 \\
Memory tokens $B$ & 256 (ablated over $\{1,4, 16, 64, 256, 1024\}$) \\
Visual stream projections & Linear \\
Action / proprioception projections & 2-layer MLP, GELU, hidden width 512 \\
Language conditioning & FiLM on queries, 2048-d instruction embedding \\
Perceiver blocks & 2, pre-LayerNorm \\
Attention heads & 8 (cross- and self-attention) \\
Residual update & $\tanh$-activated linear \\
\bottomrule
\end{tabular}
\end{table}
\paragraph{Recurrent bottleneck.}
Our streaming encoder uses gated-delta recurrence~\cite{yang2025gated} to
produce $B\times1024$ memory tokens. Each finalized video chunk contributes
four action--proprioception pairs followed by 12 pooled Wan tokens, forming
a sequence of 20 tokens. Each token receives learned modality and
within-chunk position embeddings, together with a projected normalized
chunk index.

The encoder contains two gated-delta blocks, each with eight heads and an
FP32 matrix state of size $128\times128$ per head. Each block applies
pre-normalization, a width-four causal depthwise convolution, and SiLU before
projecting queries, keys, values, write rates, and decay factors. At event
index $t$, head $h$ updates its state as
\begin{align}
 S_{t,\mathrm{ret}}^h &= \delta_t^h S_{t-1}^h,\\
 e_t^h &= v_t^h-S_{t,\mathrm{ret}}^h k_t^h,\\
 S_t^h &= S_{t,\mathrm{ret}}^h+
          \beta_t^h e_t^h(k_t^h)^\top,\qquad
 y_t^h=S_t^h q_t^h.
\end{align}
Here, $\beta_t^h$ is a sigmoid-bounded write rate and $\delta_t^h$ is a
positive learned decay factor; their projection biases are initialized to
$-1$ and $-4$, respectively. Padded events leave the state unchanged.

Language-conditioned learned queries read the recurrent states and combine
the resulting features with recent DINOv3 tokens and robot signals through
one full cross-attention/self-attention block. This bounded recent window
is processed separately from the recurrent state. During inference, we
cache the matrix states and the three preceding convolution inputs, update
each finalized chunk once, and reset the caches between episodes. 

\subsubsection{Memory Decoder and Pretraining Objectives}
\label{app:memory_objectives}

\paragraph{Decoder architecture.}
During Stage~1, a Perceiver-IO decoder reconstructs historical observations and robot signals from latent memory $m_t$. We uniformly sample up to 16 distinct historical steps $s\leq t$ without replacement, using all available steps for shorter prefixes. For each sampled step, we construct queries
\begin{equation}
q_{s,j}=e_{\mathrm{step}}(s)+e_{\mathrm{patch}}(j),\qquad
q_{s,\mathrm{sum}}=e_{\mathrm{step}}(s)+e_{\mathrm{sum}},
\end{equation}
where $e_{\mathrm{step}}(s)$ is a learned embedding of historical step $s$, $e_{\mathrm{patch}}(j)$ identifies a visual patch position across the three camera views, and $e_{\mathrm{sum}}$ is a learned summary embedding shared across steps. Each step has 192 patch queries that reconstruct DINOv3 features (64 patches per view) and one summary query that reconstructs its action and proprioceptive state. Sampled steps are sorted, and padded query groups are masked.

The decoder has two pre-normalized blocks with eight attention heads and
feedforward width 8192. In each block, the 193 queries for a step first
self-attend within that step, cross-attend to all memory tokens, and pass
through a feedforward network. Queries from different steps do not
self-attend to one another. Historical content reaches the decoder only
through $m_t$; it receives neither target features nor a separate language
input. Patch queries predict 768-dimensional DINOv3 features, while the
summary query predicts 16-dimensional action and proprioceptive vectors.
A separate learned global query attends to $m_t$ to predict progress in
$[0,1]$.

\paragraph{Multi-stream reconstruction.}
Following Sec.~\ref{sec:stage1}, we supervise reconstruction of visual features, actions, and proprioceptive states from $\mem_t$. Let $\mathcal{K}_t$ contain the $K$ valid historical steps sampled from the history prefix. The history-reconstruction component of Eq.~\ref{eq:stage1} is implemented as
\begin{equation}
\begin{aligned}
\loss_{\mathrm{rec}}=\frac{1}{K}\sum_{k\in\mathcal{K}_t}\Big[
&\frac{\lambda_v}{d_v}\big|\hat{z}^v_k-z^v_k\big|_2^2\
&+\frac{\lambda_a}{d_a}\big|\hat{\action}_k-\action_k\big|_2^2\
&+\frac{\lambda_q}{d_q}\big|\hat{\proprio}_k-\proprio_k\big|_2^2\Big],
\end{aligned}
\label{eq:rec_loss}
\end{equation}
where $\hat{z}^v_k, \hat{\action}_k,\hat{\proprio}_k$ denote decoder predictions for the sampled step $k$. The denominators count the elements in each target: $d_v=192\times768$ for visual features across all patches and views, and $d_a=d_q=16$ for actions and proprioceptive states. Before computing visual reconstruction error, we independently apply parameter-free LayerNorm to each predicted and target patch feature along its feature dimension. Visual targets remain frozen. Padded queries are excluded, and we use $(\lambda_v,\lambda_a,\lambda_q)=(1.0,0.5,0.5)$.

\paragraph{Progress estimation.}
During Stage~1 (Sec.~\ref{sec:stage1}), a learned global query $q_{\mathrm{prog}}$ cross-attends to all latent memory tokens $\mem_t$. The attention output is added to the query and passed through a learned linear head $W_\rho$ followed by a sigmoid:
\begin{equation}
\hat{\rho}_t = \sigma\!\left( W_\rho \left[ q_{\mathrm{prog}} + \operatorname{CrossAttn}\!\left(q_{\mathrm{prog}}, \mem_t\right) \right] \right) \in (0, 1).
\label{eq:progress_head}
\end{equation}
The progress prediction is therefore computed only from the compressed history in $\mem_t$. We supervise it with the normalized timestep as a coarse approximation of task progress. For a demonstration with $T$ steps indexed by $t = 0, \ldots, T-1$,
\begin{equation}
\rho_t = \frac{t}{T-1}, \qquad
\mathcal{L}_{\mathrm{prog}} = \left(\hat{\rho}_t - \rho_t\right)^2.
\label{eq:progress_loss}
\end{equation}
This target measures relative temporal position within the demonstration and requires no additional annotations. We weight this term by $\lambda_\rho = 0.5$ in Eq.~\ref{eq:stage1}. In preliminary experiments on the RoboCasa Wash and Return task, we also used reward labels from off-the-shelf reward models, RoboMeter and Robo-Dopamine, as progress targets. The normalized timestep led to higher success, likely because these labels were noisy on our demonstrations. We therefore use the normalized timestep as the progress target.

The Stage~1 objective is
\begin{equation}
\mathcal{L}_{\mathrm{mem}}
=
\mathcal{L}_{\mathrm{rec}}
+
\lambda_\rho \mathcal{L}_{\mathrm{prog}},
\end{equation}
where $\mathcal{L}_{\mathrm{rec}}$ is defined in
Eq.~\ref{eq:rec_loss}, with
$(\lambda_v,\lambda_a,\lambda_q,\lambda_\rho)
=(1.0,0.5,0.5,0.5)$.
The no-progress ablation sets $\lambda_\rho=0$ and retains all input streams.
Only the action and state tokenizer, memory encoder and decoder are trained in this stage; visual and
language features are frozen or precomputed, and training requires no VLA
policy. Stage~2 transfers the memory encoder and discards the decoder.
The first three terms supervise fine-grained reconstruction at individual past steps across all three streams, while the last term provides coarse task-progress supervision. Similar to masked video modeling~\citep{wang2023videomae}, reconstruction serves as a representation learning objective that prepares features for downstream adaptation. Our goal is to \emph{prime} the memory bottleneck with multi-stream experience and task progress, without requiring perfect history reconstruction. This gives joint policy fine-tuning informative historical features to build on, helping reduce reliance on spurious memory shortcuts.

\begin{wrapfigure}{r}{0.45\linewidth}
    \centering
    \vspace{-0.4cm}
   \includegraphics[width=\linewidth]{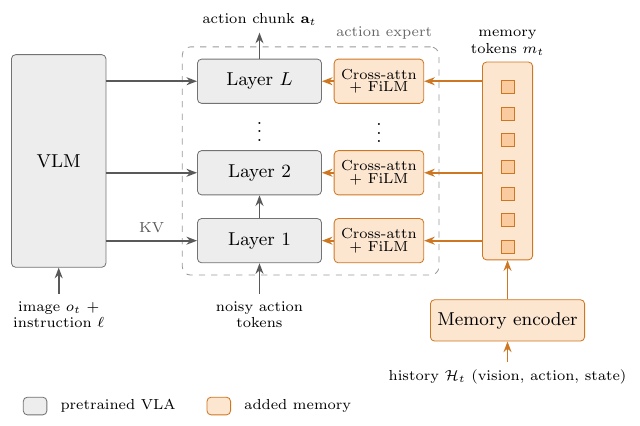}
    \caption{\textbf{Architecture of Memory Adapter for VLAs.}}
   \label{fig:memory_adapter}
    \vspace{-0.5cm}
\end{wrapfigure}

\subsubsection{Integration with Different VLA Backbones}
\label{app:vla_integration}

\paragraph{Memory adapter.}
Following~\citet{dai2026robomme}, we inject memory into the action expert rather than the VLM context. This choice outperforms VLM-context injection both in their study and in our RoboCasa experiments with heuristic memory (Fig.~\ref{fig:robocasa_results}). Memory tokens are therefore never appended to the VLM prefix, and the VLM context length and key-value cache are unchanged. We insert an adapter into each memory-enabled action-expert layer, between the layer's native attention update and its feedforward network (Fig.~\ref{fig:memory_adapter}). Let $X_\ell$ denote the action hidden states at layer $\ell$. The adapter first reads memory through cross-attention, with action tokens as queries and the memory tokens $\mem_t \in \mathbb{R}^{\memtok \times d}$ as keys and values. It then maps this read to feature-wise scale and shift parameters that modulate the normalized action features:
\begin{align}
R_\ell &= \operatorname{CrossAttn}\ell\big(\operatorname{RMSNorm}(X\ell), \operatorname{RMSNorm}(\mem_t)\big), \label{eq:app_memory_read}\\
(\gamma_\ell, \beta_\ell) &= W_\ell R_\ell, \qquad
\widetilde{X}\ell = \operatorname{RMSNorm}(X\ell) \odot (1 + \gamma_\ell) + \beta_\ell. \label{eq:app_film}
\end{align}
The modulated features then pass through the original pre-feedforward normalization, feedforward network, and residual connection. Each layer has its own adapter parameters, while all layers read the same memory tokens. We initialize $W_\ell$ with small random weights and zero bias. This introduces only a small perturbation to the pretrained policy while allowing gradients to reach the memory attention and encoder from the start of training. Because the adapter attends to a fixed number of memory tokens, its cost is independent of history length. The adapter supplements rather than replaces each backbone's native conditioning, as detailed below.

\paragraph{$\pi_0$ and $\pi_{0.5}$.}
We add the adapter to every action-expert layer, with hyperparameters listed in Table~\ref{tab:adapter_hparams}. In $\pi_0$, the state token and the timestep embedding fused into each noisy-action token are unchanged. In $\pi_{0.5}$, the timestep-conditioned AdaRMSNorm remains active after memory modulation.

\paragraph{MolmoAct2.}
We inject memory only into the flow-matching action expert. Each memory-enabled layer applies action self-attention, VLM/state cross-attention, memory cross-attention and FiLM (Eqs.~\ref{eq:app_memory_read}--\ref{eq:app_film}), and the action feedforward network, in that order. MolmoAct2's native timestep-dependent scale, shift, and residual gates are applied after memory modulation, and its state embeddings remain in the native VLM/state key-value context.

\paragraph{Policy adaptation.}
Because all three backbones consume the same token interface, a single Stage~1 memory checkpoint initializes all of them, and only the action-expert adapter is backbone-specific. By default, we fine-tune memory and policy jointly with each backbone's native action objective, keeping the vision encoder frozen. For all models, we train the memory encoder, adapter, action expert, and VLM transformer. Table~\ref{tab:ablation_definitions} summarizes alternative training configurations, and Sec.~\ref{app:ablation_configs} describes additional freezing choices.

\begin{table}[h]
\centering
\caption{Memory adapter hyperparameters for $\pi_0$ and $\pi_{0.5}$.}
\label{tab:adapter_hparams}
\begin{tabular}{ll}
\toprule
Setting & Value \\
\midrule
Memory / action-expert width & 1024 \\
Query heads & 4 \\
Key-value heads & 1 (grouped-query attention) \\
Head dimension & 256 \\
Positional encoding & Separate rotary positions for memory and action tokens \\
FiLM projection initialization & $\mathcal{N}(0, 0.002^2)$, zero bias \\
\bottomrule
\end{tabular}
\end{table}

\subsubsection{Training Hyperparameters and Resources}
\label{app:training}

Table~\ref{tab:appendix_training_hparams} lists the RMBench optimization
settings. Training uses four B200 or four H200 GPUs per run. Stage~2 uses each backbone's native flow-matching objective. For $\pi_0$ and $\pi_{0.5}$, we
sample $\epsilon\sim\mathcal{N}(0,I)$ and
$\tau\sim\operatorname{Beta}(1.5,1)$, form
$x_\tau=\tau\epsilon+(1-\tau)a$, and regress the velocity
$u_\tau=\epsilon-a$. MolmoAct2 uses the equivalent reverse-time convention.
The standard recipe uses action supervision alone.

\begin{table}[t]
\centering
\small
\caption{Optimization settings for \textsc{IMPRINT} on RMBench.}
\label{tab:appendix_training_hparams}
\begin{tabularx}{\linewidth}{@{}lXX@{}}
\toprule
Setting & Stage~1: memory pretraining & Stage~2: joint adaptation\\
\midrule
Updates & 160,000 & 100,000\\
Global batch size & 64 & 64\\
GPUs & 4 & 4\\
Optimizer & AdamW & AdamW\\
Peak learning rate & $1.0\times10^{-4}$ & $2.5\times10^{-5}$\\
Final learning rate & $1.0\times10^{-5}$ & $2.5\times10^{-6}$\\
Warm-up updates & 1,000 & 500\\
Schedule & cosine & cosine\\
Weight decay & $1.0\times10^{-4}$ & $1.0\times10^{-10}$\\
Adam $(\beta_1,\beta_2)$ &  
  (0.9,0.999)
   & $(0.9,0.95)$\\
Gradient clipping & 1.0 & 1.0\\
EMA decay & -- & 0.999\\
Training seed & 0 & 42\\
Checkpoint interval & 10,000 & 10,000\\
Action horizon & -- & 50\\
Precision & mixed BF16/FP32 & BF16\\
\bottomrule
\end{tabularx}
\end{table}

\subsection{Baseline Implementations}
\label{app:baselines}

We compare against three categories of memory-augmented policies, whose memory designs are summarized in Table~\ref{tab:memory_design_comparison}. All six store only visual history, and none stores action or proprioceptive history. 

Heuristic methods retain raw frames chosen by a fixed rule. Uniform sampling~\citep{dai2026robomme} uses only visual history, in the form of frames sampled uniformly over the episode. EventVLA~\citep{yang2026eventvla} also uses only visual history: on RMBench, it concatenates the initial frame and a window of recent frames into the VLM input.

Hierarchical methods pair a high-level VLM with a low-level policy. MemER~\citep{sridhar2026scaling} uses visual history and text. Its high-level VLM takes the recent frames and previously selected keyframes as input to select new keyframes, and passes the next subtask to the low-level policy as text. Mem-0~\citep{chen2026rmbench} likewise uses visual history and text. Its planner reads the initial frame and the text description and final frame of each completed subtask, and its executor attends to image features of the frame at the start of the current subtask and of recent frames. 

Latent methods learn continuous memory representations. HAMLET~\citep{koo2026hamlet} uses only visual history, encoding each past observation into moment tokens initialized by time-contrastive learning and aggregating them over a fixed window of past steps. MemoryVLA~\citep{shi2026memoryvla} also uses only visual history, storing perceptual tokens and a VLM cognitive token for each observation in a memory bank and retrieving from it at each step.

None of these methods uses action or proprioceptive history, both of which \ours encodes alongside visual history.

\paragraph{Source of baseline results.}
Table~\ref{tab:baseline_implementation} lists the backbone of each method and
the source of its RMBench result. For Mem-0, MemER, EventVLA, and MemoryVLA,
we report the RMBench success rates published in prior work, which follow
the same protocol of 50 demonstrations for training and 100 evaluation
rollouts per task. The Mem-0 result is reported by~\citet{chen2026rmbench},
and the MemER, EventVLA, and MemoryVLA results by~\citet{yang2026eventvla}.
For EventVLA, this is the visual-anchor configuration that its authors
evaluate on RMBench, which retains only the initial frame and a few recent
frames without its keyframe evidence memory, so it does not reflect the full
EventVLA. For MemoryVLA, it is the QwenOFT variant. We train
the non-memory VLAs, uniform sampling, and HAMLET ourselves on RMBench, and
train all baselines ourselves on RoboCasa and in the real world. Because the
reported results come from different backbones and training pipelines, they
are not controlled comparisons. Controlled comparisons are provided by the
baselines we train and by the ablations in Sec.~\ref{sec:exp_encoding}
and Sec.~\ref{sec:exp_learning}.

\begin{table}[t]
\centering
\small
\caption{Compared policies, their backbones, and the source of their
RMBench results.}
\label{tab:baseline_implementation}
\begin{tabularx}{\linewidth}{@{}llXl@{}}
\toprule
Method & Category & Backbone & RMBench result\\
\midrule
non-memory VLA & None & $\pi_0$, $\pi_{0.5}$, MolmoAct2 & Trained by us\\
Uniform sampling & Heuristic & $\pi_{0.5}$ & Trained by us\\
EventVLA (visual anchors only)$^\dagger$ & Heuristic & QwenOFT (Qwen3-VL-4B) & \citet{yang2026eventvla}\\
MemER & Hierarchical & VLM planner + VLA policy & \citet{yang2026eventvla}\\
Mem-0 & Hierarchical & Qwen3-VL-8B planner + DiT policy & \citet{chen2026rmbench}\\
HAMLET & Latent & GR00T N1.6 & Trained by us\\
MemoryVLA (QwenOFT)$^\dagger$ & Latent & QwenOFT & \citet{yang2026eventvla}\\
\midrule
\textbf{\textsc{IMPRINT}} & Latent & $\pi_0$, $\pi_{0.5}$, MolmoAct2 & Trained by us\\
\bottomrule
\end{tabularx}
\end{table}

\begin{table*}[t]
\centering
\small
\setlength{\tabcolsep}{2.6pt}
\renewcommand{\arraystretch}{1.12}
\newcommand{\grp}[1]{\multicolumn{8}{@{}l}{\textit{#1}}}
\caption{Comparison of memory designs. \cmark: yes; \pmark: partially;
\xmark: no. $^\dagger$On RMBench, EventVLA is evaluated only in its
visual-anchor configuration, which keeps the initial frame and a few recent
frames and omits its keyframe evidence memory~\citep{yang2026eventvla}. The
reported result therefore reflects this heuristic variant rather than the
full EventVLA, and the column describes this evaluated variant.}
\label{tab:memory_design_comparison}
\begin{tabular}{@{}l cc cc cc >{\columncolor{oursbg}}c@{}}
\toprule
& \multicolumn{2}{c}{Heuristic} & \multicolumn{2}{c}{Hierarchical} & \multicolumn{2}{c}{Latent} & \\
\cmidrule(lr){2-3}\cmidrule(lr){4-5}\cmidrule(lr){6-7}
& \shortstack{Uniform\\sampling} & EventVLA$^\dagger$ & MemER & Mem-0 & HAMLET & MemoryVLA & \textbf{\textsc{IMPRINT}}\\
\midrule
\grp{What memory stores}\\
Visual history          & \cmark & \cmark & \cmark & \cmark & \cmark & \cmark & \cmark\\
Action history          & \xmark & \xmark & \xmark & \xmark & \xmark & \xmark & \cmark\\
Proprio.\ history & \xmark & \xmark & \xmark & \xmark & \xmark & \xmark & \cmark\\
Memory form             & Frames & Frames & \begin{tabular}[c]{@{}c@{}}Frames,\\text\end{tabular} & \begin{tabular}[c]{@{}c@{}}Frames,\\text\end{tabular} & Latent & Latent & Latent\\
\midrule
\grp{How memory is supervised}\\
Action loss             & \cmark & \cmark & \xmark & \cmark & \cmark & \cmark & \cmark\\
Visual objective        & \xmark & \xmark & \xmark & \xmark & \cmark & \xmark & \cmark\\
Action/proprio.\ objective & \xmark & \xmark & \xmark & \xmark & \xmark & \xmark & \cmark\\
Subtask/keyframe labels & \xmark & \xmark & \cmark & \cmark & \xmark & \xmark & \xmark\\
Task progress           & \xmark & \xmark & \xmark & \xmark & \xmark & \xmark & \cmark\\
\midrule
\grp{Training and inference}\\
No annotations needed   & \cmark & \cmark & \xmark & \xmark & \cmark & \cmark & \cmark\\
Trained without the policy & \xmark & \xmark & \cmark & \cmark & \cmark & \xmark & \cmark\\
No high-level VLM & \cmark & \cmark & \xmark & \xmark & \cmark & \cmark & \cmark\\
Covers the full history & \xmark & \xmark & \xmark & \xmark & \xmark & \cmark & \cmark\\

Shown on multiple VLAs  & \xmark & \xmark & \xmark & \xmark & \cmark & \xmark & \cmark\\
\bottomrule
\end{tabular}
\end{table*}

\subsection{Ablation Configurations}
\label{app:ablation_configs}

Table~\ref{tab:ablation_definitions} defines every variant used in our
ablations and indicates where each is reported. The main paper studies four
aspects of \textsc{IMPRINT}: memory encoding (Sec.~\ref{sec:exp_encoding}),
the training recipe and pretraining objectives
(Sec.~\ref{sec:exp_learning}), the choice of VLA backbone, and where memory
is injected into the policy. Each ablation changes a single factor relative
to the default configuration described in Sec.~\ref{app:imprint_implementation}.
Unless stated otherwise, the demonstrations, tokenizers, action horizon,
optimizer, training budget, and evaluation protocol are held fixed, and all
variants are evaluated with the same number of trials per task as the main
results.

\paragraph{Memory encoding.}
The capacity ablation (Fig.~\ref{fig:memory_capacity}) varies only the
number of memory tokens, $B\in\{1, 4,16,64,256, 1024\}$, while holding the memory
width, tokenizers, encoder depth, pretraining objectives, and policy adapter
fixed. It uses the full-attention encoder and is evaluated on RMBench and on
the memory and non-memory task subsets of RoboCasa. The encoder
ablation (Sec.~\ref{sec:exp_encoding}) keeps $B=256$ and replaces the
recurrent gated-delta encoder with the full-attention Perceiver encoder
described in Sec.~\ref{app:memory_encoders}. Both encoders are evaluated on
RMBench and RoboCasa, and their latency is measured with the protocol in
Sec.~\ref{app:timing_protocol}.

\paragraph{Training recipe.}
The recipe ablation (Fig.~\ref{fig:training_ablation}, left) compares three
ways of training the memory encoder. \emph{End-to-end} initializes the
encoder randomly and trains it only through the action loss in
Eq.~\ref{eq:action_objective}. \emph{Frozen memory} initializes the
encoder from Stage~1 and keeps it fixed during Stage~2, so only the adapter
and policy are trained. \emph{\textsc{IMPRINT}} initializes the encoder from
Stage~1 and updates it jointly with the adapter and policy. All three
variants are evaluated on RMBench, and the same checkpoints are used for the
representation analysis in Sec.~\ref{app:representation_analysis}.

\paragraph{Pretraining objectives.}
The objective ablation (Fig.~\ref{fig:training_ablation}, middle) varies the
Stage~1 loss weights in Eq.~\ref{eq:stage1} while keeping Stage~2
adaptation, the decoder, and the sampling of up to 16 historical steps
fixed. \emph{Without progress} sets $\lambda_\rho=0$. \emph{Visual only}
keeps only visual reconstruction, and \emph{action and state only} keeps only
action and proprioceptive reconstruction. These variants remove supervision
but not inputs, so every variant still encodes all three streams. All
variants are evaluated on RMBench.

\paragraph{Backbones and memory injection.}
The backbone study (Fig.~\ref{fig:flexible_vla}) applies the same memory
pipeline and adaptation procedure to $\pi_0$, $\pi_{0.5}$, and MolmoAct2,
changing only the backbone-specific adapter described in
Sec.~\ref{app:vla_integration}. Each backbone is compared with its
non-memory counterpart trained with the same Stage~2 budget on RMBench. The
injection study (Fig.~\ref{fig:robocasa_results}) supplies the same uniformly
sampled frames either to the action expert through the memory adapter or to
the VLM context as additional image tokens, and is evaluated on RoboCasa.

\begin{table*}[t]
\centering
\small
\setlength{\tabcolsep}{4pt}
\caption{Configurations used in the ablation study. \emph{Stage~1 init.}:
whether the memory encoder starts from Stage~1 pretraining. \emph{Memory
updated}: whether the encoder is optimized during Stage~2. Unlisted
settings match the default \textsc{IMPRINT} configuration.}
\label{tab:ablation_definitions}
\begin{tabularx}{\textwidth}{@{}llcc>{\raggedright\arraybackslash}Xl@{}}
\toprule
Study & Variant & \shortstack{Stage~1\\init.} & \shortstack{Memory\\updated} & Change from default & Reported in\\
\midrule
Encoding
 & Memory capacity & yes & yes & $B\in\{1,4,16,64,256, 1024\}$ (full-attention encoder) & Fig.~\ref{fig:memory_capacity}\\
 & Full attention & yes & yes & Perceiver attention instead of recurrence & Fig.~\ref{fig:inference_cost}, Sec.~\ref{sec:exp_encoding}\\
\midrule
Recipe
 & End-to-end & no & yes & Random encoder, action loss only & Fig.~\ref{fig:training_ablation}\\
 & Frozen memory & yes & no & Encoder fixed during Stage~2 & Fig.~\ref{fig:training_ablation}\\
 & \textsc{IMPRINT} & yes & yes & None (default) & Fig.~\ref{fig:training_ablation}\\
\midrule
Objectives
 & Without progress & yes & yes & $\lambda_\rho=0$ & Fig.~\ref{fig:training_ablation}\\
 & Visual only & yes & yes & $\lambda_a=\lambda_q=\lambda_\rho=0$ & Fig.~\ref{fig:training_ablation}\\
 & Action and state only & yes & yes & $\lambda_v=\lambda_\rho=0$ & Fig.~\ref{fig:training_ablation}\\
\midrule
Integration
 & Backbone & yes & yes & $\pi_0$, $\pi_{0.5}$, or MolmoAct2 & Fig.~\ref{fig:flexible_vla}\\
 & Injection location & -- & -- & Sampled frames in action expert or VLM context & Fig.~\ref{fig:robocasa_results}\\
\midrule
Inputs & Vision only      & yes & yes & Action and state inputs and targets removed & Table~\ref{tab:input_stream_ablation} \\
       & Vision + Action  & yes & yes & State input and target removed             & Table~\ref{tab:input_stream_ablation} \\
       & Vision + State   & yes & yes & Action input and target removed            & Table~\ref{tab:input_stream_ablation} \\
\bottomrule
\end{tabularx}
\end{table*}

\section{Benchmarks and Evaluation Protocols}
\label{app:benchmarks}
\label{app:experimental_protocols}

We evaluate memory-based manipulation tasks on RMBench, memory and
non-memory tasks on a customized RoboCasa benchmark, and three real-world
tasks. The following sections describe the tasks, datasets, and evaluation
procedures.

\subsection{RMBench}
\label{app:rmbench_tasks}

RMBench comprises nine manipulation tasks built on RoboTwin 2.0 with the
Aloha-AgileX embodiment. It distinguishes tasks requiring recall of one event,
M(1), from those requiring a sequence of events, M($n$).
We use the demo clean setting, original task instructions, and
50 expert demonstrations per task, for 450 demonstrations in total.
Table~\ref{tab:rmbench_task_details} summarizes the tasks and episode limits.

\begin{table*}[t]
\centering
\small
\caption{RMBench tasks and episode limits, measured in simulator steps.}
\label{tab:rmbench_task_details}
\begin{tabularx}{\textwidth}{@{}lcXr@{}}
\toprule
Task & Type & Task and memory requirement & Limit\\
\midrule
Observe and pickup & M(1) & Pick the object matching a target shown before occlusion. & 250\\
Rearrange blocks & M(1) & Move the center block to the empty mat, press, and move the initially mat-placed block to the center. & 700\\
Put back block & M(1) & Move the block to the center, press, and return it to its original mat. & 500\\
Swap blocks & M(1) & Use the spare tray to exchange the blocks' initial tray assignments, then press once. & 1000\\
Swap T & M(1) & Exchange the initial positions and orientations of two T-shaped blocks. & 600\\
\midrule
Battery & M($n$) & Use gauge feedback from tested orientation pairs to place both batteries correctly. & 1000\\
Block ranking & M($n$) & Use feedback from failed arrangements to find the hidden left-to-right block order. & 3500\\
Cover blocks & M($n$) & Remember the initial color order, cover from left to right, and uncover in red--green--blue order. & 1500\\
Press button & M($n$) & Recall two displayed numbers, press the left and middle buttons those numbers of times, and confirm. & 1500\\
\bottomrule
\end{tabularx}
\end{table*}

Episode seeds determine object identities, poses, colors, and hidden task
variables. Success follows each environment's state-based criteria.
For example, Cover Blocks requires the complete cover/uncover sequence and
rejects out-of-order transitions.

\subsection{Customized RoboCasa Benchmark}
\label{app:robocasa_tasks}

Our customized RoboCasa benchmark contains 12 tasks, six memory-based
and six non-memory (Table~\ref{tab:robocasa_task_groups}). Four of the
memory tasks follow ReMemBench~\citep{shah2026scaling} and require recalling an object's
location or its association with a container: Retrieve Fruit, Retrieve Oil,
Wash and Return, and Wash and Return to Same Location. The other two, Sweeten
Hot Chocolate and Rinse Cutting Board, require tracking task progress. In
each memory task, the policy must retain a variable that the
current observation no longer reveals, such as an earlier object location, a
source container, an exact initial position, the number of sugar cubes already
deposited, or the accumulated rinse duration. The non-memory tasks can be
specified from the current observation and instruction alone, although they
may still require multi-step closed-loop manipulation.

Each task has 50 expert demonstrations, for 600 demonstrations in total. For
evaluation, we restore 50 recorded initial states per task, each consisting
of the simulator XML, state, and episode metadata, rather than sampling new
scenes procedurally. These 50 configurations do not correspond to 50 distinct
kitchens. Some tasks use a single layout and style and vary objects,
placements, and task parameters, whereas others span up to 34 layouts and 35
styles. Policies observe three synchronized $256\times256$ RGB views
as left agentview, right agentview, and
wrist view. An episode ends upon success or after a task-specific
horizon of policy steps. Retrieve Fruit and Retrieve Oil begin with a
scripted exploration phase where it moves around left to right to view the full scene before the policy takes control, so their total simulator
duration can exceed the listed horizon.

\begin{table*}[t]
\centering
\scriptsize
\setlength{\tabcolsep}{3pt}
\renewcommand{\arraystretch}{1.12}
\caption{Tasks in the customized RoboCasa benchmark. \emph{Horizon}: maximum
number of policy-controlled simulator steps. \emph{L/S}: number of distinct
layouts and styles among the 50 evaluation configurations. Instructions are
abbreviated.}
\label{tab:robocasa_task_groups}
\begin{tabularx}{\textwidth}{@{}>{\raggedright\arraybackslash}p{2.6cm}>{\raggedright\arraybackslash}Xrc>{\raggedright\arraybackslash}p{4.2cm}@{}}
\toprule
Task & Instruction and memory requirement & Horizon & L/S & Variation across configurations\\
\midrule
\multicolumn{5}{@{}l}{\textit{Memory}}\\
Retrieve fruit & Place the \{fruit\} in the sink; recall which side of the sink it was observed on. & 1600 & 1/1 & 9 fruit categories; target at the far left or right of the sink.\\
Retrieve oil & Pick up the olive-oil bottle; recall which of two bottles is the target after both sides are shown. & 1600 & 1/1 & Four initial/target-side combinations (LL, LR, RL, RR).\\
Wash and return & Wash the fruit and return it to its original container; recall the source container. & 1600 & 1/1 & Left or right source container; fruit, receptacle, and placements vary.\\
Wash and return to same location & Wash the fruit and return it to its exact initial location; recall that location. & 1600 & 1/1 & Initial fruit location varies.\\
Sweeten hot chocolate & Place \{1, 2, 3\} sugar cubes in the saucepan; track how many have been placed. & 2100 & 15/17 & Target counts 1/2/3 in 20/20/10 configurations; 3--5 cubes per scene, including 0--3 extras.\\
Rinse cutting board & Rinse the board with hot water for ten seconds, then turn off the faucet; track rinse duration. & 2250 & 34/18 & Board instance, pose, sink, and kitchen vary.\\
\midrule
\multicolumn{5}{@{}l}{\textit{Non-memory}}\\
Pan transfer & Dump the vegetables from the pan onto the plate and return the pan to the stove. & 1800 & 1/1 & Pan, food, plate, and placements vary.\\
Organize condiments & Move only the condiments or shakers from the counter into the cabinet. & 2250 & 33/35 & Three targets and one distractor; 8 target and 36 distractor categories.\\
Counter to sink & Pick the \{object\} from the counter and place it in the sink. & 600 & 28/29 & 29 target categories, 43 target assets, and two distractors per scene.\\
Place equal ice cubes & Place two of the four ice cubes into each glass of water. & 2850 & 25/34 & Four cubes, two glasses, and one bowl, drawn from 24 assets.\\
Scrub cutting board & Scrub the board with the sponge, then release it. & 1200 & 30/32 & Six cutting-board and four sponge assets.\\
Frying-pan adjustment & Move the pan to another burner and turn that burner on. & 1200 & 31/30 & Ten pan assets; source and destination burners vary.\\
\bottomrule
\end{tabularx}
\end{table*}

\subsection{Real-World Setup and Tasks}
\label{app:realworld_tasks}

We use a bimanual YAM robot with one head camera and two wrist cameras.
The robot is controlled in joint space with 14-dimensional action and state
vectors, padded to 32 dimensions for the $\pi_0$ interface. Real-world
policies predict chunks of 30 actions and execute all of them for all methods. The visualization of those tasks are shown in Fig.~\ref{fig:teaser_figure} and Fig.~\ref{fig:real_world_tasks}.

\begin{figure}
    \centering
    \includegraphics[width=\linewidth]{figs/pi0_nonmemory_success_initial_views.pdf}
    \caption{\textbf{Real World Tasks with Three Camera Views}}
    \label{fig:real_world_tasks}
\end{figure}
\paragraph{Dust Table.}
The task instruction is ``move the objects away to dust the table and put them
back in the original positions.'' The robot must remember the objects'
original locations while clearing and cleaning the table. It must also
remember whether the table has already been dusted, to avoid dusting it
repeatedly. The original locations are recoverable from visual history, while
the dusting progress is reflected in the robot's past actions and
proprioceptive states as well as visual histories. The objects are placed on the table in two sets of
bottles. For each set, the left and right positions of the bottles are
swapped with probability 0.5, and the duster is placed on the left or right
side of the table with equal probability.

\paragraph{Scan Bottle.}
The instruction is ``scan the bottles and put them on the other side.''
The robot scans and transfers the bottles while tracking which have already
been scanned. The bottle locations are recoverable from visual history, while
the scanning status and how long it has held still is reflected in the robot's past actions and
proprioceptive states. The bottles start on either the left or the right side
of the table, and we randomize which arm performs the scan and which bottle
is scanned first.

\begin{wrapfigure}{r}{0.6\linewidth}
    \centering
    \vspace{-0.6cm}\includegraphics[width=\linewidth]{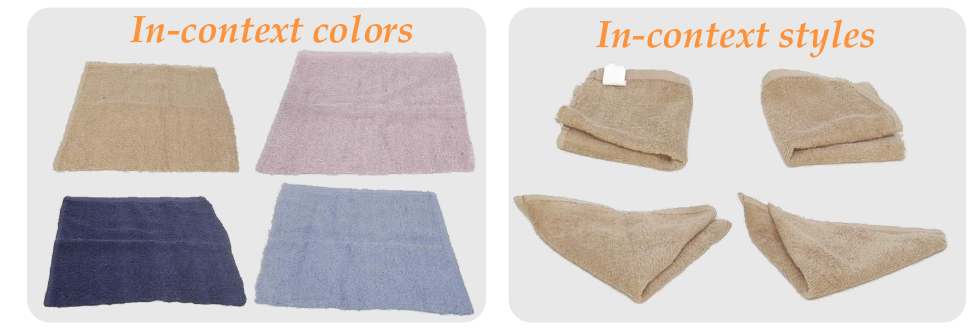}
    \caption{\textbf{Real World Fold Cloth Task Configuration}}
    \vspace{-0.6cm}\label{fig:cloth_styles}
\end{wrapfigure}

\paragraph{Fold Cloth.}
The instruction is ``follow the example to fold the towel.'' The robot must
retain the demonstrated folding configuration and the motion to fold the cloth while executing the fold. The
towels come in four colors, and each can be folded in one of four styles as shown in Fig.~\ref{fig:cloth_styles}. For
each episode, one robot demonstration with the same color and style is
provided as an in-context example and concatenated to the memory. Imitating
the example requires its actions and proprioceptive states to reproduce the folding
motion.

\subsection{Demonstration Datasets}
\label{app:datasets}

We use 50 demonstrations per task for 9 tasks in Rmbench, 50 demonstrations per task for 12 Robocasa tasks and 100 demonstrations per task for 3 real world tasks. Rmbench demonstrations are collected with scripted policies. Robocasa uses human teleoperated demos with spacemouse from Robocasa and Remembench. Real world task uses leader-follower arms in bimanual Yam platforma to collect demonstrations with kinesthetic teaching. We randomize the object locations and the memory condition for each demonstration in the real world tasks. The representation analyses
in Sec.~\ref{app:representation_analysis} use training demonstrations to
characterize the learned memory.

\subsection{Rollout Protocol and Success Metrics}
\label{app:evaluation}

We evaluate policies on two simulation benchmarks, RMBench and RoboCasa, and on three real-world tasks. For each setting, we first describe the rollout protocol and then the task success criteria.

\subsubsection{RMBench}

We follow the evaluation protocol of~\citep{chen2026rmbench}, running 100 accepted trials per task starting from seed 100000. Each task runs on two L40S GPUs: one serves the policy, and the other handles simulation and rendering. The policy deterministically predicts and executes chunks of 50 actions, using a causal memory buffer that holds finalized history chunks and a recent observation window. Success criteria for all 9 tasks follow~\citep{chen2026rmbench}.

\subsubsection{RoboCasa}

\paragraph{Protocol.}
We run 50 rollouts for each of the 12 tasks, one per fixed, demonstration-derived scene configuration. Each rollout restores the corresponding simulator state, object placements, instruction, and camera configuration. The $\pi_{0.5}$ policy deterministically predicts 20-action chunks and executes five actions before replanning. All tasks use the same three-view observation pipeline as in training, consolidate finalized history chunks, and retain the recent-observation window.

Tasks differ in how the episode is initialized. For retrieval tasks, we reconstruct the recorded exploration prefix by replaying simulator states and inserting the associated 12-dimensional actions into memory with the same causal alignment used in training; these actions populate memory but are not re-executed. Wash-and-return tasks have no scripted exploration and start directly from the restored reset state, and all remaining tasks start from their demonstration-aligned reset.

\paragraph{Success criteria.}
Success is determined by task-specific predicates on the simulator state. Throughout, we say an object is \emph{released} when the gripper finger gap exceeds $0.075$ and the end-effector--object distance exceeds $0.10$\,m.

\textit{Memory tasks.}
\begin{itemize}
    \item \textbf{Retrieve Fruit.} The policy must remember whether the instructed fruit is on the left or right and place it in the sink. Success requires the target fruit to be fully inside the sink basin and released.
    \item \textbf{Retrieve Oil.} The policy must remember which oil bottle is the target and retrieve it from the counter. Success requires the target bottle to be lifted more than $0.08$\,m above its initial height.
    \item \textbf{Wash and Return Left/Right.} The policy must place the fruit in the sink and then return it to its remembered source container. The evaluator first requires a full sink-entry event (with $0.03$\,m of vertical padding on the sink region). Afterwards, the fruit's $x$ and $y$ coordinates must each lie within $0.08$\,m of the source-container center, and the fruit must be released.
    \item \textbf{Wash and Return to the Same Location.} Same as above, except the fruit must be returned to its original counter position: after sink entry, $|x - x_0| < 0.05$\,m and $|y - y_0| < 0.05$\,m, and the fruit must be released. Success is latched once all conditions are met.
    \item \textbf{Sweeten Hot Chocolate.} The policy must place the instructed number $g \in \{1,2,3\}$ of sugar cubes into the saucepan and then return the arm toward its reset pose. Any cube, including distractors, counts toward $g$. A placement event is recorded when a cube stays inside the saucepan, ungrasped, for 10 consecutive frames. Success is computed from the stored trajectory and requires exactly $g$ placement events, no cube grasped at the end, maximum mobile-base joint drift of at most $2{\times}10^{-4}$, and at least one post-placement end-effector position within $0.15$\,m of its initial position.
    \item \textbf{Rinse Cutting Board.} The policy must hold the cutting board under running water and then close the faucet. The rinse event is latched once the board stays under water, with the faucet on and water temperature at least $0.6$, for more than 100 consecutive evaluator steps. Afterwards, the faucet must be off, with its handle within $0.2$\,rad of the closed position.
\end{itemize}

\textit{Non-memory tasks.}
\begin{itemize}
    \item \textbf{Pan Transfer.} The policy must move the vegetable from the pan to the plate while leaving the pan on its original burner. Success requires the vegetable to be inside the plate, and the pan to be released, in contact with the stove, centered within $0.08$\,m of the recorded burner, and tilted less than $15^\circ$ from upright. Any gripper--vegetable contact during the rollout invalidates success.
    \item \textbf{Organize Condiments.} The policy must move three condiments into the cabinet while leaving a distractor on the counter. Success requires all three targets inside the cabinet, the distractor in contact with the counter, and the end effector more than $0.25$\,m from all four objects. The cabinet-door state is not checked.
    \item \textbf{Pick and Place from Counter to Sink.} Success requires the target object's center to lie inside the sink and the end effector to be more than $0.25$\,m from the object.
    \item \textbf{Place Equal Ice Cubes.} The policy must distribute four ice cubes so that each of two glasses contains exactly two. A cube counts as inside a glass if it is detected through physical contact within the native $0.5$\,m receptacle threshold, or if it touches a cube detected this way; each glass must contain at least one directly detected cube. At completion, the end effector must be more than $0.15$\,m from every cube.
    \item \textbf{Scrub Cutting Board.} The policy must grasp the sponge, scrub the board over a sufficiently large area, and release the sponge. While the grasped sponge contacts the board, the evaluator must record at least five distinct contact positions, each at least $0.02$\,m from all earlier ones, spanning a planar range of at least $0.10$\,m. The final end-effector--sponge distance must exceed $0.15$\,m.
    \item \textbf{Frying-Pan Adjustment.} The policy must move the pan to a different burner and turn that burner on. Success requires the pan to contact the stove with its center within $0.08$\,m of a burner other than the starting one, and the destination burner's knob to be on. No release condition is imposed.
\end{itemize}

\subsubsection{Real World}
\label{app:real_world_eval}
\paragraph{Protocol.}
Table~\ref{tab:real_world_protocol} summarizes the rollouts for each task. Before every rollout, the operator manually resets the workspace and randomizes object placement. The $\pi_0$ policy predicts 30-step, 14-dimensional joint-and-gripper action chunks. Real-time chunking is disabled, and each full chunk is executed synchronously at a nominal 30\,Hz before replanning, corresponding to roughly one second of motion plus policy-refresh latency. For Dust Table and Scan Bottle, episodic memory starts empty and is populated online with the observations, robot states, and actions actually issued to the robot. For Fold Cloth, memory is initialized with a full demonstration as an in-context example, which is encoded as context but not physically replayed.

\begin{table}[h]
\centering
\caption{Real-world rollout protocol.}
\label{tab:real_world_protocol}
\begin{tabular}{lcl}
\toprule
Task & Rollouts & Variation \\
\midrule
Dust Table  & 20 & 2 object sets $\times$ duster left/right (5 rollouts each) \\
Scan Bottle & 20 & bottles on left/right (10 rollouts each) \\
Fold Cloth  & 32 & 16 towel color/style contexts (2 rollouts each) \\
\bottomrule
\end{tabular}
\end{table}

\paragraph{Success criteria.}
\begin{itemize}
    \item \textbf{Dust Table.} The robot must place the objects down upright, dust the table exactly once, and return the objects upright to the small table in the correct left/right arrangement.
    \item \textbf{Scan Bottle.} The robot must pick up the bottles one at a time, scan each exactly once, and place them on the other side in an aligned vertical row without collisions.
    \item \textbf{Fold Cloth.} The robot must reproduce the folding steps of the demonstration and reach a final shape close to the demonstrated one. Repeated grasp attempts are allowed.
\end{itemize}

\begin{figure}
    \centering
    \includegraphics[width=\linewidth]{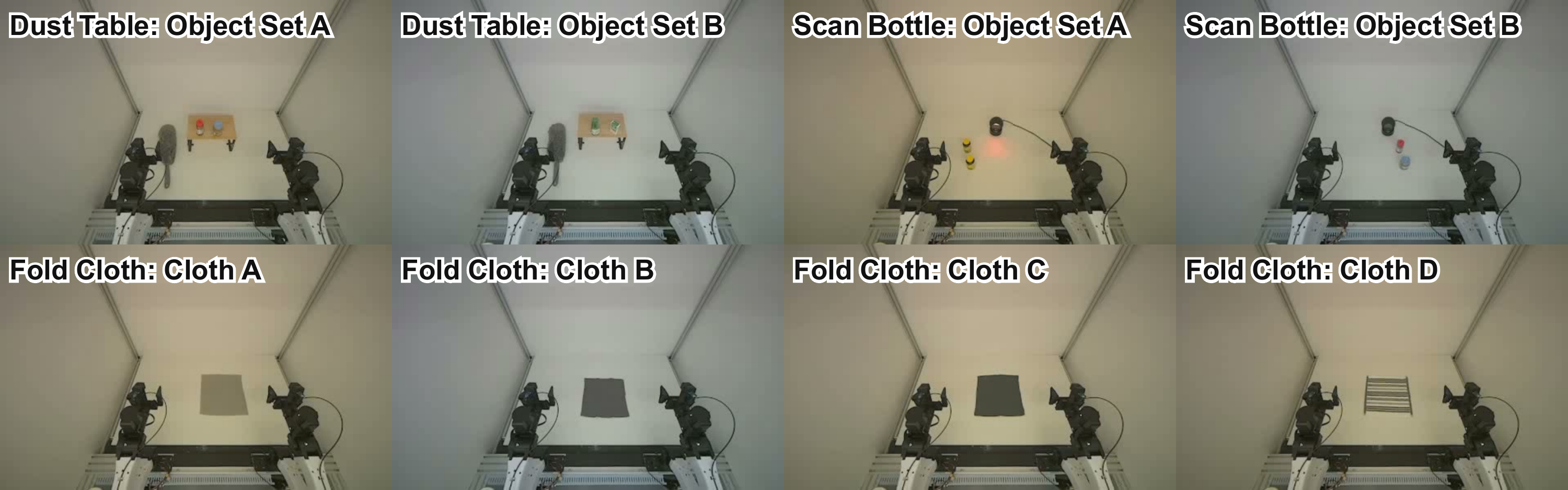}
    \caption{\textbf{OOD Scenarios for Each Task in the Real World}}
    \label{fig:ood_scenes}
\end{figure}

\paragraph{OOD Evaluation.}
To test whether the memory overfits to the objects and scenes seen in the
demonstrations, we also evaluate on out-of-distribution objects. A good memory
objective should extract the information needed for the task while still
generalizing to new objects. For Dust Table, we introduce two new sets of
objects whose shapes differ from the in-distribution objects. For Scan
Bottle, we likewise introduce two new sets of objects. For Fold Cloth, we use
four new towels in colors not seen during training
(Fig.~\ref{fig:ood_scenes}). All OOD evaluations follow the same protocol and
number of rollouts as the in-distribution evaluation.

\subsection{Statistical Reporting}
\label{app:statistics}

We follow the statistical reporting protocol of \citep{lbmtri2025}. Each rollout is treated as an independent Bernoulli trial, so estimating a policy's success rate on a task amounts to estimating the parameter $p$ of the Bernoulli distribution that generates its success/failure outcomes. In simulation, initial conditions are set by fixed random seeds; in the real world, the operator resets and randomizes the scene before every rollout.

\paragraph{Violin plots.}
Each violin shows the Bayesian posterior over $p$ given the observed outcomes, under a uniform prior. With $s$ successes out of $n$ rollouts and a $\mathrm{Beta}(1,1)$ prior, the posterior is $\mathrm{Beta}(s+1,\, n-s+1)$. The width of the violin at success rate $x$ is proportional to the posterior density at $x$, and the horizontal marker denotes the empirical success rate $s/n$. The violin thus reflects uncertainty in the true success rate given the number of rollouts, not variation across tasks.

\paragraph{Aggregation across tasks.}
For aggregate plots, we pool outcomes across the tasks of a benchmark and plot $\mathrm{Beta}\big(\sum_i s_i + 1,\, \sum_i (n_i - s_i) + 1\big)$. Because all tasks within a benchmark use the same number of rollouts, the pooled success rate equals the unweighted mean of per-task success rates. As noted by \citet{lbmtri2025}, pooling treats outcomes from different tasks as exchangeable rather than strictly i.i.d.

\section{Detailed Experiment Results \& Analysis For Main Paper}
\label{app:additional_results}

This section reports per-task results for every experiment in
Sec.~\ref{sec:experiments} and follows the same order.
Sec.~\ref{app:results_overall} corresponds to the comparison with existing
memory-augmented VLAs (Sec.~\ref{sec:exp_overall_eval}),
Sec.~\ref{app:results_encoding} to memory encoding
(Sec.~\ref{sec:exp_encoding}), and Sec.~\ref{app:results_learning} as well as Sec.~\ref{app:representation_analysis} to memory
learning and VLA backbones (Sec.~\ref{sec:exp_learning}). Unless noted
otherwise, all entries are task success rates in percent. RMBench uses 100
evaluation rollouts per task and RoboCasa uses 50. Averages are
task-balanced: we first compute success within each task and then average
tasks with equal weight.

\subsection{Comparison with Memory-Augmented VLAs}
\label{app:results_overall}

\subsubsection{Per-task Results on RMBench}
\label{app:rmbench_per_task}

Table~\ref{tab:rmbench_per_task} groups the RMBench tasks by whether they
require recalling a single event, M(1), or a sequence of events, M(n)
(App.~\ref{app:rmbench_tasks}). On M(1) tasks, several methods perform well
once the relevant frame is retained. EventVLA, which always keeps the initial
frame, averages 79.0\% on these tasks, close to 81.0\% for recurrent
\textsc{IMPRINT}. The gap widens on M(n) tasks, where recurrent
\textsc{IMPRINT} averages 62.5\%, compared with 54.0\% for EventVLA and at
most 38.8\% for the other baselines. Press Button, which requires counting
how many times each button has been pressed, shows this difference most
clearly. Every baseline stays at or below 3\%, whereas \textsc{IMPRINT}
reaches 21\% with full attention and 31\% with the recurrent encoder. Because
each press doesn't change the scene and the robot needs to keep along history to count the press properly,  only \textsc{IMPRINT} retains the history with the proper compression. Observe
and Pickup and Battery remain difficult for all methods, with no method
exceeding 21\% and 41\%, respectively because the former has a diverse set of similar objects that might be hard to tell and the latter requires high precision to grasp the round objects as battery and failure of grasping is irrecoverable. 

\begin{table*}[t]
\centering
\scriptsize
\setlength{\tabcolsep}{2.7pt}
\renewcommand{\arraystretch}{1.08}
\caption{Per-task success rates on the nine RMBench memory tasks,
grouped into single-event, M(1), and multi-event, M(n), tasks. The final
column is the task-balanced average reported in Fig.~\ref{fig:rmbench_results}.
The best result for each task and for the average is shown in bold.
$^\dagger$Results reported by prior work (App.~\ref{app:baselines}).}
\label{tab:rmbench_per_task}
\resizebox{\textwidth}{!}{%
\begin{tabular}{@{}lrrrrrrrrrr@{}}
\toprule
& \multicolumn{5}{c}{M(1)} & \multicolumn{4}{c}{M(n)} & \\
\cmidrule(lr){2-6}\cmidrule(lr){7-10}
Method &
\shortstack{Observe /\\pick} &
\shortstack{Rearrange\\blocks} &
\shortstack{Put back\\block} &
\shortstack{Swap\\blocks} &
Swap T &
Battery &
\shortstack{Block\\ranking} &
\shortstack{Cover\\blocks} &
\shortstack{Press\\button} &
Average \\
\midrule
$\pi_{0.5}$ (no memory) & 8 & 63 & 20 & 16 & 8 & 30 & 66 & 15 & 3 & 25.4  \\
Uniform sampling                 & 5  & 79 & \textbf{100} & 62 & 30 & 28 & 82   & 13 & 3 & 44.7 \\
EventVLA$^\dagger$               & \textbf{21} & 96 & 95  & 96 & 87 & 35 & 81   & 97 & 3 & 67.8 \\
MemER$^\dagger$                  & 7  & 17 & 0   & 14 & 7  & 27 & 0    & 6  & 0 & 8.7 \\
Mem-0$^\dagger$                  & 4  & 89 & 90  & 67 & 14 & 28 & 18   & 68 & 0 & 42.0 \\
HAMLET                           & 6  & 76 & 23  & 25 & 13 & 26 & 70   & 14 & 0 & 28.1 \\
MemoryVLA$^\dagger$              & 2  & 53 & 81  & 76 & 9  & 33 & 53   & 69 & 0 & 41.7 \\
\midrule
\textsc{IMPRINT} (full)          & 9 & \textbf{100} & \textbf{100} & \textbf{100} & \textbf{100} & \textbf{41} & 83 & \textbf{100} & 21 & 72.7 \\
\textsc{IMPRINT} (recurrent)     & 6 & 99  & \textbf{100} & \textbf{100} & \textbf{100} & 35 & \textbf{84} & \textbf{100} & \textbf{31} & \textbf{72.8} \\
\bottomrule
\end{tabular}%
}
\end{table*}

\subsubsection{Per-task Results on RoboCasa}
\label{app:robocasa_per_task}

Table~\ref{tab:robocasa_per_task} shows that naively adding historical frames
to the policy does not reliably help. Both heuristic variants have lower
overall success than the non-memory $\pi_{0.5}$ (48.8\% and 41.5\% vs.\
52.2\%). Supplying the frames as additional VLM-context images gives almost
no benefit on memory tasks (39.0\% vs.\ 38.7\%). On Retrieve Oil and
Wash and Return, for example, it performs within 2 points of the non-memory
policy (50\% vs.\ 48\% and 46\% vs.\ 46\%). Supplying the same frames to the
action expert is more effective, raising memory-task success to 54.0\%. The
largest gains appear when the required memory is a single visual snapshot:
88\% on Retrieve Fruit and 98\% on Retrieve Oil, where the policy only needs
to recall where the target was seen. We attribute this difference to the
injection point. Adding images to the VLM context changes the input
distribution of the pretrained VLM, whereas injecting memory into the action
expert leaves the VLM input unchanged, consistent with the findings
of~\citet{dai2026robomme}. \textsc{IMPRINT} therefore uses the action expert
as its injection point.

Neither heuristic variant, however, preserves performance on non-memory tasks. Adding
raw frames lowers success from 65.7\% to 43.7\% in the action
expert and to 44.0\% in the VLM context for non-memory tasks. Examples include Pan Transfer (74\% to
48\% and 44\%), Organize Condiments (72\% to 32\% and 48\%), and Frying-Pan
Adjustment (76\% to 52\% and 22\%). When the current observation is
sufficient, the extra frames likely act as distractors that the policy cannot
learn to ignore.

\textsc{IMPRINT} injects memory at the same point as the action-expert
variant, so the remaining gap reflects the memory representation rather than
the injection point. Replacing raw frames with compressed multi-stream memory
raises success from 54.0\% to 65.3\% for memory tasks and 
43.7\% to 68.0\% for non-memory tasks, above the non-memory $\pi_{0.5}$. The gains come mainly from
tasks that depend on what the robot has already done. Recurrent
\textsc{IMPRINT} improves Wash and Return from 58\% to 96\%, Wash and Return
to Same Location from 36\% to 58\%, and Sweeten Hot Chocolate from 36\% to
64\%. On non-memory tasks, it matches or exceeds the non-memory $\pi_{0.5}$
on four of six tasks. Rinse Cutting Board stays at or below 8\% and Place
Equal Ice Cubes at or below 20\% for every method.

\begin{table*}[t]
\centering
\scriptsize
\setlength{\tabcolsep}{3.2pt}
\renewcommand{\arraystretch}{1.08}
\caption{Per-task success rates on the customized RoboCasa benchmark
(Fig.~\ref{fig:robocasa_results}). \emph{Heuristic action} supplies uniformly
sampled historical frames to the action expert, whereas \emph{heuristic VLM}
supplies them as additional VLM-context images. The best result for each task
and each average is shown in bold.}
\label{tab:robocasa_per_task}
\resizebox{\textwidth}{!}{%
\begin{tabular}{@{}llrrrrr@{}}
\toprule
Task & Type &
$\pi_{0.5}$ &
\shortstack{Heuristic\\action} &
\shortstack{Heuristic\\VLM} &
\shortstack{\textsc{IMPRINT}\\full} &
\shortstack{\textsc{IMPRINT}\\recurrent} \\
\midrule
Retrieve fruit                    & Memory      & 42 & 88 & 60 & 84 & \textbf{90} \\
Retrieve oil                      & Memory      & 48 & \textbf{98} & 50 & 76 & 80 \\
Wash and return                   & Memory      & 46 & 58 & 46 & 86 & \textbf{96} \\
Wash and return to same location  & Memory      & 46 & 36 & 44 & \textbf{62} & 58 \\
Rinse cutting board               & Memory      & 4  & \textbf{8}  & 0  & 0  & 4 \\
Sweeten hot chocolate             & Memory      & 46 & 36 & 34 & 58 & \textbf{64} \\
\midrule
Pan transfer                      & Non-memory & 74 & 48 & 44 & \textbf{76} & 72 \\
Frying-pan adjustment             & Non-memory & \textbf{76} & 52 & 22 & 72 & 70 \\
Organize condiments               & Non-memory & 72 & 32 & 48 & 62 & \textbf{74} \\
Place equal ice cubes             & Non-memory & 16 & 6  & 6  & 6  & \textbf{20} \\
Scrub cutting board               & Non-memory & 60 & 50 & 56 & 66 & \textbf{76} \\
Counter to sink                   & Non-memory & 96 & 74 & 88 & \textbf{98} & 96 \\
\midrule
Memory task average               & --          & 38.7 & 54.0 & 39.0 & 61.0 & \textbf{65.3} \\
Non-memory task average          & --          & 65.7 & 43.7 & 44.0 & 63.3 & \textbf{68.0} \\
Overall average                   & --          & 52.2 & 48.8 & 41.5 & 62.2 & \textbf{66.7} \\
\bottomrule
\end{tabular}%
}
\end{table*}

\subsubsection{Detailed Real-world Results}
\label{app:realworld_detailed}

Table~\ref{tab:realworld_detailed} reports the rollout counts behind
Fig.~\ref{fig:real_world_results}. \textsc{IMPRINT} improves over the
strongest baseline, uniform sampling, by 30, 25, and 25 points on Dust Table,
Scan Bottle, and Fold Cloth.

The baselines fail in different ways. MemER is limited by its hierarchical
design. In our rollouts, its high-level VLM often predicts the wrong subtask,
for example returning to a bottle that has already been scanned. Even when
the prediction is correct, the low-level policy's actions change abruptly
when the subtask switches, which frequently disrupts an ongoing grasp or
placement. HAMLET does not improve over $\pi_0$ on any task. Its memory often
retains the wrong information, possibly because the time-contrastive objective is not learning the critical information for compressing the memory. Its action predictions are also imprecise and produce jerky motions
that cause manipulation failures even when the remembered information is
correct.

Fold Cloth is the lowest-scoring task for $\pi_0$, HAMLET, and uniform
sampling. Reproducing the in-context example requires both its visual
appearance, to identify the folding style, and the demonstrated actions and
proprioceptive states, to reproduce the folding motion. Baselines that retain
only visual history can observe the example but not recover how it was
executed. \textsc{IMPRINT} encodes all three streams of the example and
reaches 72\% on this task.

Under the OOD object variations (App.~\ref{app:real_world_eval}),
\textsc{IMPRINT} retains most of its performance on Scan Bottle (70\% vs.\
75\%) but drops on Dust Table (45\%) and Fold Cloth (31\%), which involve
longer horizons and more precise placement.

\begin{table*}[t]
\centering
\small
\setlength{\tabcolsep}{7pt}
\renewcommand{\arraystretch}{1.08}
\caption{Real-world results (Fig.~\ref{fig:real_world_results}). Each cell
reports successful rollouts / total rollouts, followed by the success rate in
parentheses. The average is the mean of the per-task success rates. The best
in-distribution result for each task and the average is shown in bold. OOD
denotes the held-out object variations.}
\label{tab:realworld_detailed}
\begin{tabular}{@{}lcccc@{}}
\toprule
Method & Dust Table & Scan Bottle & Fold Cloth & Average \\
\midrule
$\pi_0$              & 6/20 (30\%)  & 5/20 (25\%)  & 7/32 (22\%)  & 25.6\% \\
HAMLET               & 5/20 (25\%)  & 5/20 (25\%)  & 6/32 (19\%)  & 22.9\% \\
MemER                & 8/20 (40\%)  & 7/20 (35\%)  & 12/32 (38\%) & 37.5\% \\
Uniform sampling     & 10/20 (50\%) & 10/20 (50\%) & 15/32 (47\%) & 49.0\% \\
\textsc{IMPRINT}     & \textbf{16/20 (80\%)} & \textbf{15/20 (75\%)} & \textbf{23/32 (72\%)} & \textbf{75.6\%} \\
\midrule
\textsc{IMPRINT} (OOD) & 9/20 (45\%) & 14/20 (70\%) & 10/32 (31\%) & 48.8\% \\
\bottomrule
\end{tabular}
\end{table*}

\subsection{Memory Encoding}
\label{app:results_encoding}

\subsubsection{Memory Capacity}
\label{app:capacity_per_task}

Table~\ref{tab:capacity_per_task} shows that task performance generally improves with memory capacity up to $B=256$. As the number of memory tokens grows from 4 to 256, average success rises on RMBench (67.0\% to 72.7\%) and on RoboCasa memory tasks (54.0\% to 61.0\%). Non-memory success stays near 57\% through 64 tokens and rises to 63.3\% at 256. Additional capacity mainly helps tasks that require retaining several pieces of information. On RMBench, Cover Blocks, which requires the order of three colors, rises from 68\% with 4 tokens to 100\% with 256 tokens, and Block Ranking rises from 73\% to 83\%. On RoboCasa, Sweeten Hot Chocolate improves steadily from 40\% to 58\%. In contrast, tasks that depend on a single location or assignment, such as Rearrange Blocks and Put Back Block, already reach 100\% with a single token.

To test the limits of this trend, we additionally train variants with $B=1$ and $B=1024$ on RMBench, keeping all other settings identical. At the low end, a single memory token already reaches 67.0\%, matching $B=4$, and attains the highest Press Button success of any capacity (29\%). This suggests that a highly compressed memory can still retain coarse counting information, but it underperforms on tasks that require several pieces of information, such as Block Ranking (64\%) and Cover Blocks (74\%). At the high end, increasing capacity to $B=1024$ does not yield further gains: average success decreases to 70.0\%, below $B=256$ and comparable to $B=64$ (70.6\%). Relative to $B=256$, the largest drops occur on Block Ranking (83\% to 68\%), Swap Blocks (100\% to 93\%), and Press Button (21\% to 15\%), while Battery improves (41\% to 46\%). Notably, $B=1024$ attains the lowest Stage-1 losses in Table~\ref{tab:stage1_capacity_losses} but not the highest task success, reinforcing that reconstruction accuracy alone does not determine control performance. Overall, memory benefits from sufficient capacity without harming tasks that do not rely on memory, but it does not need to grow with the observation history, and excess capacity may leave the representation less constrained without adding task-relevant information. We therefore use $B=256$ by default.

\paragraph{Stage-1 Supervision-Loss Evaluation.}
We evaluate checkpoints at 160k updates for
$B \in \{1,4,16,\\ 64,256,1024\}$ on all 450 RMBench
validation dataset from policy rollouts in Fig.~\ref{fig:memory_capacity}. We deterministically select four uniformly
spaced causal prefixes per rollout and 16 historical timesteps
per prefix, yielding 1,800 prefixes and 28,800 historical-step queries
per checkpoint.

We evaluate the same Stage-1 objective:
\begin{equation}
\mathcal{L}_{\mathrm{mem}}
=
\mathcal{L}_{\mathrm{vision}}
+ 0.5\mathcal{L}_{\mathrm{action}}
+ 0.5\mathcal{L}_{\mathrm{state}}
+ 0.5\mathcal{L}_{\mathrm{progress}}.
\end{equation}
Visual reconstruction uses 192 DINOv3 patch features per queried
timestep, with parameter-free LayerNorm applied independently to
predictions and targets. Reconstruction errors are averaged over
feature dimensions and queried timesteps. Progress is predicted at
each prefix endpoint using the normalized timestep
$\rho_t = t/(T-1)$ for a rollout of length $T$.
We average losses over prefixes within each task and equally across
the nine tasks, reporting standard errors across task means.

\begin{table}[t]
\centering
\small
\setlength{\tabcolsep}{5pt}
\caption{Stage-1 losses across memory capacities. Total loss is
mean $\pm$ SEM across tasks; component losses are unweighted.
The second header row indicates scaling.}
\label{tab:stage1_capacity_losses}
\begin{tabular}{rccccc}
\toprule
$B$ & Total & Vision & Action & State & Progress \\
& $\times 10^{-2}$ & $\times 10^{-2}$
& $\times 10^{-4}$ & $\times 10^{-4}$ & $\times 10^{-4}$ \\
\midrule
1    & $3.649 \pm 0.295$ & 3.599 & 2.37 & 2.43 & 5.13 \\
4    & $3.308 \pm 0.273$ & 3.278 & 1.31 & 1.35 & 3.31 \\
16   & $3.406 \pm 0.277$ & 3.368 & 1.50 & 1.55 & 4.41 \\
64   & $3.389 \pm 0.282$ & 3.359 & 1.50 & 1.52 & 2.94 \\
256  & $3.338 \pm 0.280$ & 3.311 & 1.27 & 1.31 & 2.81 \\
1024 & $3.256 \pm 0.270$ & 3.232 & 1.08 & 1.10 & 2.63 \\
\bottomrule
\end{tabular}
\end{table}

Losses generally decrease with capacity but are non-monotonic
(Table~\ref{tab:stage1_capacity_losses}). Increasing capacity from
4 to 16 tokens raises reconstruction errors despite improving
downstream memory-task success (Fig.~\ref{fig:memory_capacity}).
This is consistent with using reconstruction to learn useful memory
representations for policy adaptation, rather than treating
reconstruction accuracy as a direct measure of control performance.

\begin{table}[t]
\centering
\caption{Per-task success as the capacity of the full-attention memory encoder is varied (Fig.~\ref{fig:memory_capacity}). All other architecture, training, and evaluation settings are fixed. $B=1$ and $B=1024$ are evaluated on RMBench only (--). The best result in each row is shown in bold; rows where all settings tie are left unmarked.}
\label{tab:capacity_per_task}
\small
\setlength{\tabcolsep}{5pt}
\begin{tabular}{llcccccc}
\toprule
Task & Type & $B{=}1$ & $B{=}4$ & $B{=}16$ & $B{=}64$ & $B{=}256$ & $B{=}1024$ \\
\midrule
\multicolumn{8}{l}{\textit{RMBench}} \\
Observe / pick   & Memory & 7  & 8   & 8   & 8   & 9            & \textbf{10}  \\
Rearrange blocks & Memory & \textbf{100} & \textbf{100} & \textbf{100} & \textbf{100} & \textbf{100} & 99 \\
Put back block   & Memory & 100 & 100 & 100 & 100 & 100 & 100 \\
Swap blocks      & Memory & 97 & 97  & 99  & 99  & \textbf{100} & 93  \\
Swap T           & Memory & 99 & 99  & 99  & 99  & \textbf{100} & 99  \\
Battery          & Memory & 33 & 40  & 44  & 39  & 41           & \textbf{46}  \\
Block ranking    & Memory & 64 & 73  & 79  & 81  & \textbf{83}  & 68  \\
Cover blocks     & Memory & 74 & 68  & 83  & 83  & \textbf{100} & \textbf{100} \\
Press button     & Memory & \textbf{29} & 18 & 15 & 26 & 21 & 15 \\
\midrule
RMBench average  & Memory & 67.0 & 67.0 & 69.7 & 70.6 & \textbf{72.7} & 70.0 \\
\midrule
\multicolumn{8}{l}{\textit{RoboCasa}} \\
Retrieve fruit                   & Memory & -- & 80 & 74 & 80 & \textbf{84} & -- \\
Retrieve oil                     & Memory & -- & 66 & \textbf{78} & 70 & 76 & -- \\
Wash and return                  & Memory & -- & \textbf{92} & 84 & 82 & 86 & -- \\
Wash and return to same location & Memory & -- & 46 & \textbf{62} & 60 & \textbf{62} & -- \\
Rinse cutting board              & Memory & -- & 0  & 0  & 0  & 0  & -- \\
Sweeten hot chocolate            & Memory & -- & 40 & 44 & 54 & \textbf{58} & -- \\
\midrule
RoboCasa memory average          & Memory & -- & 54.0 & 57.0 & 57.7 & \textbf{61.0} & -- \\
\midrule
Pan transfer                     & Non-memory & -- & \textbf{78} & 68 & 70 & 76 & -- \\
Frying-pan adjustment            & Non-memory & -- & 58 & 58 & 58 & \textbf{72} & -- \\
Organize condiments              & Non-memory & -- & 48 & 52 & 60 & \textbf{62} & -- \\
Place equal ice cubes            & Non-memory & -- & 4  & \textbf{10} & 6 & 6 & -- \\
Scrub cutting board              & Non-memory & -- & 60 & 60 & 56 & \textbf{66} & -- \\
Counter to sink                  & Non-memory & -- & \textbf{98} & 94 & 96 & \textbf{98} & -- \\
\midrule
RoboCasa non-memory average      & Non-memory & -- & 57.7 & 57.0 & 57.7 & \textbf{63.3} & -- \\
\bottomrule
\end{tabular}
\end{table}

\subsubsection{Recurrent and Full-attention Encoders}
\label{app:encoder_per_task}

The per-task comparison between the two encoders is included in
Tables~\ref{tab:rmbench_per_task} and~\ref{tab:robocasa_per_task}. On
RMBench, the two encoders reach nearly identical average success (72.8\% vs.\
72.7\%) and differ by at most 10 points on any task, with the largest
differences on Press Button (31\% recurrent vs.\ 21\% full) and Battery
(35\% vs.\ 41\%). On RoboCasa, the recurrent encoder improves average
success from 62.2\% to 66.7\% and is higher on 8 of 12 tasks, with the
largest gains on Place Equal Ice Cubes (20\% vs.\ 6\%), Organize Condiments
(74\% vs.\ 62\%), Scrub Cutting Board (76\% vs.\ 66\%), and Wash and Return
(96\% vs.\ 86\%).

The two encoders also differ in inference cost. The full-attention encoder
revisits the entire history at every query, so its cost grows with episode
length. The recurrent encoder instead updates a fixed-size state once per
chunk, so its cost stays bounded. As a result, the recurrent encoder reduces
latency from 130.8\,ms to 122.8\,ms per action chunk
(Fig.~\ref{fig:inference_cost}; detailed inference time analysis in
App.~\ref{app:timing_protocol}) by averaging over the entire trajectory. It also grows more slowly than full attention variant when the horizon length increases. Overall, the recurrent encoder matches the
full-attention encoder on RMBench, improves it on RoboCasa, and runs faster,
which is why we use it as the default \textsc{IMPRINT} encoder.

\subsection{Memory Learning and VLA Backbones}
\label{app:results_learning}

\subsubsection{Training Recipe and Stage~1 Objectives}
\label{app:ablation_per_task}

Table~\ref{tab:learning_per_task} shows that Stage~1 training alone is
sufficient for some tasks. With frozen memory, Swap T rises from 45\% to
99\% and Rearrange Blocks from 90\% to 99\%. Other tasks require joint
adaptation. Cover Blocks rises only from 15\% with end-to-end training to
35\% with frozen memory, but reaches 100\% after joint adaptation. This is
consistent with the analysis in Fig.~\ref{fig:memory_bottleneck_analysis},
where frozen memory already encodes the color order but the policy cannot yet
use it. On Battery, Block Ranking, and Press Button, frozen memory is even
below end-to-end training, and only joint adaptation improves success.

The memory objective ablation shows how reconstruction supervision from different streams affects task performance. Visual
reconstruction is essential for Cover Blocks, whose memory is the initial
color order: without it, success drops from 100\% to 52\%, whereas visual
reconstruction alone keeps 100\%. Block Ranking follows the same pattern
(61\% without visual reconstruction vs.\ 85\% with it alone). Conversely,
action and state reconstruction benefit Rearrange Blocks (100\% vs.\ 78\% with
visual reconstruction only) and Press Button (23\% vs.\ 15\%). Removing
progress estimation mainly hurts multi-stage tasks, lowering Rearrange Blocks
to 91\%, Swap Blocks to 90\%, and Block Ranking to 73\%.

\begin{table*}[t]
\centering
\scriptsize
\setlength{\tabcolsep}{3.2pt}
\renewcommand{\arraystretch}{1.08}
\caption{Per-task success for the training-recipe and Stage~1 objective
ablations on RMBench (Fig.~\ref{fig:training_ablation}). The best result in
each row is shown in bold separately within each ablation; rows where all
variants tie are left unmarked.}
\label{tab:learning_per_task}
\resizebox{\textwidth}{!}{%
\begin{tabular}{@{}lrrrrrrr@{}}
\toprule
& \multicolumn{3}{c}{Training recipe} &
\multicolumn{4}{c}{Stage~1 objective} \\
\cmidrule(lr){2-4}\cmidrule(lr){5-8}
Task &
E2E &
\shortstack{\textsc{IMPRINT}\\frozen} &
\textsc{IMPRINT} &
\textsc{IMPRINT} &
\shortstack{Without\\progress} &
\shortstack{Visual\\only} &
\shortstack{Action +\\state} \\
\midrule
Observe / pick    & \textbf{12} & 7   & 9   & \textbf{9}   & 6   & 6   & 5 \\
Rearrange blocks  & 90  & 99  & \textbf{100} & \textbf{100} & 91  & 78  & \textbf{100} \\
Put back block    & 100 & 100 & 100 & 100 & 100 & 100 & 100 \\
Swap blocks       & 90  & 99  & \textbf{100} & \textbf{100} & 90  & 89  & 89 \\
Swap T            & 45  & 99  & \textbf{100} & \textbf{100} & 97  & 99  & 98 \\
Battery           & 27  & 20  & \textbf{41}  & \textbf{41}  & 35  & 30  & 36 \\
Block ranking     & 73  & 68  & \textbf{83}  & 83  & 73  & \textbf{85}  & 61 \\
Cover blocks      & 15  & 35  & \textbf{100} & \textbf{100} & \textbf{100} & \textbf{100} & 52 \\
Press button      & 5   & 2   & \textbf{21}  & 21  & \textbf{26}  & 15  & 23 \\
\midrule
\textit{Average}  & 50.8 & 58.8 & \textbf{72.7} & \textbf{72.7} & 68.7 & 66.9 & 62.7 \\
\bottomrule
\end{tabular}%
}
\end{table*}

\begin{table*}[t]
\centering
\scriptsize
\setlength{\tabcolsep}{3.5pt}
\renewcommand{\arraystretch}{1.08}
\caption{Per-task success with and without \textsc{IMPRINT} across three VLA
backbones on RMBench (Fig.~\ref{fig:flexible_vla}). For each backbone, the
better result of the two variants is shown in bold; ties are left unmarked.}
\label{tab:backbone_per_task}
\resizebox{\textwidth}{!}{%
\begin{tabular}{@{}lrrrrrr@{}}
\toprule
& \multicolumn{2}{c}{$\pi_{0.5}$} &
\multicolumn{2}{c}{$\pi_0$} &
\multicolumn{2}{c}{MolmoAct2} \\
\cmidrule(lr){2-3}\cmidrule(lr){4-5}\cmidrule(lr){6-7}
Task &
\shortstack{No\\memory} & \textsc{IMPRINT} &
\shortstack{No\\memory} & \textsc{IMPRINT} &
\shortstack{No\\memory} & \textsc{IMPRINT} \\
\midrule
Observe / pick    & 8    & \textbf{9}   & \textbf{8}  & 6   & 4  & 4 \\
Rearrange blocks  & 63   & \textbf{100} & 13 & \textbf{99}  & 52 & \textbf{100} \\
Put back block    & 20   & \textbf{100} & 49 & \textbf{100} & 0  & \textbf{100} \\
Swap blocks       & 16   & \textbf{100} & 4  & \textbf{100} & 11 & \textbf{65} \\
Swap T            & 8    & \textbf{100} & 0  & \textbf{98}  & 15 & \textbf{100} \\
Battery           & 30   & \textbf{41}  & 27 & \textbf{30}  & \textbf{7}  & 5 \\
Block ranking     & 66 & \textbf{83}  & 54 & \textbf{56}  & 62 & \textbf{74} \\
Cover blocks      & 15   & \textbf{100} & 0  & \textbf{78}  & 18 & \textbf{83} \\
Press button      & 3    & \textbf{21}  & \textbf{2}  & 0   & 0  & \textbf{6} \\
\midrule
\textit{Average}  & 25.4 & \textbf{72.7} & 17.4 & \textbf{63.0} & 18.8 & \textbf{59.7} \\
\bottomrule
\end{tabular}%
}
\end{table*}
\subsection{Analysis of the Learned Memory}
\label{app:representation_analysis}
\label{app:memory_utilization}

This section describes the analyses in Fig.~\ref{fig:memory_bottleneck_analysis}.
All of them use Cover Blocks, where the robot must recall the initial
left-to-right order of the red, green, and blue blocks, a six-way
classification over their permutations. We compare three memory encoders:
end-to-end training from scratch (E2E), frozen Stage~1 memory (Frozen), and
jointly adapted \textsc{IMPRINT} (Ours). For each encoder, we reconstruct the
causal history at 64 uniformly spaced timesteps in each of the 50 Cover Blocks
demonstrations and pass the visual, action, proprioceptive, and language
inputs through the fixed memory encoder. We concatenate the mean- and
max-pooled features of its $B=256$ output tokens to obtain a 2,048-dimensional
representation. The six color orders appear in 11, 7, 10, 10, 3, and 9
demonstrations for B--G--R, B--R--G, G--B--R, G--R--B, R--B--G, and R--G--B,
respectively. Because these demonstrations were also used to train the
memory, the analyses describe the learned representation rather than
generalization to new trajectories.

\paragraph{Linear probe.}
We measure whether the learned latent memory representation automatically include color order information by
training a six-way linear softmax classifier on the features, giving
3,200 samples per encoder. We use five-fold cross-validation stratified by
class at the episode level, so all timesteps of an evaluation episode are
excluded from training. Features are standardized with training-fold
statistics, and the classifier is optimized for 300 full-batch gradient steps
with a learning rate of 0.2 and $\ell_2$ regularization of $10^{-3}$.
Timestep accuracy scores each sampled prefix, and episode accuracy averages
the logits over an episode before predicting its class. We also report
balanced accuracy to account for the class imbalance. The task-success
column reports closed-loop rollouts on Cover Blocks and is not used to train
the probe. As shown in Table~\ref{tab:cover_order_probe}, Stage~1 training
raises timestep accuracy from 19.94\% to 83.31\%, and joint adaptation raises
it further to 91.53\%.

\begin{table}[t]
\centering
\small
\caption{Cover Blocks color-order probes and closed-loop task success
(\%). Probes are evaluated on held-out episode folds drawn from the memory
model's training demonstrations.}
\label{tab:cover_order_probe}
\begin{tabular}{@{}lrrrr@{}}
\toprule
Memory training & \shortstack{Timestep\\accuracy} &
\shortstack{Balanced\\accuracy} & \shortstack{Episode\\accuracy} &
\shortstack{Task\\success}\\
\midrule
E2E & 19.94 & 17.29 & 42.00 & 15.00\\
Frozen & 83.31 & 83.68 & 90.00 & 35.00\\
\textsc{IMPRINT} & \textbf{91.53} & \textbf{85.95} &
\textbf{98.00} & \textbf{100.00}\\
\bottomrule
\end{tabular}
\end{table}

\paragraph{t-SNE visualization.}
We visualize only the second half of each episode, closer to the uncovering
phase in which the color order guides behavior, which gives 32 points per
episode and 1,600 points per encoder. The features are standardized and reduced
to 50 dimensions with PCA, and then embedded with t-SNE using Euclidean
distance, perplexity 30, PCA initialization, an automatic learning rate,
1,500 iterations, and seed 42. Class labels are used only for coloring and
for computing separation afterward. Point color indicates the color-order
class, and saturation and opacity increase with episode progress from 50\% to
100\%.

The mean silhouette coefficient in the two-dimensional embedding is $-0.08$
for E2E, $0.23$ for Frozen, and $0.36$ for \textsc{IMPRINT}. In the second
half of the episode, the blocks are largely covered, so the color order can
no longer be read from the current observation and must come from memory.
The E2E memory shows no grouping by color order: its negative silhouette
indicates that episodes with different orders are mixed together, consistent
with its near-chance probe accuracy (19.94\% vs.\ 16.7\% chance). After
Stage~1 training alone, the Frozen memory already groups episodes by color
order, even though Stage~1 receives neither explicit color-order labels nor gradients from a downstream policy. This shows that reconstructing the multi-stream history is
enough for the memory to retain this hidden variable from the early frames.
Joint adaptation separates the groups further, indicating that the action
objective refines the memory toward the information the policy needs. Together,
these results show that the learned memory organizes past experience by the
task-relevant variable rather than by the current scene. Because t-SNE is
fitted separately for each encoder, only the structure within each panel is
comparable, not coordinates across panels. The linear probe provides the
quantitative measure of how much color-order information the memory retains.

\paragraph{Attention visualization.}
At inference step 750, we record the cross-attention weights of the
full-attention memory encoder. In each of its two blocks and eight heads,
these weights connect the 256 latent queries to history tokens. Averaging over blocks,
heads, and queries yields one attention score per history token. We then assign the valid tokens to five streams: head-camera,
left-wrist, and right-wrist observations, actions, and proprioceptive states.
Visual memory tokens are mapped back to their temporal chunk and camera view,
while action and state tokens keep their original timestep. Scores are ranked
separately within each stream and displayed from timestep 0 to 750 with a
purple-to-red colormap, where warmer colors indicate higher attention within
that stream. In the displayed episode, Frozen and \textsc{IMPRINT} place
substantial attention on the initial head-camera frames that reveal the color
order, whereas E2E emphasizes proprioceptive history. This visualization
shows which inputs the memory attends to in one episode and does not
establish their causal contribution. The recurrent encoder has no
cross-attention over history tokens and would require a separate attribution
method.

\begin{figure}
      \centering
      \includegraphics[width=\linewidth]{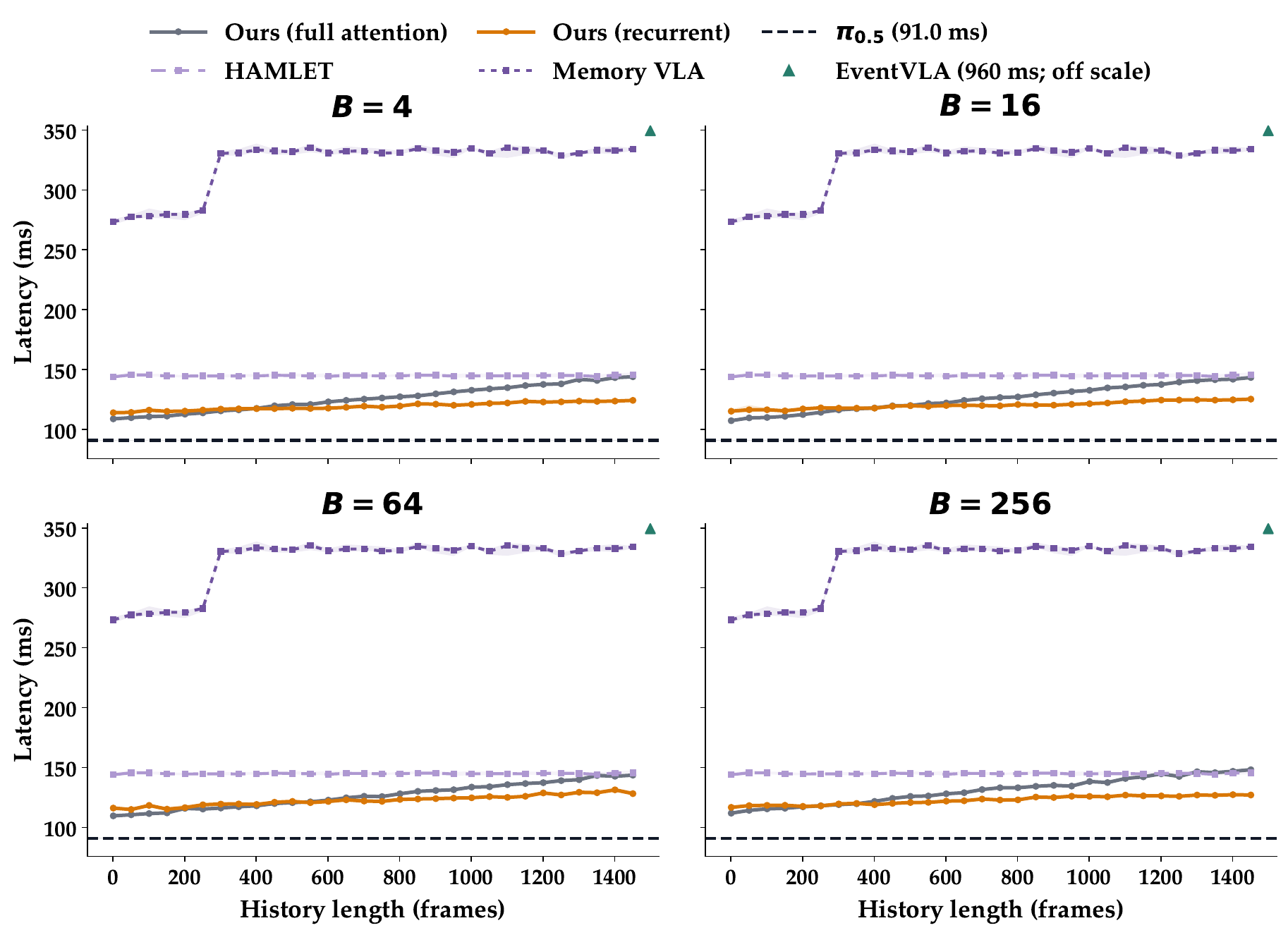}
      \caption{\textbf{Inference Time Analysis For Different Bottleneck Size and Different Methods}}
      \label{fig:inference_time_plot}
  \end{figure}
\subsection{Inference-Time Analysis}
  \label{app:timing_protocol}

  \paragraph{Protocol.}
  We measure the synchronized wall-clock latency of one policy call on an NVIDIA
  L40S GPU with batch size one. Unless stated otherwise, each call predicts a
  50-step action chunk. We vary the available history from 1 to 1,451 environment
  frames in increments of 50 frames. Curves report median latency, and shaded
  regions denote the interquartile range. Environment stepping, video recording,
  and the subsequent history-buffer update are excluded. For our method,
  recent-image features are computed once per history state and reused across
  timing repetitions, matching the steady-state cached evaluation path. The full result of runtime analysis is in Fig.~\ref{fig:inference_time_plot}.

  \paragraph{Comparison with existing methods.}
  The non-memory $\pi_{0.5}$ policy requires $91.0$\,ms per 50-action prediction,
  providing a lower bound corresponding to the underlying policy without memory.
  At $B=256$, our recurrent memory requires $116.8$\,ms at the shortest history
  and $127.1$\,ms at 1,451 frames. Thus, incorporating long-term memory introduces
  a $36.1$\,ms overhead over $\pi_{0.5}$ at the longest history, while retaining
  a sub-$130$\,ms policy latency.

  HAMLET requires approximately $144$--$146$\,ms throughout the episode. Its
  moment-token extraction, rolling-memory update, memory Transformer, and
  50-action generation are all included in this measurement. HAMLET uses a
  fixed $K=8$ rolling window with a stride of 50 frames. Its cache contents are
  updated at every policy call, but the number of processed moment tokens remains
  constant; consequently, its latency does not increase with episode length.
  At 1,451 frames, our recurrent $B=256$ model is approximately $18.6$\,ms, or
  $12.8\%$, faster than HAMLET.

  Memory VLA is substantially more expensive, increasing from $273.3$\,ms to
  $334.2$\,ms. It predicts a  50-step action chunk, so
  we report its measured latency without rescaling. The abrupt increase between
  251 and 301 frames occurs when its 256-entry memory becomes full. Every
  subsequent update activates token-merging consolidation, increasing latency
  from approximately $283$\,ms to $330$\,ms.
  Memory VLA is approximately $2.6\times$ slower than our recurrent model at the
  longest history. EventVLA has a reported latency of $960$\,ms and is shown
  off-scale for reference.

  \paragraph{Effect of memory capacity.}
  Table~\ref{tab:inference_capacity} reports latency as the number of memory
  tokens $B$ varies. These measurements slice the query bank of the same
  $B=256$ variant; they isolate inference cost and are not independently
  trained capacity variants.

  \begin{table}[t]
  \centering
  \caption{Policy-call latency in milliseconds at the shortest and longest
  history. All variants of our method predict 50 actions.}
  \label{tab:inference_capacity}
  \begin{tabular}{c|cc|cc}
  \toprule
  & \multicolumn{2}{c|}{Full attention}
  & \multicolumn{2}{c}{Recurrent} \\
  $B$ & 1 frame & 1,451 frames & 1 frame & 1,451 frames \\
  \midrule
  4   & 109.0 & 144.1 & 114.1 & 124.4 \\
  16  & 107.4 & 143.4 & 115.3 & 125.4 \\
  64  & 109.7 & 143.7 & 116.2 & 128.3 \\
  256 & 111.9 & 148.2 & 116.8 & 127.1 \\
  \bottomrule
  \end{tabular}
  \end{table}

  Increasing $B$ from 4 to 256 adds only $4.1$\,ms for full attention and
  $2.7$\,ms for recurrent memory at the longest history. The relationship is not
  strictly monotonic because these differences are small relative to runtime
  variation and fixed-shape policy computation. Memory capacity therefore has a
  much smaller effect on latency than the mechanism used to process temporal
  history.

  \paragraph{Scaling with history horizon.}
  Full attention grows by $34$--$36$\,ms between 1 and 1,451 frames, corresponding
  to a fitted slope of approximately $24$--$25$\,ms per 1,000 frames. Depending
  on $B$, this represents a $31$--$34\%$ increase. The recurrent variants grow by
  only $10$--$12$\,ms, with slopes of approximately $6.6$--$9.7$\,ms per 1,000
  frames and relative increases of $8.8$--$10.4\%$.

  The difference is explained by the history representations. Full attention
  constructs an input whose active history grows from approximately 194 to 4,008
  tokens. At $B=256$, history preparation consequently increases from $1.1$ to
  $33.7$\,ms. In contrast, recurrent memory incrementally compresses completed
  history into a fixed-size state and reads only that state together with a short
  recent window. Its history-preparation cost increases from $1.0$ to only
  $8.5$\,ms. Action generation itself remains nearly constant at approximately
  $79$--$82$\,ms for both architectures.

  Full attention is slightly faster for very short histories because recurrent
  memory pays a fixed update and read overhead. The recurrent implementation
  becomes faster after approximately 250--550 frames, depending on $B$. At the
  longest history it reduces latency by $15.4$--$21.1$\,ms, or $10.7$--$14.2\%$,
  relative to full attention. These results show that recurrent compression
  provides its main computational advantage as temporal context grows, while
  adding only modest overhead over the non-memory policy.

\section{Additional Ablation Results}
\label{app:additional_experiments}

These comparisons isolate choices in the visual representation and in the historical input streams supplied to the memory encoder. Table~\ref{tab:ablation_definitions} defines the corresponding configurations.

\subsection{Ablations on Visual Encoder For Input Stream \& Supervision Objective}
\label{sec:visual-feature-ablation}

Frame-level visual features enter memory learning at two points: the recent
visual stream, which covers frames not yet encoded by the video tokenizer,
and the Stage~1 visual reconstruction target. We compare two feature choices
that change both points together. The \emph{DINOv3} variant uses the same
DINOv3 features for the recent stream and the reconstruction target, so the
encoder receives the same representation that the decoder learns to predict.
Each image is encoded by a frozen DINOv3 ViT-B/16 at $256\times256$
resolution, and its patch grid is average-pooled to 64 tokens of width 768
per camera view. The \emph{SigLIP} variant instead takes the recent stream
from the SigLIP vision encoder of the policy($\pi_{0.5}$) and uses the image-token hidden
states of its PaliGemma language model as the reconstruction target, each
pooled to 64 tokens of width 2048 per view. Unlike the DINOv3 variant, the
input and target are therefore different representations. All other
components, including the memory bottleneck, the action and proprioceptive
streams, the training objectives, and the evaluation protocol, are held
fixed. We do not evaluate mixed configurations, such as DINOv3 inputs with
SigLIP targets.

\paragraph{Results.}
Table~\ref{tab:visual-feature-ablation} shows that \textsc{Imprint} performs well with both feature choices, with a small overall advantage for DINOv3. On RoboCasa, DINOv3 improves memory-task success from 57.7\% to 61.0\%, while both variants reach 63.3\% on non-memory tasks, giving a 1.7-point gain in overall success (62.2\% vs.\ 60.5\%). On RMBench, DINOv3 improves average success by 2.1 points (72.7\% vs.\ 70.6\%). The largest gains occur on Sweeten Hot Chocolate ($+10$) and Wash and Return to Same Location ($+8$) on RoboCasa, and on Block Ranking ($+11$) and Battery ($+9$) on RMBench. These tasks require tracking changes in object arrangement or state, which is consistent with dense patch features being better suited to capturing local visual changes. SigLIP is better on a few tasks, most notably Press Button ($-7$) and Observe and Pick Up ($-3$) on RMBench and Rinse Cutting Board ($-4$) on RoboCasa, so the advantage is not uniform across tasks. Because the two variants change the recent stream and the reconstruction target together, this comparison measures their combined effect and does not isolate the contribution of either one. Since DINOv3 performs better on average and does not depend on the policy's visual encoder, we use it as the default for all experiments in Sec.~\ref{sec:experiments}.

\begin{table*}[t]
    \centering
    \caption{Effect of the visual features used for both the recent visual
    stream and the Stage~1 reconstruction target. \emph{DINOv3} uses pooled
    DINOv3 patch features for both; \emph{SigLIP} uses pooled SigLIP features
    as input and PaliGemma image-token hidden states as the target. Entries
    are per-task success rates (\%), and $\Delta$ is DINOv3 minus SigLIP. The
    better result in each row is shown in bold, and ties are bolded for both
    variants.}
    \label{tab:visual-feature-ablation}
    \setlength{\tabcolsep}{4.5pt}
    \begin{minipage}[t]{0.56\textwidth}
        \centering
        \small
        \textbf{(a) RoboCasa}\\[3pt]
        \begin{tabular}{lrrr}
            \toprule
            Task & DINOv3 & SigLIP & $\Delta$ \\
            \midrule
            \multicolumn{4}{l}{\textit{memory}} \\
            Retrieve fruit                    & \textbf{84} & 78          & $+6$ \\
            Retrieve oil                      & \textbf{76} & \textbf{76} & $0$ \\
            Wash and return                   & \textbf{86} & \textbf{86} & $0$ \\
            Wash and return to same location  & \textbf{62} & 54          & $+8$ \\
            Rinse cutting board               & 0           & \textbf{4}  & $-4$ \\
            Sweeten hot chocolate             & \textbf{58} & 48          & $+10$ \\
            \addlinespace
            \textit{Average}                  & \textbf{61.0} & 57.7      & $+3.3$ \\
            \midrule
            \multicolumn{4}{l}{\textit{Non-memory}} \\
            Pan transfer                      & \textbf{76} & \textbf{76} & $0$ \\
            Frying-pan adjustment             & 72          & \textbf{74} & $-2$ \\
            Organize condiments               & \textbf{62} & 60          & $+2$ \\
            Place equal ice cubes             & 6           & \textbf{8}  & $-2$ \\
            Scrub cutting board               & 66          & \textbf{68} & $-2$ \\
            Counter to sink                   & \textbf{98} & 94          & $+4$ \\
            \addlinespace
            \textit{Average}                  & \textbf{63.3} & \textbf{63.3} & $0$ \\
            \midrule
            Overall average                   & \textbf{62.2} & 60.5      & $+1.7$ \\
            \bottomrule
        \end{tabular}
    \end{minipage}
    \hfill
    \begin{minipage}[t]{0.40\textwidth}
        \centering
        \small
        \textbf{(b) RMBench}\\[3pt]
        \begin{tabular}{lrrr}
            \toprule
            Task & DINOv3 & SigLIP & \(\Delta\) \\
            \midrule
            Observe/pickup              & 9            & \textbf{12}  & \(-3\) \\
            Rearrange blocks            & \textbf{100} & 98           & \(+2\) \\
            Put back block              & \textbf{100} & \textbf{100} & \(0\) \\
            Swap blocks                 & \textbf{100} & 93           & \(+7\) \\
            Swap T                      & \textbf{100} & \textbf{100} & \(0\) \\
            Battery                     & \textbf{41}  & 32           & \(+9\) \\
            Blocks ranking              & \textbf{83}           & 72  & \(+11\) \\
            Cover blocks                & \textbf{100} & \textbf{100} & \(0\) \\
            Press button                & 21  & \textbf{28}           & \(-7\) \\
            \midrule
            \textbf{Overall average}    & \textbf{72.7} & 70.6        & \(+2.1\) \\
            \bottomrule
        \end{tabular}
    \end{minipage}
\end{table*}

\subsection{Ablation on Historical Input Streams}
\label{app:input_stream_ablation}

The Stage-1 objective ablation in Sec.~\ref{sec:exp_learning} removes \emph{supervision} for individual streams while still encoding all three streams. Here we instead remove the \emph{inputs} themselves, to measure how much each history stream contributes to memory. We fix the full-attention encoder at $B=256$ and keep visual history, recent DINOv3 features, and language conditioning in all variants. \emph{Vision only} removes both action and proprioceptive inputs and their reconstruction objectives. \emph{Vision + Action} retains executed actions and action reconstruction but removes proprioception. \emph{Vision + State} retains proprioception and state reconstruction but removes actions. Disabled streams are excluded from the encoder's key--value sequence, and their reconstruction losses are not computed. All other settings match the default configuration.

\paragraph{Results.}
Table~\ref{tab:input_stream_ablation} shows that both non-visual streams contribute, and that the full combination performs best (72.7\%). Among the ablated variants, Vision + Action is strongest at 68.4\%, compared with 60.6\% for Vision only and 60.9\% for Vision + State. Relative to visual history alone, adding actions improves Swap T (45\% to 99\%), Swap Blocks (81\% to 100\%), Press Button (7\% to 18\%), and Battery (30\% to 38\%), tasks in which the relevant history consists of interaction events such as which blocks have been moved or how many presses have occurred. These events are difficult to infer from temporally sampled visual observations alone.

Adding proprioception without actions leaves the average nearly unchanged, but this masks large offsetting per-task effects. Vision + State matches Vision + Action on Swap T (99\%) and nearly on Swap Blocks (98\%), yet sharply degrades Block Ranking (80\% to 46\%) and Cover Blocks (100\% to 70\%). One possible explanation is that proprioceptive history alone offers an easy but incomplete cue, which the memory may rely on at the expense of visual information; this is consistent with the E2E variant in Fig.~\ref{fig:memory_bottleneck_analysis} attending mainly to state history on Cover Blocks. In contrast, combining state with actions and vision improves over Vision + Action by 4.3 points, with the largest gains on Cover Blocks (83\% to 100\%) and Block Ranking (72\% to 83\%). This suggests that the streams are complementary when supervised jointly.

Finally, removing streams from the input is more harmful than removing only their supervision. Encoding only visual history yields 60.6\%, whereas encoding all three streams with visual reconstruction only (Table~\ref{tab:learning_per_task}) yields 66.9\%, with Swap T rising from 45\% to 99\%. Access to action and proprioceptive history therefore helps even without explicit reconstruction targets for those streams, and supervising them provides a further gain.

\begin{table}[t]
\centering
\caption{Historical input-stream ablation on RMBench with the full-attention encoder at $B=256$. Each variant removes the listed streams from both the memory input and the Stage-1 reconstruction targets. The best result in each column is shown in bold; columns where all variants tie are left unmarked.}
\label{tab:input_stream_ablation}
\resizebox{\linewidth}{!}{%
\begin{tabular}{lcccccccccc}
\toprule
Historical inputs & Obs./pick & Rearrange & Put back & Swap blocks & Swap T & Battery & Ranking & Cover & Press & Avg. \\
\midrule
Vision only          & \textbf{9} & 93  & 100 & 81  & 45  & 30 & 80 & \textbf{100} & 7  & 60.6 \\
Vision + Action      & 6 & \textbf{100} & 100 & \textbf{100} & 99  & 38 & 72 & 83  & 18 & 68.4 \\
Vision + State       & \textbf{9} & 94  & 100 & 98  & 99  & 28 & 46 & 70  & 4  & 60.9 \\
\midrule
Vision + Action + State (\textsc{Imprint}) & \textbf{9} & \textbf{100} & 100 & \textbf{100} & \textbf{100} & \textbf{41} & \textbf{83} & \textbf{100} & \textbf{21} & \textbf{72.7} \\
\bottomrule
\end{tabular}}
\end{table}

\section{Per-Task Failure Analysis}

\label{app:failure_analysis}

We analyze representative rollouts to examine errors in task-state tracking
and physical execution. Memory-related errors involve incorrect target
identities, action orders, repetition counts, or remembered locations and
orientations. Manipulation errors involve failed grasps, dropped objects,
or inaccurate placements despite selecting the intended target. All
comparisons below refer to the illustrated rollouts.

\subsection{RMBench Analysis}
\label{app:qualitative_rmbench}

For each task, we compare \ours with HAMLET and Uniform Sampling, followed
by a comparison with the Frozen and End-to-End training variants. Together,
these examples illustrate challenges in recalling task-relevant information
and executing the corresponding actions precisely.

\begin{figure}[tbp]
    \centering
    \includegraphics[width=0.9\linewidth]{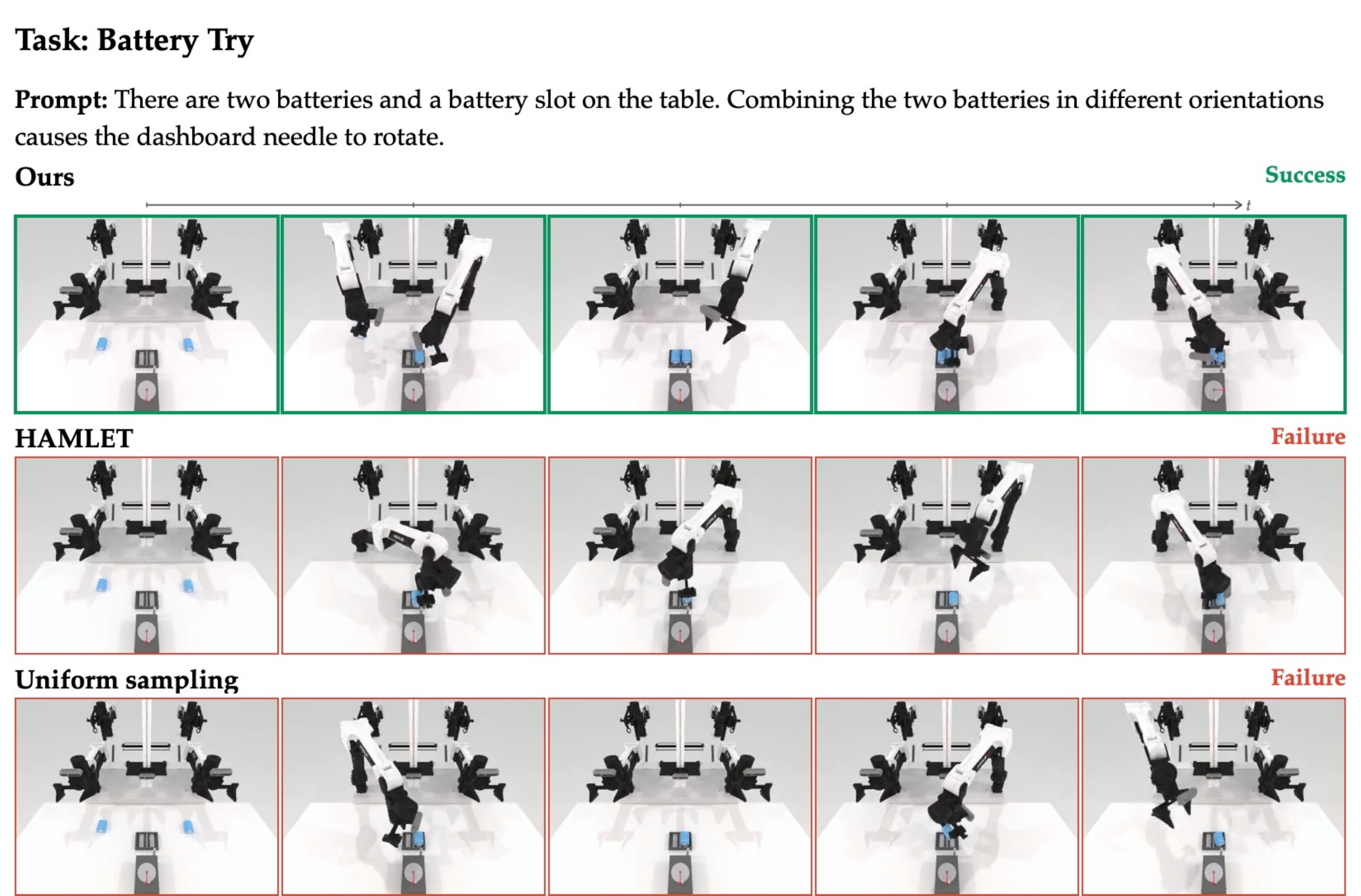}
    \caption{\textbf{RMBench: Battery Try.} Representative rollouts of \ours, HAMLET, and Uniform Sampling.}
    \label{fig:rmbench_baselines_battery_try}
\end{figure}

\begin{figure}[tbp]
    \centering
    \includegraphics[width=0.9\linewidth]{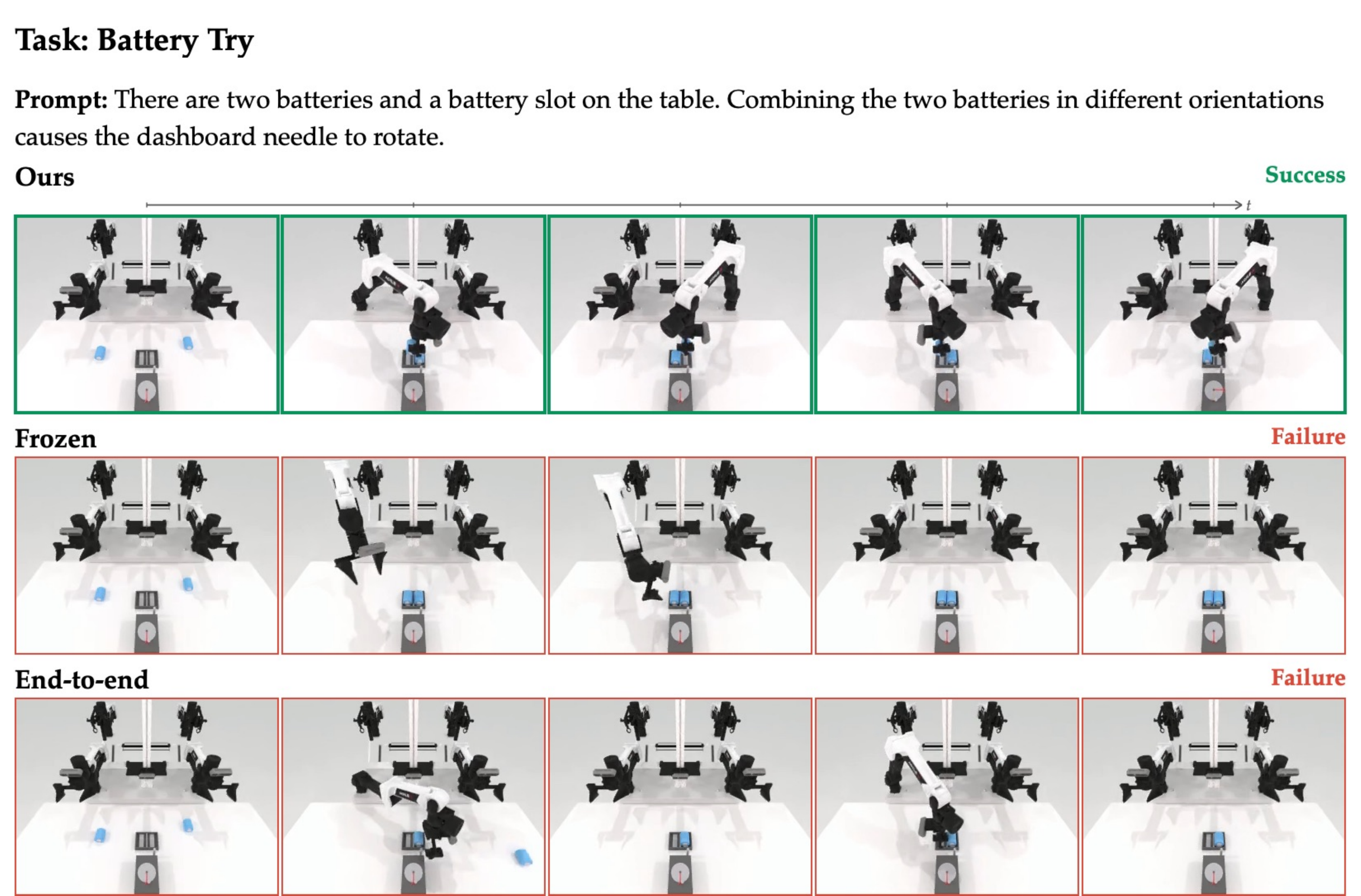}
    \caption{\textbf{RMBench: Battery Try.} Representative rollouts of \ours and the Frozen and End-to-End training variants.}
    \label{fig:rmbench_ablation_battery_try}
\end{figure}

\paragraph{Battery Try.}
In the baseline comparison
(Fig.~\ref{fig:rmbench_baselines_battery_try}), \ours completes the task.
Both baselines insert one battery but knock the remaining battery off the
table during grasping. The training-recipe comparison
(Fig.~\ref{fig:rmbench_ablation_battery_try}) reveals additional failure
modes: Frozen completes an insertion but stalls before the required
rotation, while End-to-End pushes a battery away and cannot complete the
task. In contrast, \ours inserts both batteries and performs the required
reorientation. These examples illustrate the need for both correct action
sequencing and precise manipulation.


\begin{figure}[tbp]
    \centering
    \includegraphics[width=0.9\linewidth]{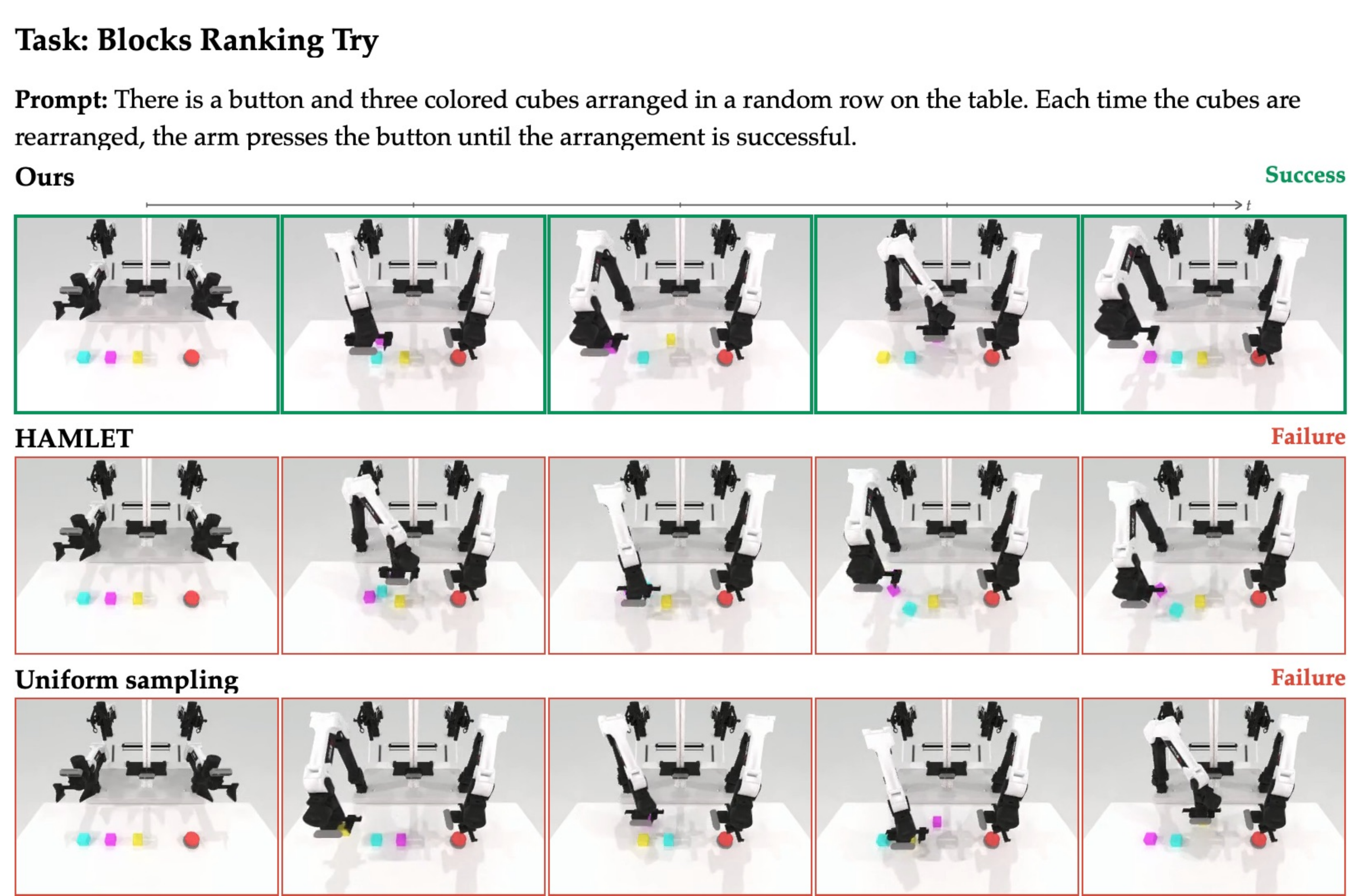}
    \caption{\textbf{RMBench: Block Ranking Try.} Representative rollouts of \ours, HAMLET, and Uniform Sampling.}
    \label{fig:rmbench_baselines_block_ranking_try}
\end{figure}

\begin{figure}[tbp]
    \centering
    \includegraphics[width=0.9\linewidth]{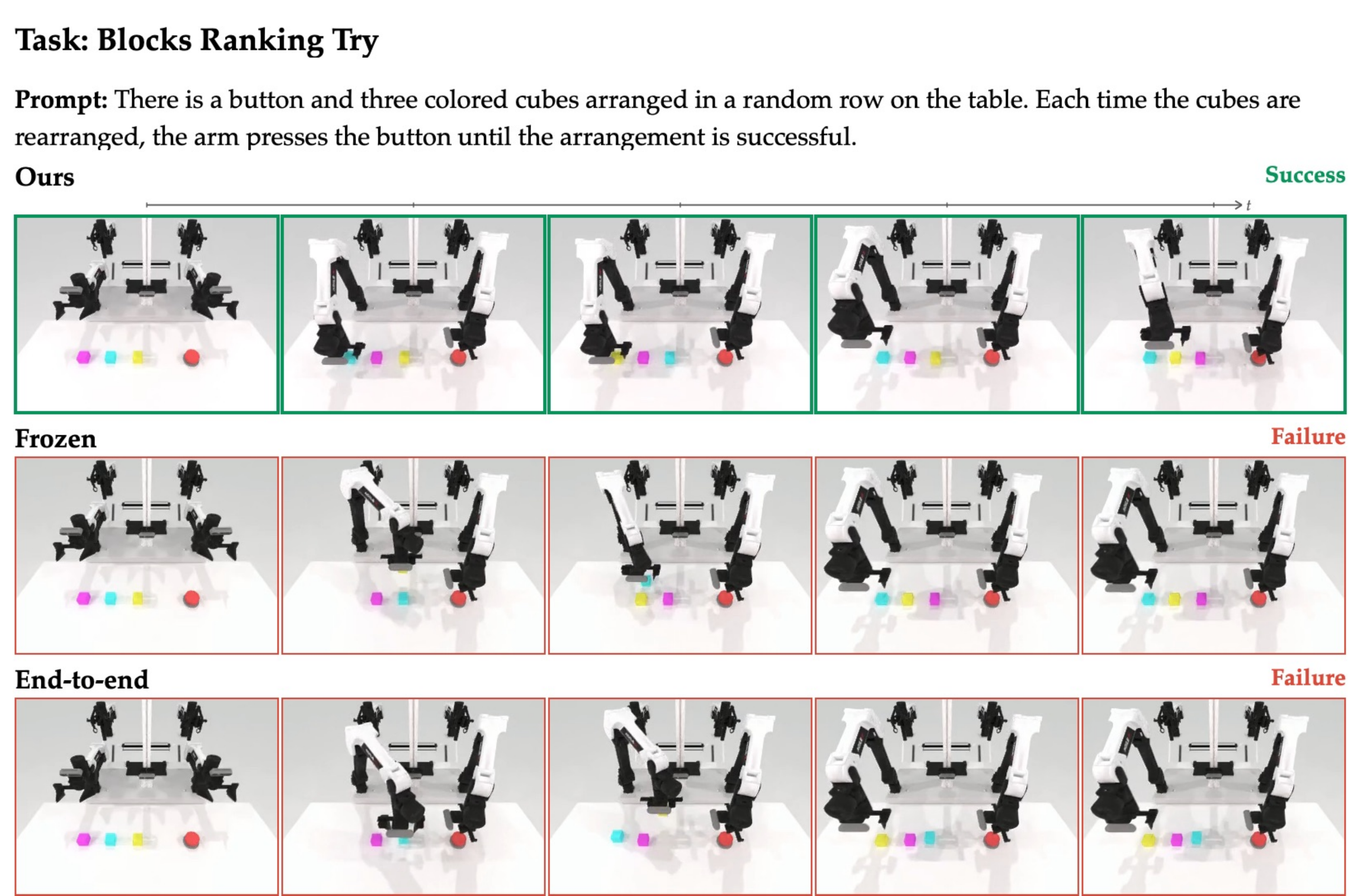}
    \caption{\textbf{RMBench: Block Ranking Try.} Representative rollouts of \ours and the Frozen and End-to-End training variants.}
    \label{fig:rmbench_ablation_block_ranking_try}
\end{figure}

\paragraph{Block Ranking Try.}
After repeated rearrangement, \ours places the blocks in the target
configuration (Fig.~\ref{fig:rmbench_baselines_block_ranking_try}). The
baselines either exceed the time limit or place blocks inaccurately. In
the training-recipe comparison
(Fig.~\ref{fig:rmbench_ablation_block_ranking_try}), Frozen presses the
button prematurely, while End-to-End fails to reconstruct the target
configuration. Only \ours completes the illustrated task, which requires
tracking changes in the arrangement and accurately reproducing the
resulting configuration.


\begin{figure}[tbp]
    \centering
    \includegraphics[width=0.9\linewidth]{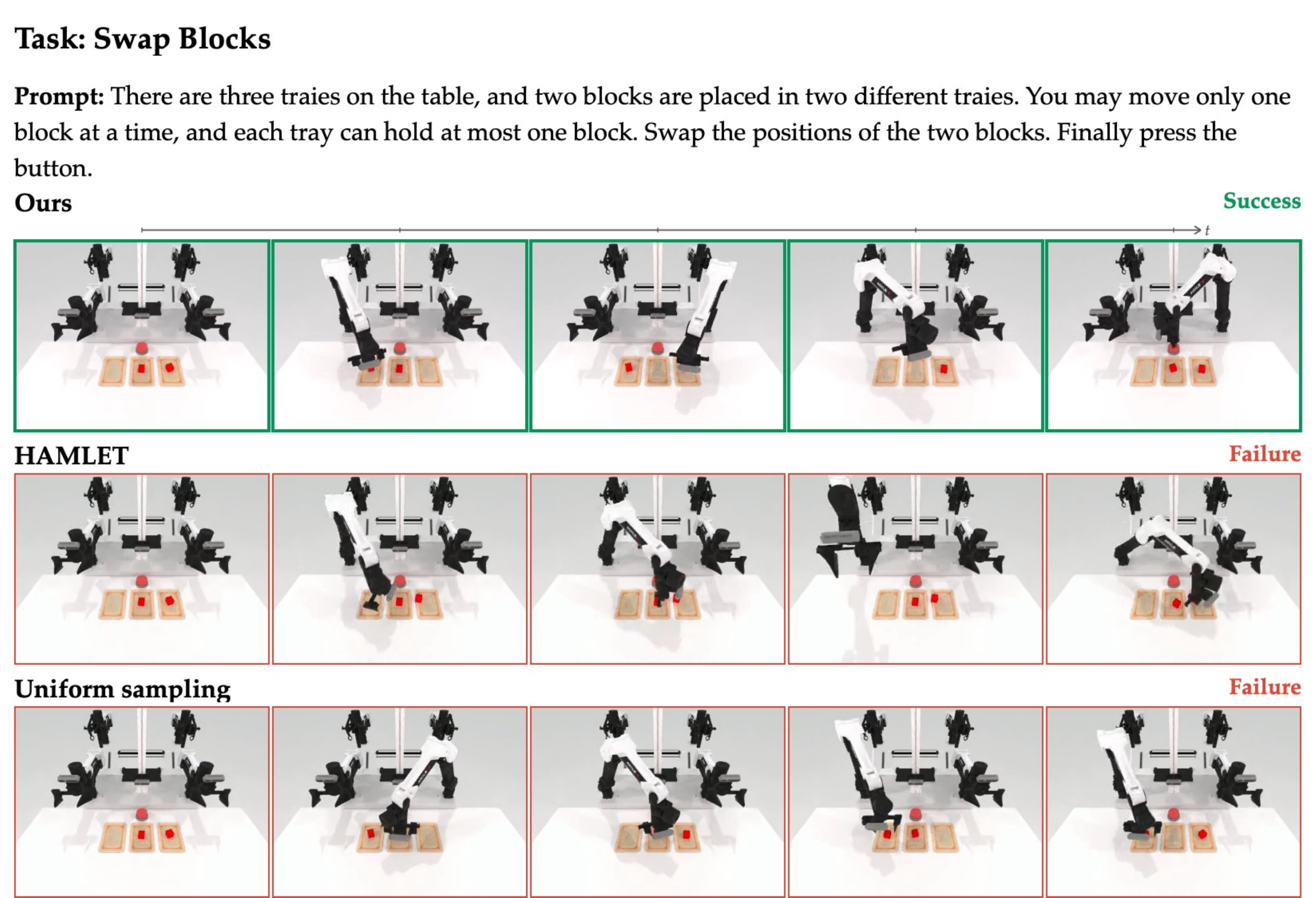}
    \caption{\textbf{RMBench: Swap Blocks.} Representative rollouts of \ours, HAMLET, and Uniform Sampling.}
    \label{fig:rmbench_baselines_swap_block}
\end{figure}

\begin{figure}[tbp]
    \centering
    \includegraphics[width=0.9\linewidth]{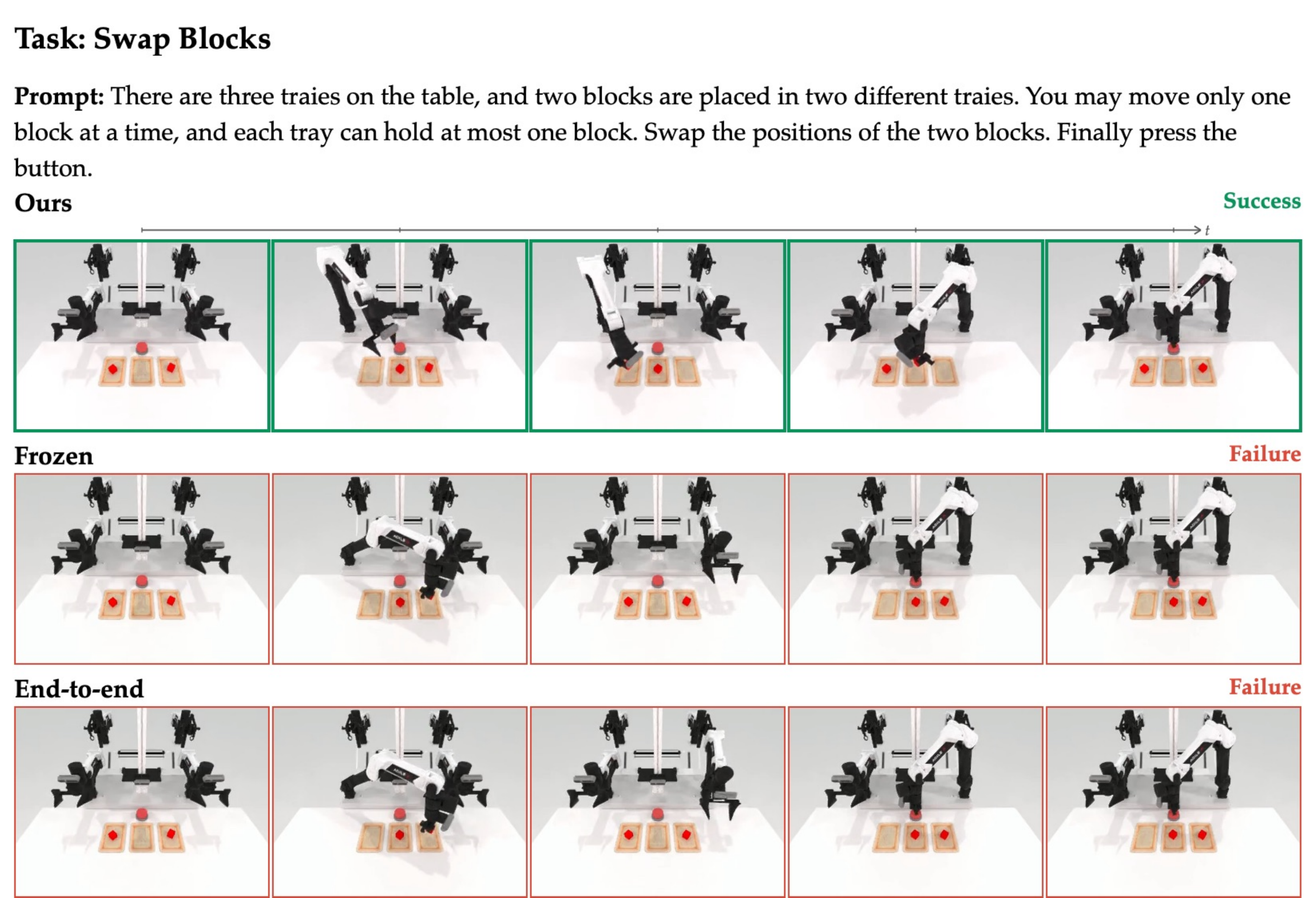}
    \caption{\textbf{RMBench: Swap Blocks.} Representative rollouts of \ours and the Frozen and End-to-End training variants.}
    \label{fig:rmbench_ablation_swap_block}
\end{figure}

\paragraph{Swap Blocks.}
As shown in Fig.~\ref{fig:rmbench_baselines_swap_block}, \ours uses the
vacant slot on the left as a temporary buffer and swaps the blocks into
the correct slots. HAMLET makes an imprecise placement, while Uniform
Sampling follows an incorrect action sequence and exceeds the time limit.
In the training-recipe comparison
(Fig.~\ref{fig:rmbench_ablation_swap_block}), both Frozen and End-to-End
fail to place the blocks correctly in the required slots, whereas \ours
completes the swap.


\begin{figure}[tbp]
    \centering
    \includegraphics[width=0.9\linewidth]{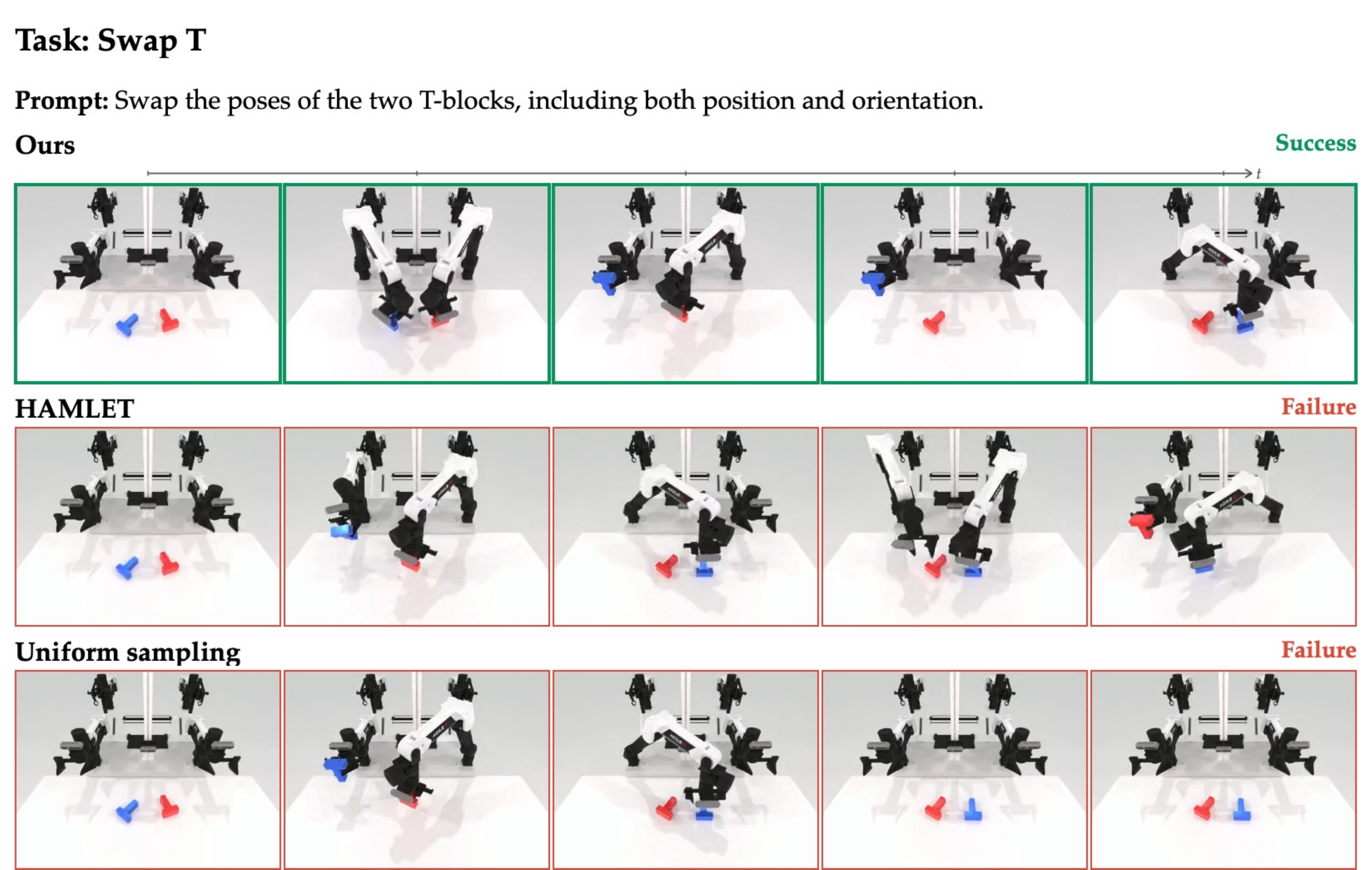}
    \caption{\textbf{RMBench: Swap T.} Representative rollouts of \ours, HAMLET, and Uniform Sampling.}
    \label{fig:rmbench_baselines_swap_t}
\end{figure}

\begin{figure}[tbp]
    \centering
    \includegraphics[width=0.9\linewidth]{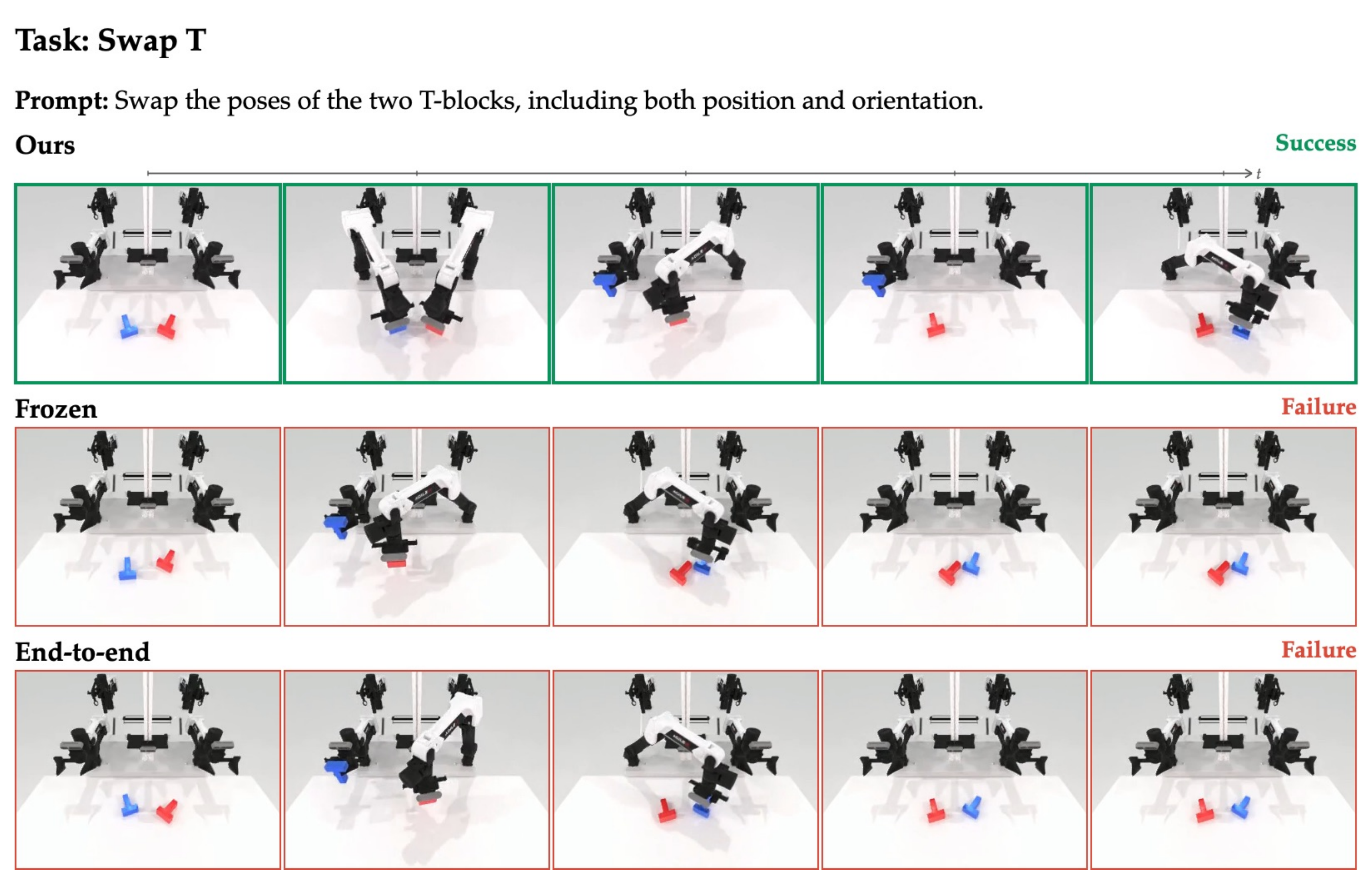}
    \caption{\textbf{RMBench: Swap T.} Representative rollouts of \ours and the Frozen and End-to-End training variants.}
    \label{fig:rmbench_ablation_swap_t}
\end{figure}

\paragraph{Swap T.}
In Fig.~\ref{fig:rmbench_baselines_swap_t}, \ours swaps the T-shaped blocks
and places them within the required position and orientation tolerances.
Both baselines exchange the blocks but fail to achieve sufficiently precise
final placements. In Fig.~\ref{fig:rmbench_ablation_swap_t}, Frozen swaps
the blocks but leaves the blue block at an incorrect position and
orientation, while End-to-End fails to complete the swap. \ours completes
both the exchange and the precise final placement.


\begin{figure}[tbp]
    \centering
    \includegraphics[width=0.9\linewidth]{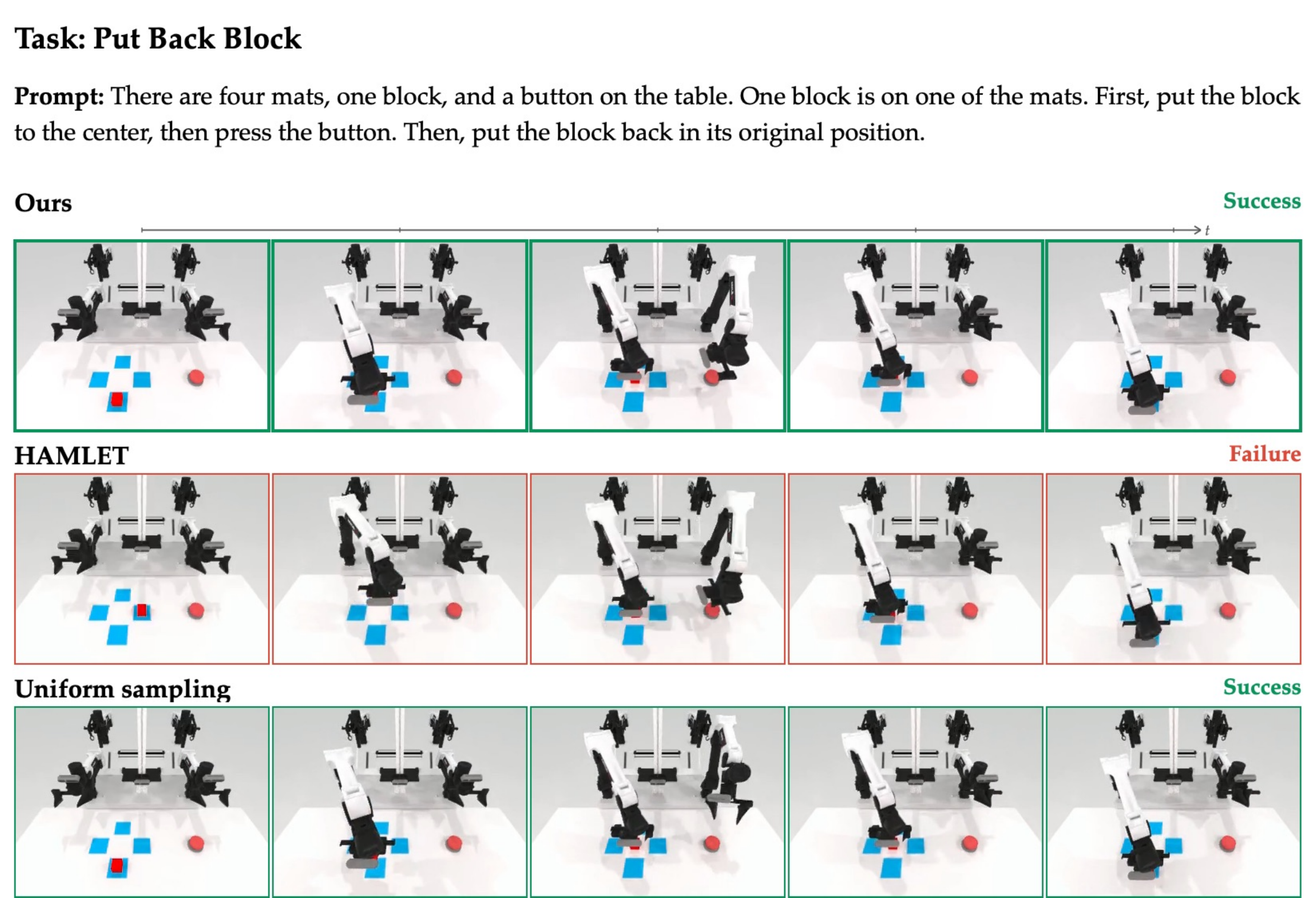}
    \caption{\textbf{RMBench: Put Back Block.} Representative rollouts of \ours, HAMLET, and Uniform Sampling.}
    \label{fig:rmbench_baselines_put_back_block}
\end{figure}

\begin{figure}[tbp]
    \centering
    \includegraphics[width=0.9\linewidth]{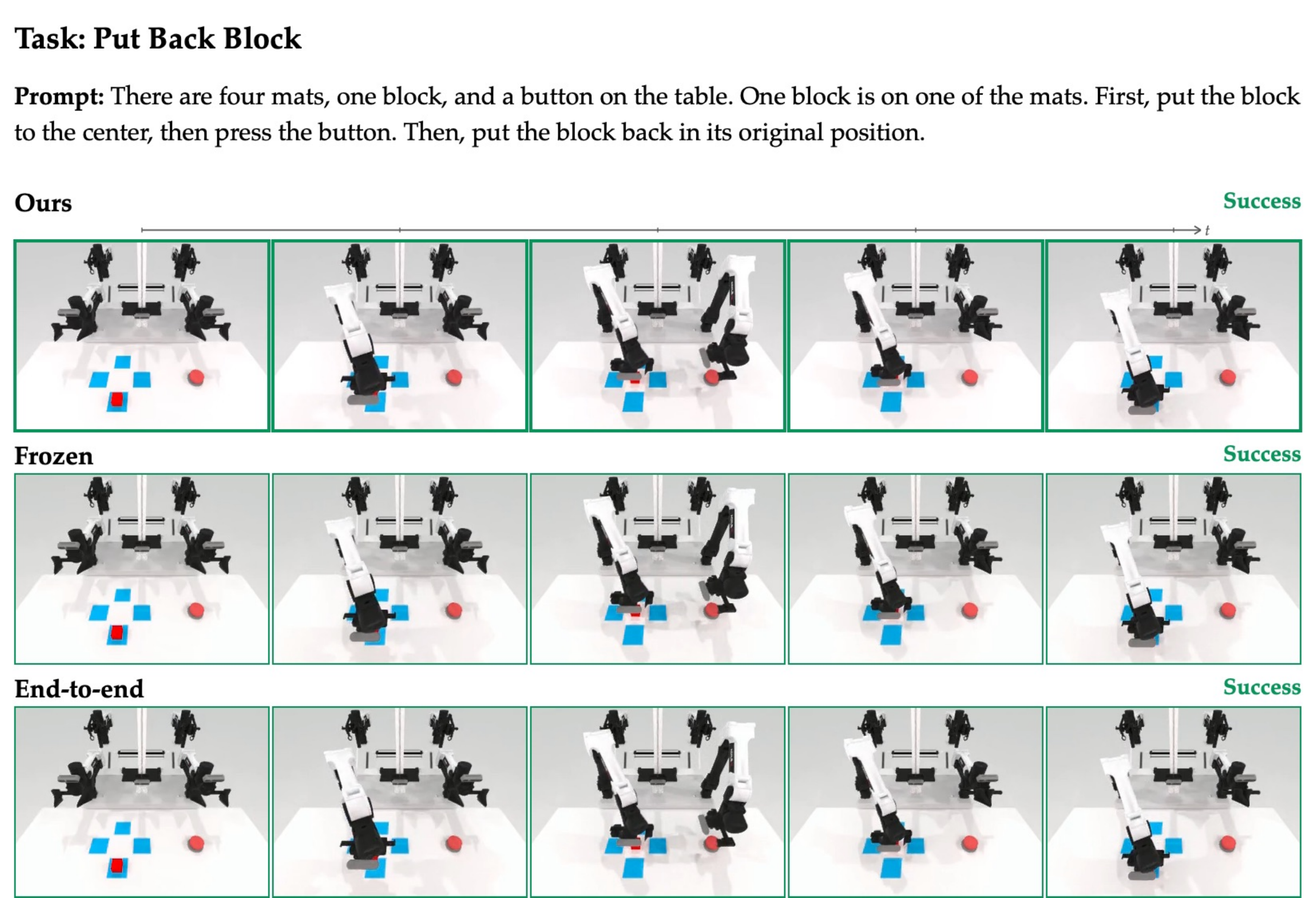}
    \caption{\textbf{RMBench: Put Back Block.} Representative rollouts of \ours and the Frozen and End-to-End training variants.}
    \label{fig:rmbench_ablation_put_back_block}
\end{figure}

\paragraph{Put Back Block.}
In the baseline comparison
(Fig.~\ref{fig:rmbench_baselines_put_back_block}), both \ours and Uniform
Sampling return the block to the correct location, while HAMLET selects
the wrong location. In the training-recipe comparison
(Fig.~\ref{fig:rmbench_ablation_put_back_block}), all three variants return
the block correctly and press the button.


\begin{figure}[tbp]
    \centering
    \includegraphics[width=0.9\linewidth]{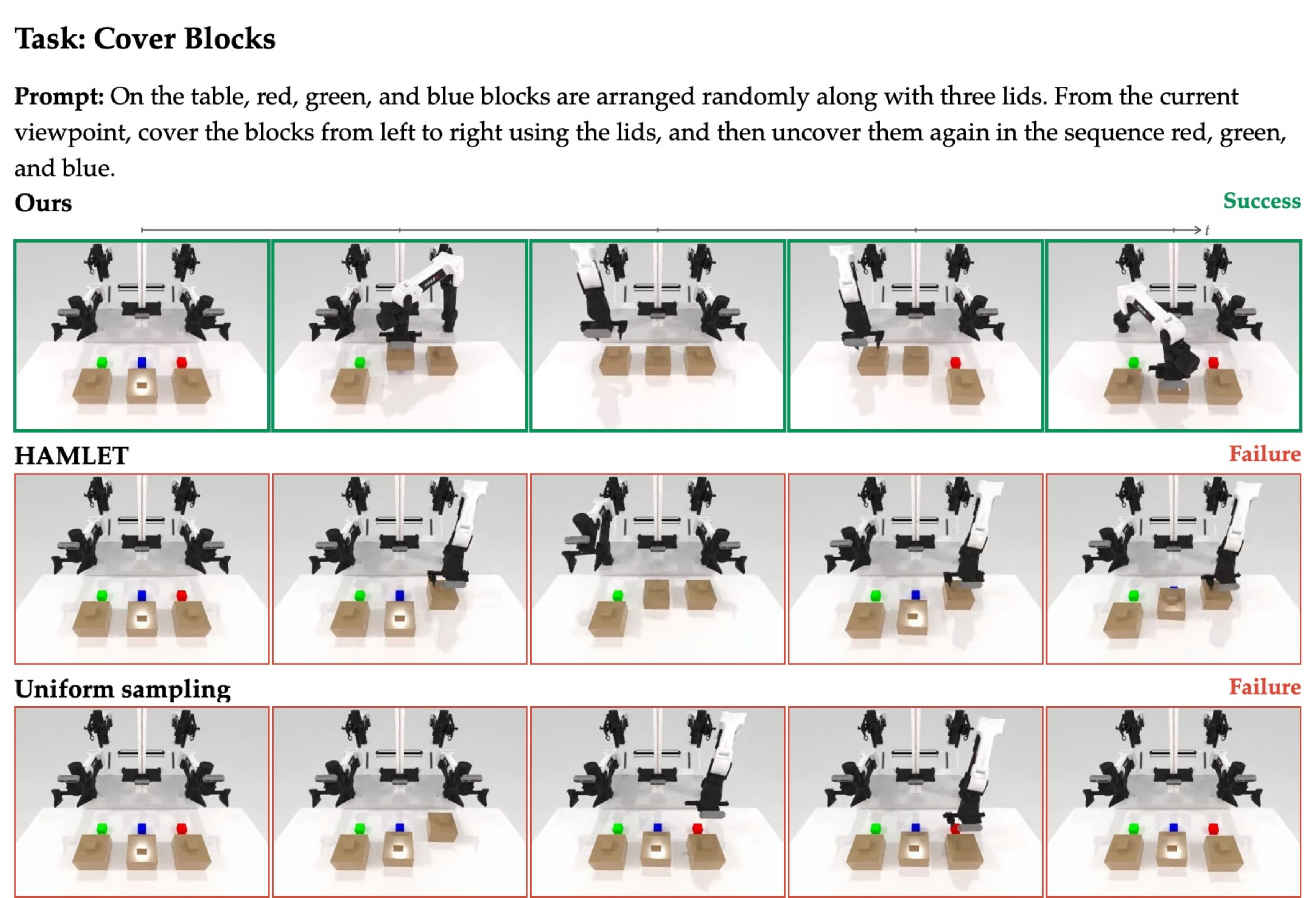}
    \caption{\textbf{RMBench: Cover Blocks.} Representative rollouts of \ours, HAMLET, and Uniform Sampling.}
    \label{fig:rmbench_baselines_Cover_block}
\end{figure}

\begin{figure}[tbp]
    \centering
    \includegraphics[width=0.9\linewidth]{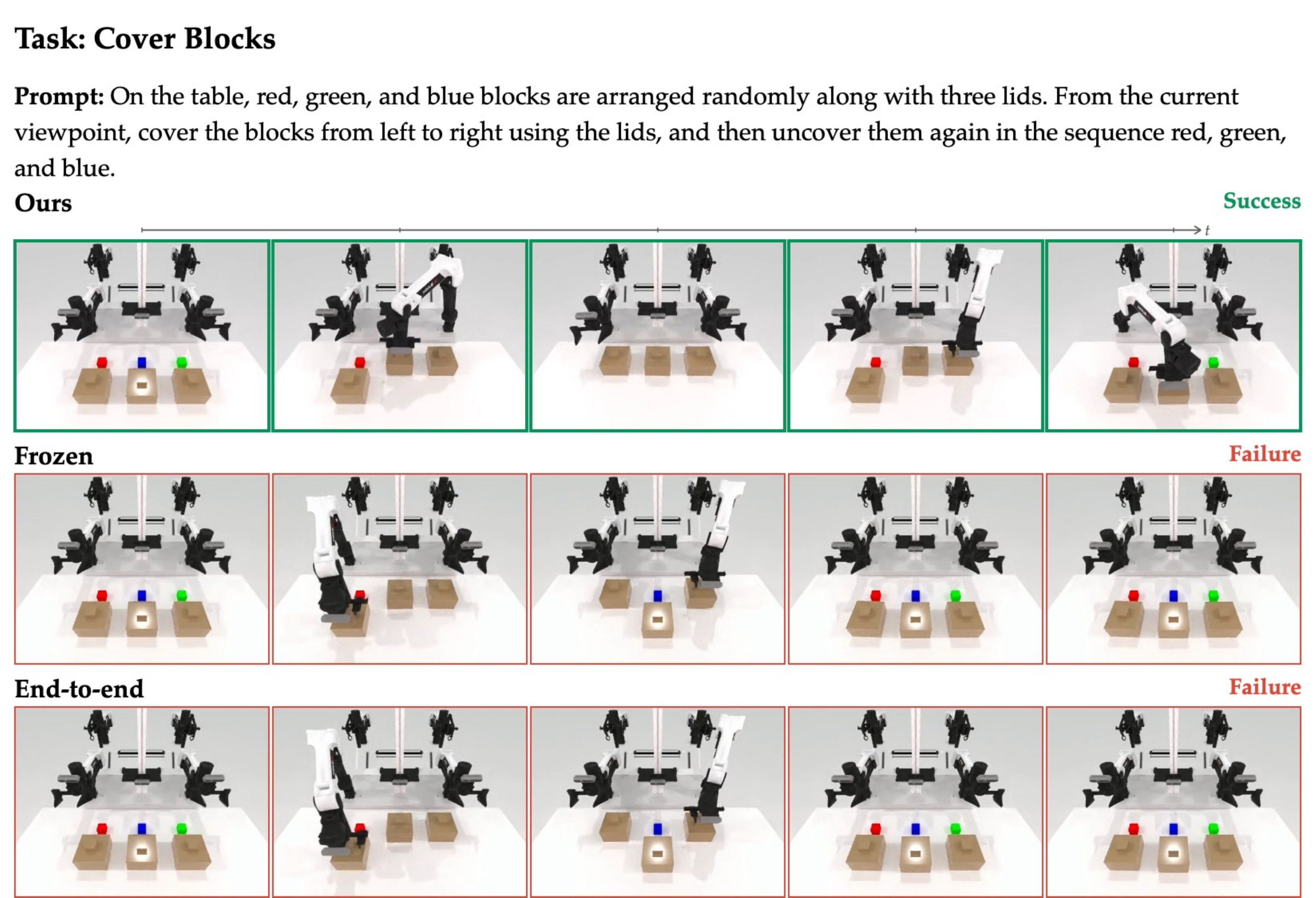}
    \caption{\textbf{RMBench: Cover Blocks.} Representative rollouts of \ours and the Frozen and End-to-End training variants.}
    \label{fig:rmbench_ablation_Cover_block}
\end{figure}

\paragraph{Cover Blocks.}
In Fig.~\ref{fig:rmbench_baselines_Cover_block}, \ours uncovers the blocks
in the required red--green--blue order. HAMLET instead follows a
green--blue--red order, while Uniform Sampling repeatedly covers and
uncovers the red block. In the training-recipe comparison
(Fig.~\ref{fig:rmbench_ablation_Cover_block}), both Frozen and End-to-End
incorrectly uncover the blue block first, whereas \ours follows the
required order. These rollouts illustrate difficulties in retaining the
order and locations of previously covered blocks.


\begin{figure}[tbp]
    \centering
    \includegraphics[width=0.9\linewidth]{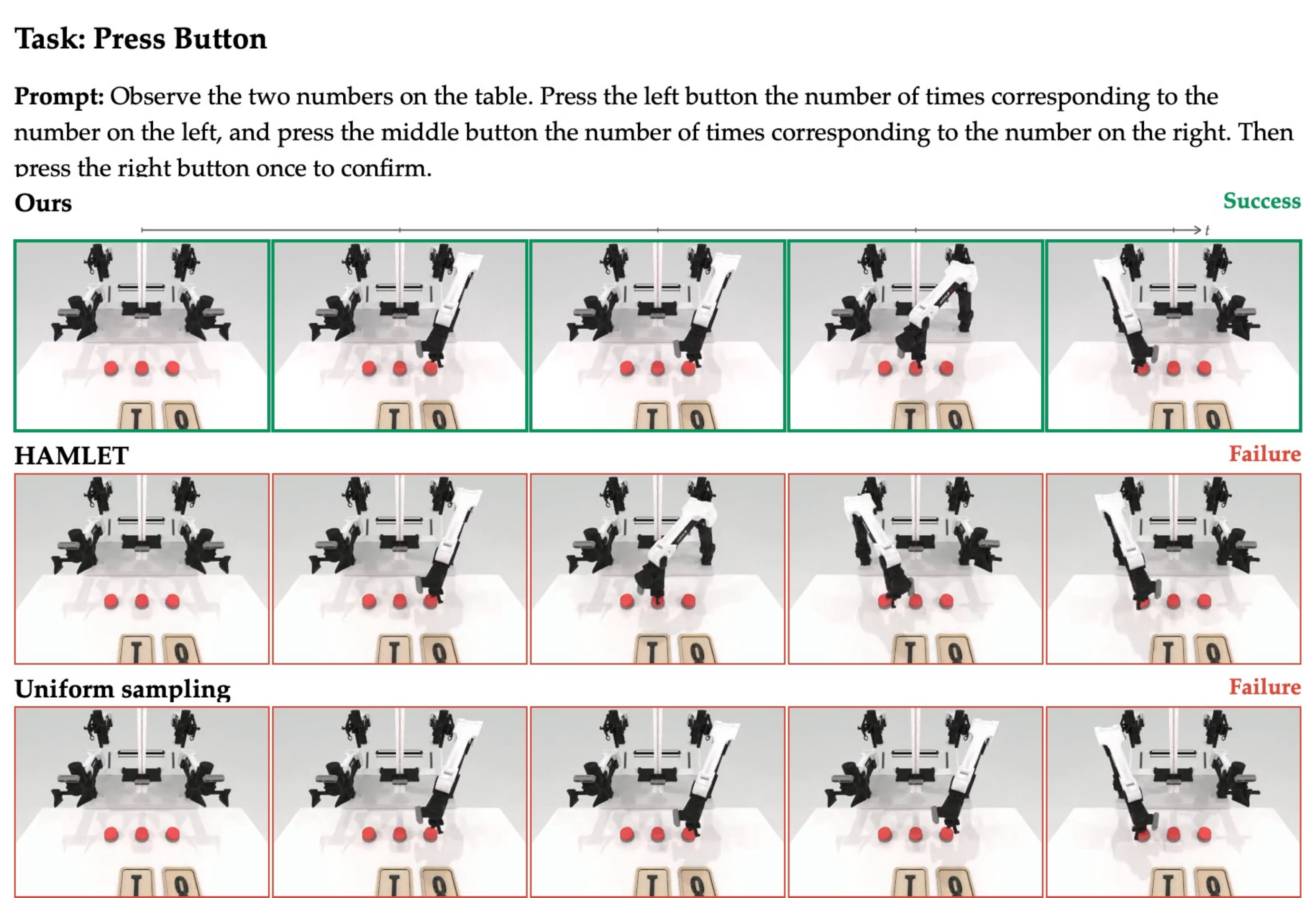}
    \caption{\textbf{RMBench: Press Button.} Representative rollouts of \ours, HAMLET, and Uniform Sampling.}
    \label{fig:rmbench_baselines_press_button}
\end{figure}

\begin{figure}[tbp]
    \centering
    \includegraphics[width=0.9\linewidth]{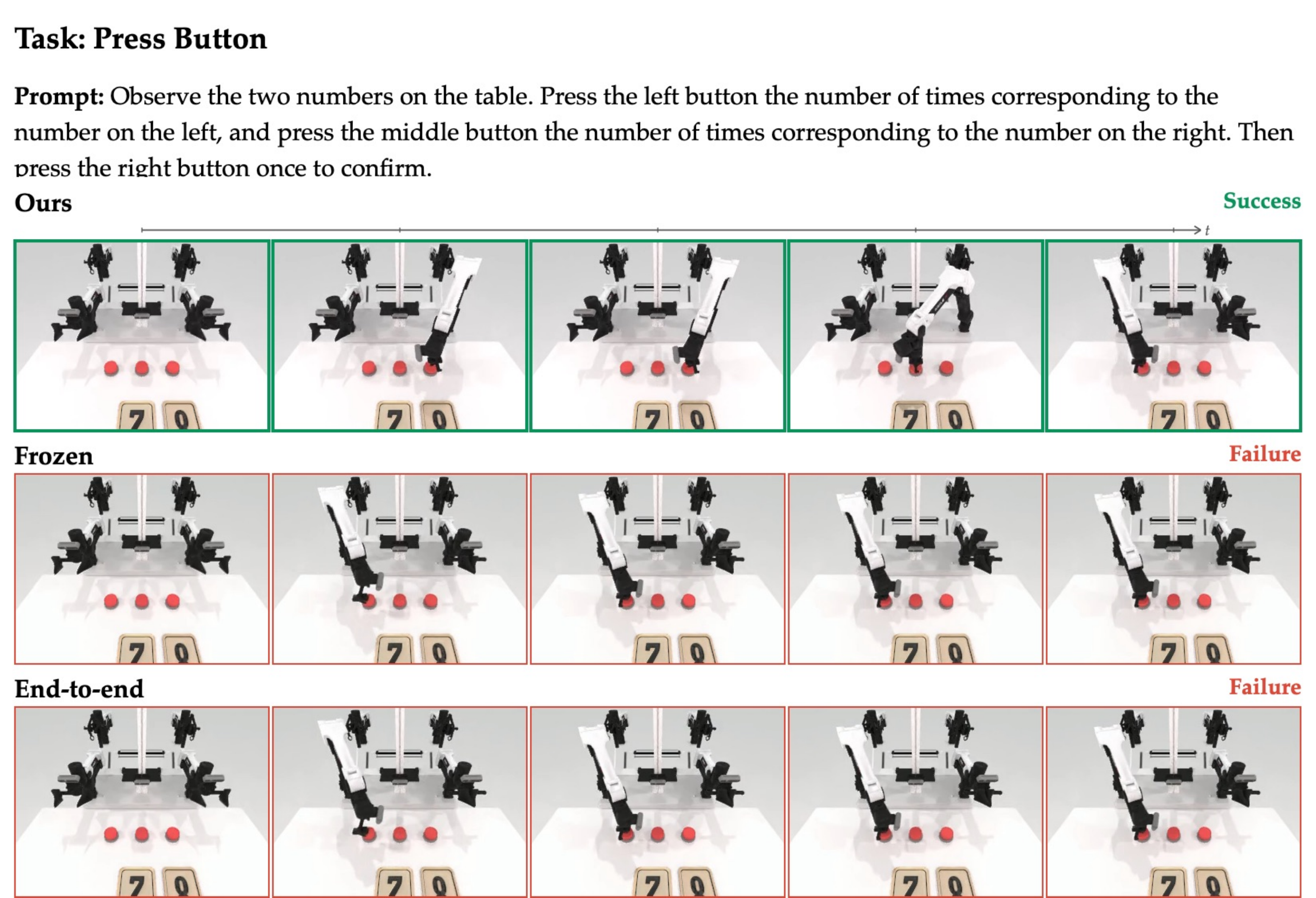}
    \caption{\textbf{RMBench: Press Button.} Representative rollouts of \ours and the Frozen and End-to-End training variants.}
    \label{fig:rmbench_ablation_press_button}
\end{figure}

\paragraph{Press Button.}
In Fig.~\ref{fig:rmbench_baselines_press_button}, \ours completes the
required eight presses on the right button and one press on the left.
Uniform Sampling presses the right button only once, and HAMLET presses
it only twice. In the separate training-recipe example
(Fig.~\ref{fig:rmbench_ablation_press_button}), \ours completes two presses
on the left button and eight on the right, while both ablations fail to
press the right button enough times. These examples illustrate errors in
tracking and completing the required repetition counts.


\begin{figure}[tbp]
    \centering
    \includegraphics[width=0.9\linewidth]{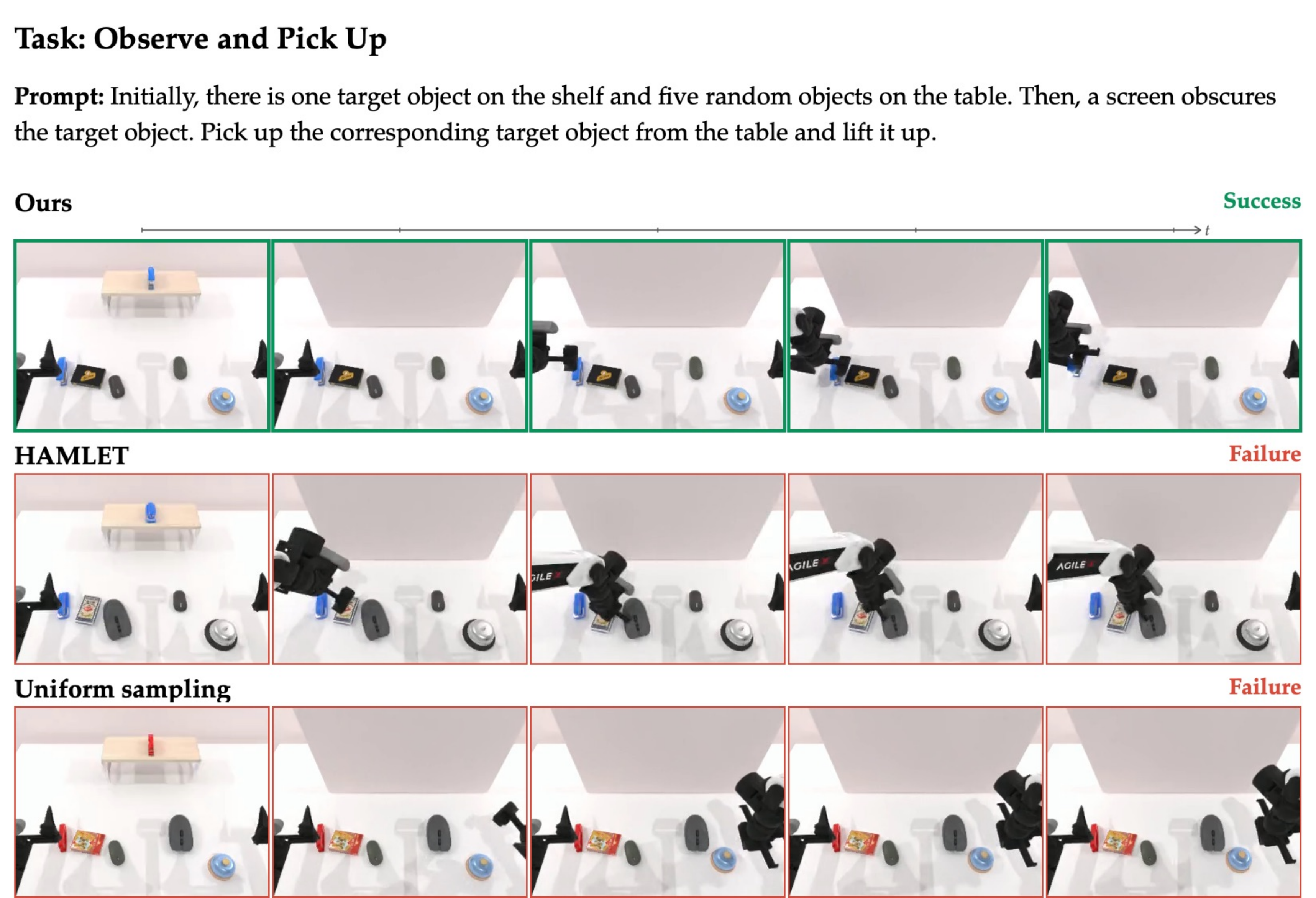}
    \caption{\textbf{RMBench: Observe and Pick Up.} Representative rollouts of \ours, HAMLET, and Uniform Sampling.}
    \label{fig:rmbench_baselines_observe_and_pickup}
\end{figure}

\begin{figure}[tbp]
    \centering
    \includegraphics[width=0.9\linewidth]{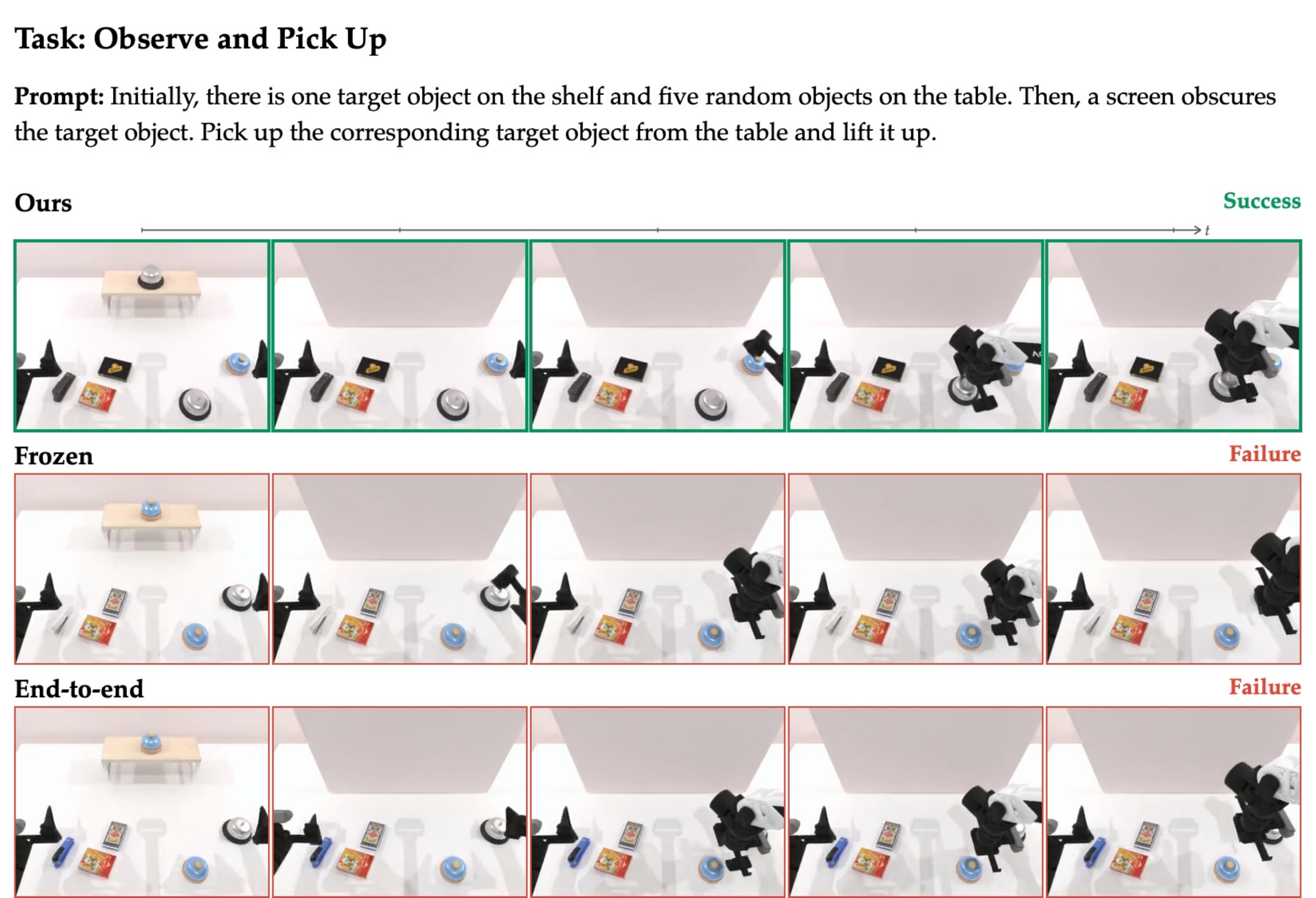}
    \caption{\textbf{RMBench: Observe and Pick Up.} Representative rollouts of \ours and the Frozen and End-to-End training variants.}
    \label{fig:rmbench_ablation_observe_and_pickup}
\end{figure}

\paragraph{Observe and Pick Up.}
In Fig.~\ref{fig:rmbench_baselines_observe_and_pickup}, \ours grasps the
target object shown during the initial observation, while both baselines
fail to locate the correct object. Similarly, in the training-recipe
comparison (Fig.~\ref{fig:rmbench_ablation_observe_and_pickup}), only \ours
successfully picks up the target object; Frozen and End-to-End fail to
retrieve it. These examples concern retaining the identity and location
of an object observed earlier in the episode.


\begin{figure}[tbp]
    \centering
    \includegraphics[width=0.9\linewidth]{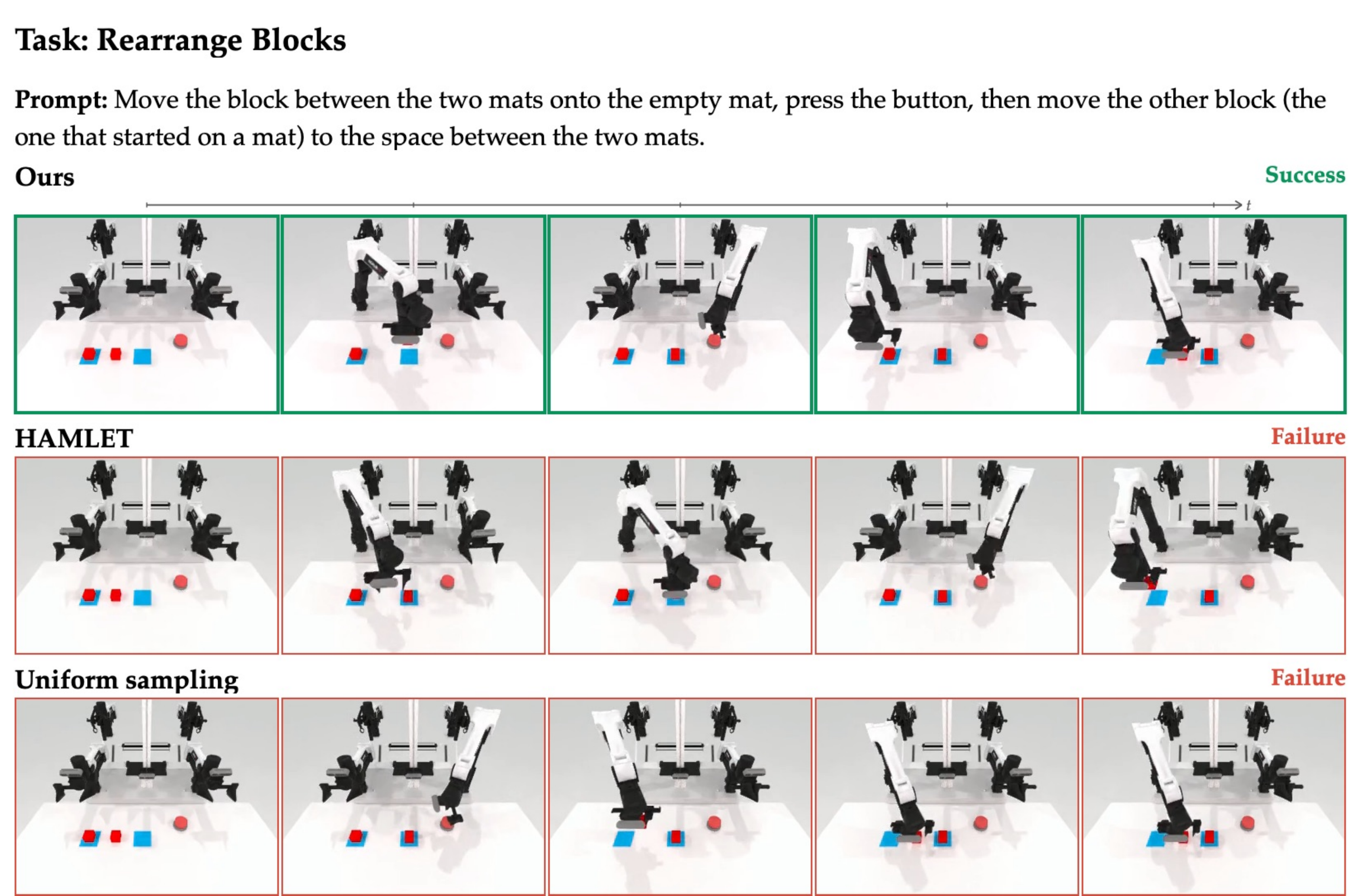}
    \caption{\textbf{RMBench: Rearrange Blocks.} Representative rollouts of \ours, HAMLET, and Uniform Sampling.}
    \label{fig:rmbench_baselines_rearrange_block}
\end{figure}

\begin{figure}[tbp]
    \centering
    \includegraphics[width=0.9\linewidth]{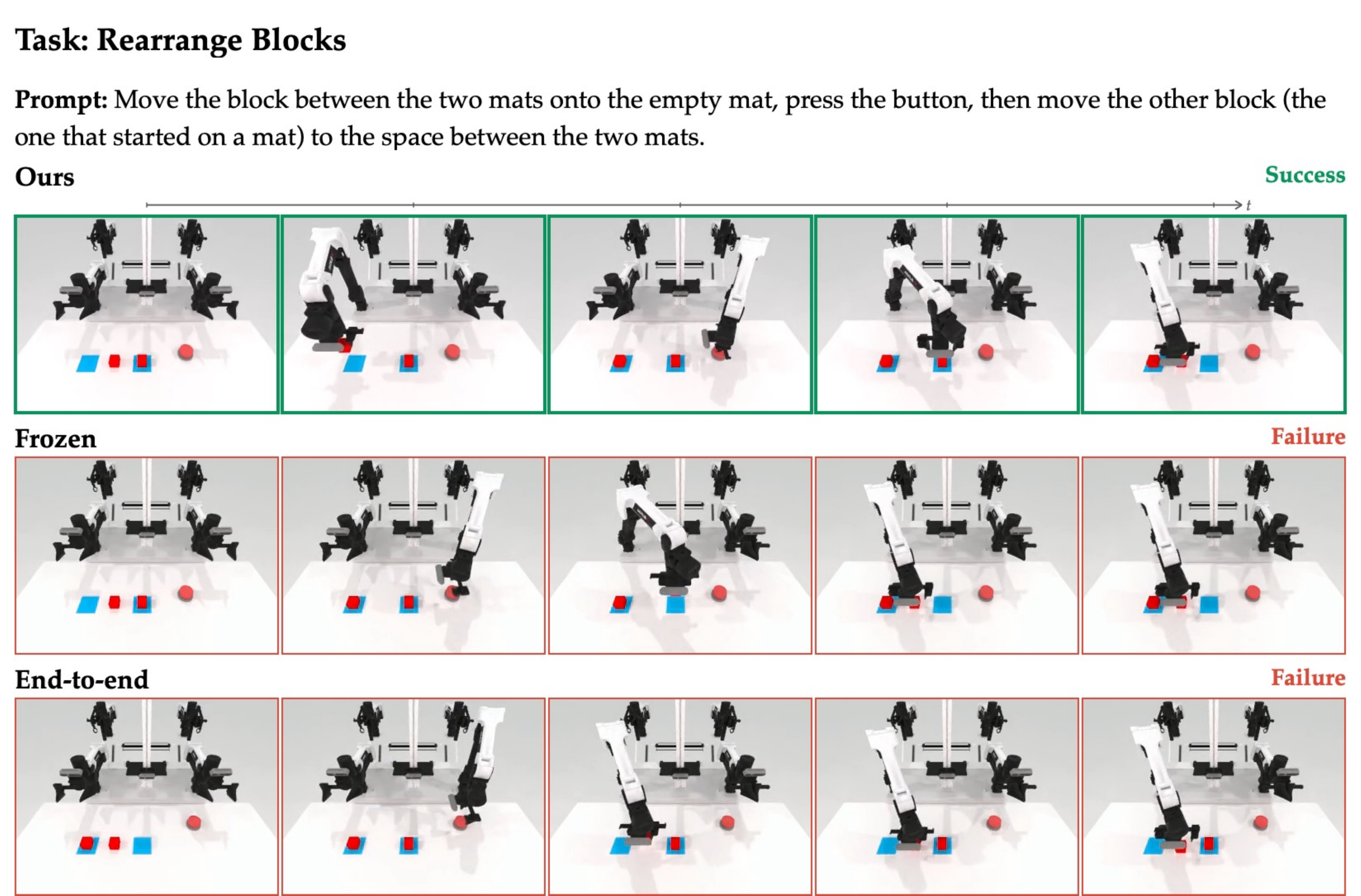}
    \caption{\textbf{RMBench: Rearrange Blocks.} Representative rollouts of \ours and the Frozen and End-to-End training variants.}
    \label{fig:rmbench_ablation_rearrange_block}
\end{figure}

\paragraph{Rearrange Blocks.}
In the baseline comparison
(Fig.~\ref{fig:rmbench_baselines_rearrange_block}), only \ours returns the
blocks to the target configuration and releases the gripper. Uniform Sampling baseline follows the correct sequence and places both blocks correctly. It fails because the right gripper remains closed around the second block instead of releasing. In the training-recipe comparison
(Fig.~\ref{fig:rmbench_ablation_rearrange_block}), Frozen approaches the
target arrangement but places a block inaccurately, while End-to-End fails
to achieve the required arrangement. \ours completes the rearrangement
with accurate final placements.


\subsection{RoboCasa Analysis}
\label{app:qualitative_robocasa}

We compare \ours with $\pi_{0.5}$ and two Uniform Sampling baselines that
use VLM-context or action-expert conditioning. The rollouts illustrate
errors in remembering locations, tracking counts and task progress, and
executing precise interactions. They also include tasks without explicit
memory requirements.

\begin{figure}[tbp]
    \centering
    \includegraphics[width=0.9\linewidth]{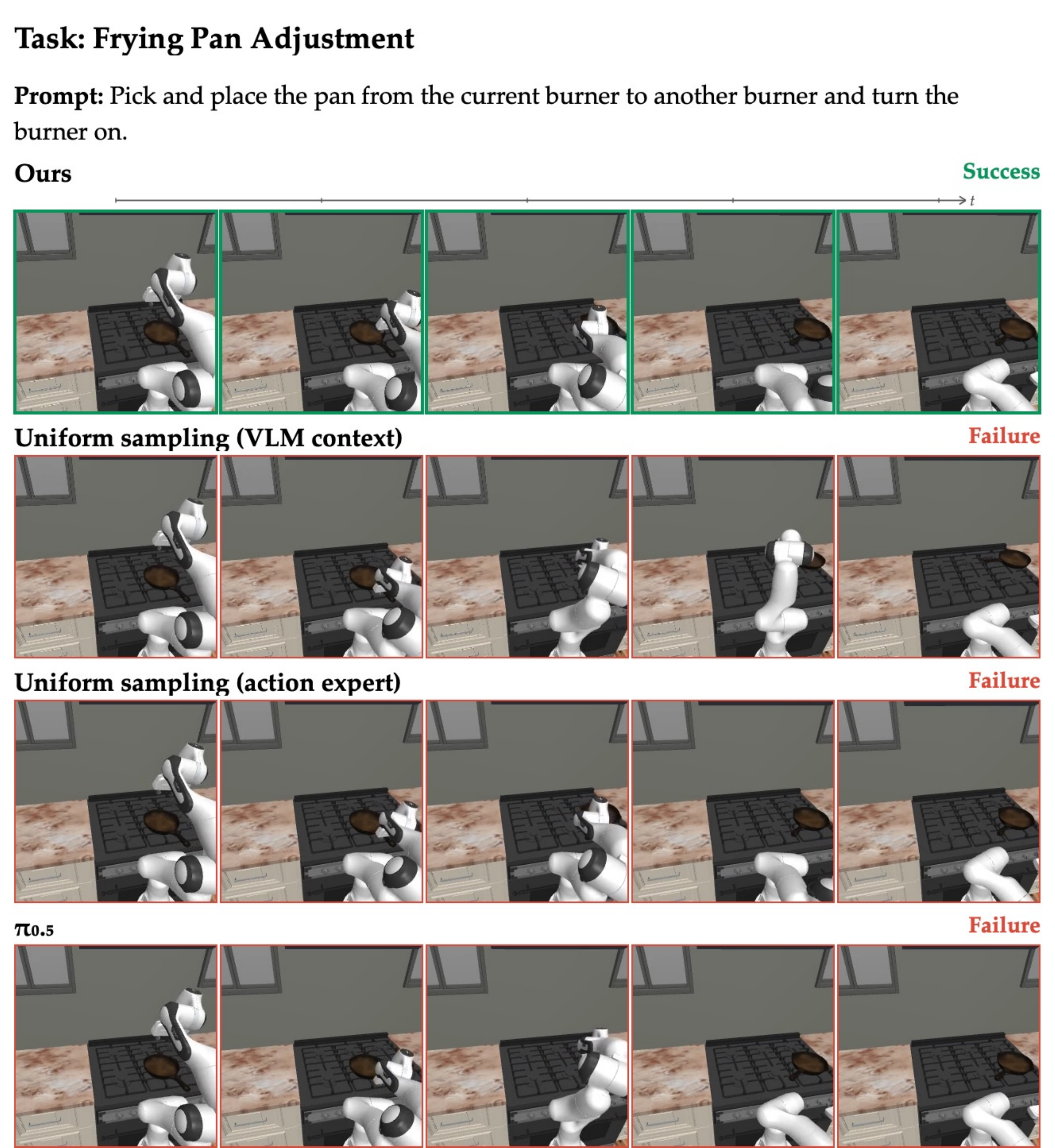}
    \caption{\textbf{RoboCasa: Frying Pan Adjustment.} Representative rollouts of \ours, $\pi_{0.5}$, and Uniform Sampling with VLM-context or action-expert conditioning.}
    \label{fig:robocasa_frying_pan}
\end{figure}

\paragraph{Frying Pan Adjustment.}
As shown in Fig.~\ref{fig:robocasa_frying_pan}, \ours repositions the frying
pan and turns on the stove. All other methods fail to turn on the stove,
leaving the task incomplete.

\begin{figure}[tbp]
    \centering
    \includegraphics[width=0.9\linewidth]{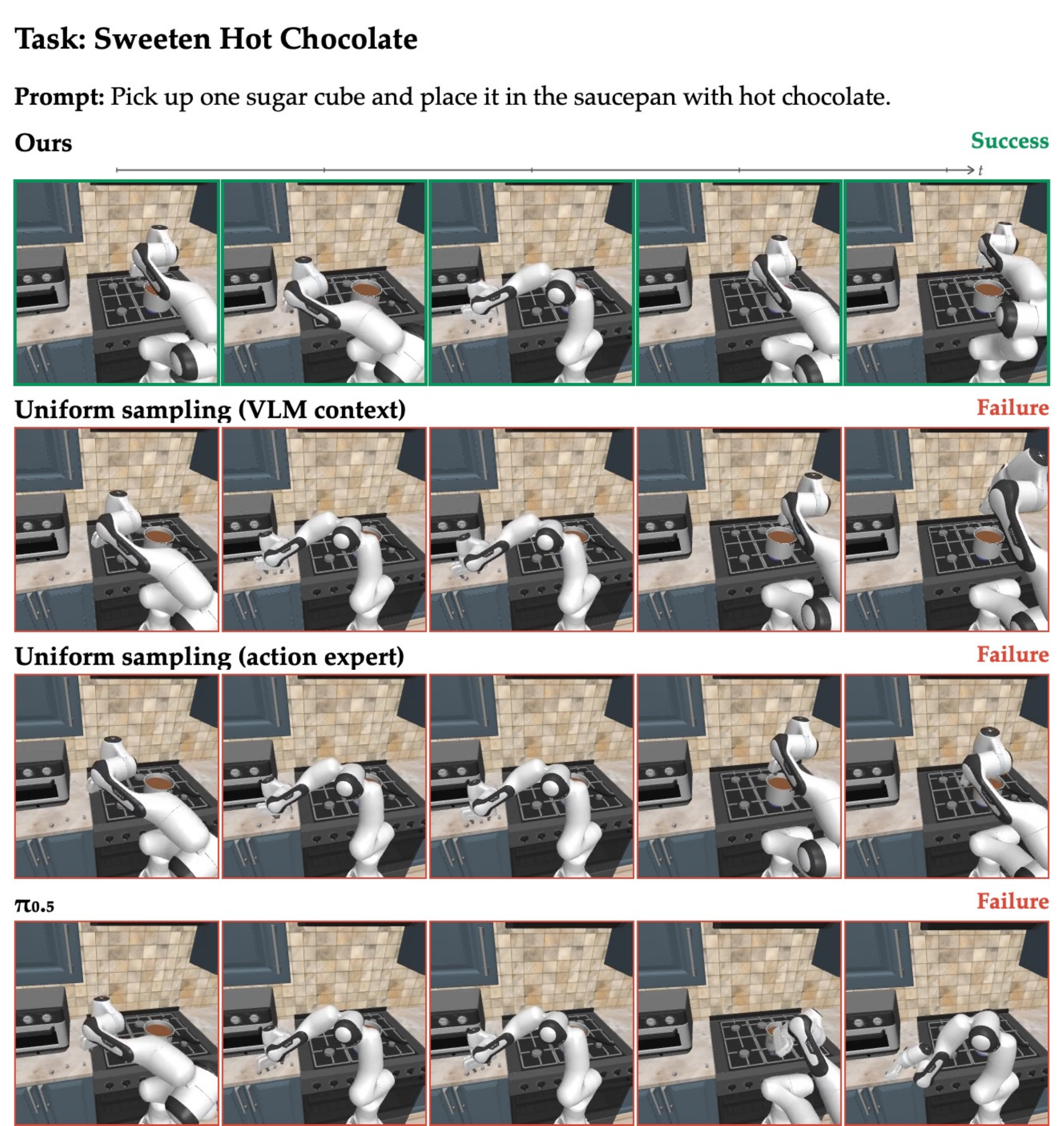}
    \caption{\textbf{RoboCasa: Sweeten Hot Chocolate.} Representative rollouts of \ours, $\pi_{0.5}$, and Uniform Sampling with VLM-context or action-expert conditioning.}
    \label{fig:robocasa_sweeten_hot_chocolate}
\end{figure}

\paragraph{Sweeten Hot Chocolate.}
In Fig.~\ref{fig:robocasa_sweeten_hot_chocolate}, \ours places exactly one
sugar cube in the pot and returns to the reset pose. $\pi_{0.5}$ fails to
place the cube inside the pot, while both Uniform Sampling baselines add
more than one cube. The illustrated failures therefore include both
inaccurate placement and an incorrect number of repetitions.

\begin{figure}[tbp]
    \centering
    \includegraphics[width=\linewidth]{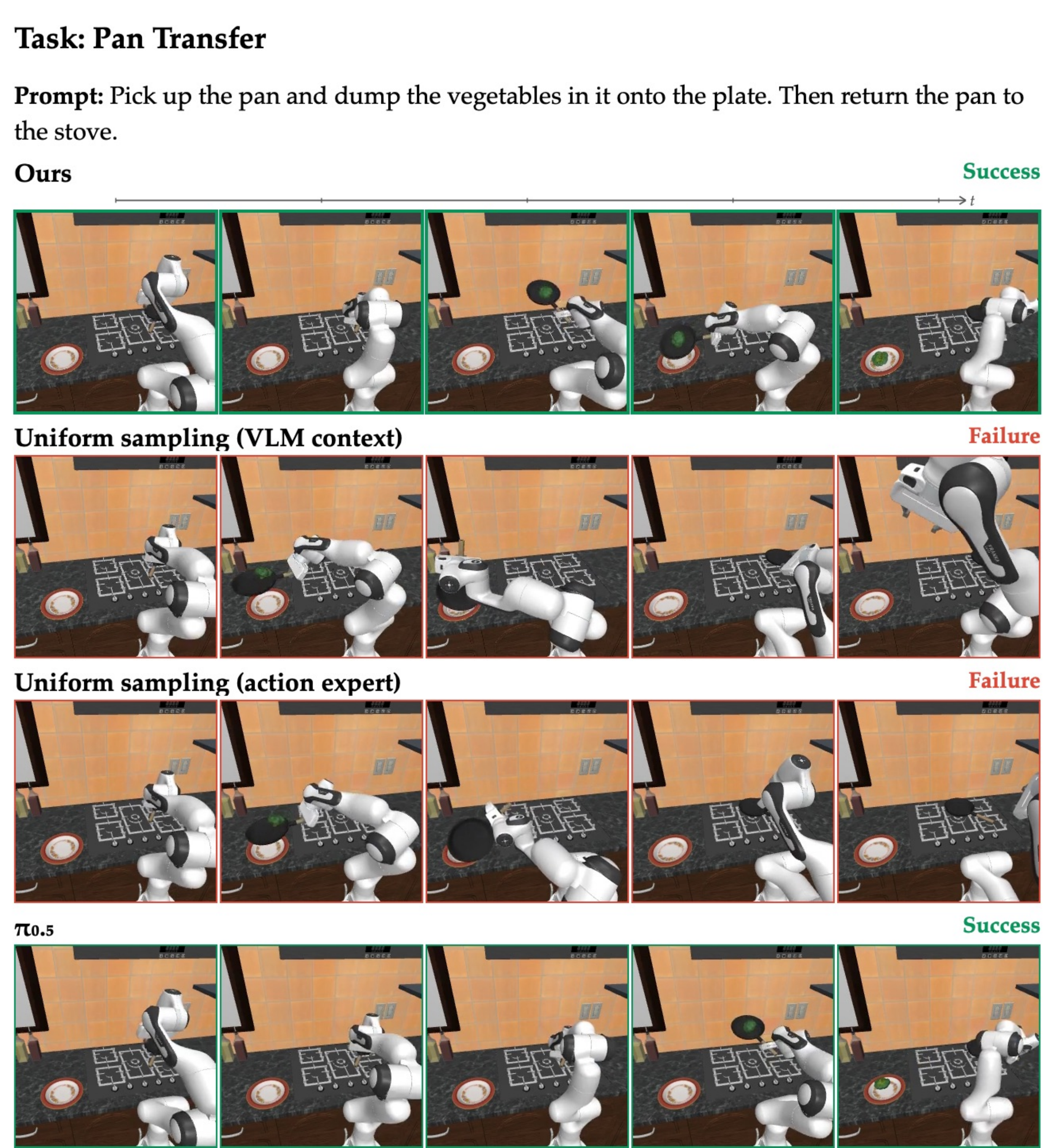}
    \caption{\textbf{RoboCasa: Pan Transfer.} Representative rollouts of \ours, $\pi_{0.5}$, and Uniform Sampling with VLM-context or action-expert conditioning.}
    \label{fig:robocasa_pan_transfer}
\end{figure}

\paragraph{Pan Transfer.}
Both \ours and $\pi_{0.5}$ transfer the food to the plate and return the pan
to its original position on the stove (Fig.~\ref{fig:robocasa_pan_transfer}).
Both Uniform Sampling baselines fail to transfer the food successfully
and to return the pan to the correct stove position.

\begin{figure}[tbp]
    \centering
    \includegraphics[width=\linewidth]{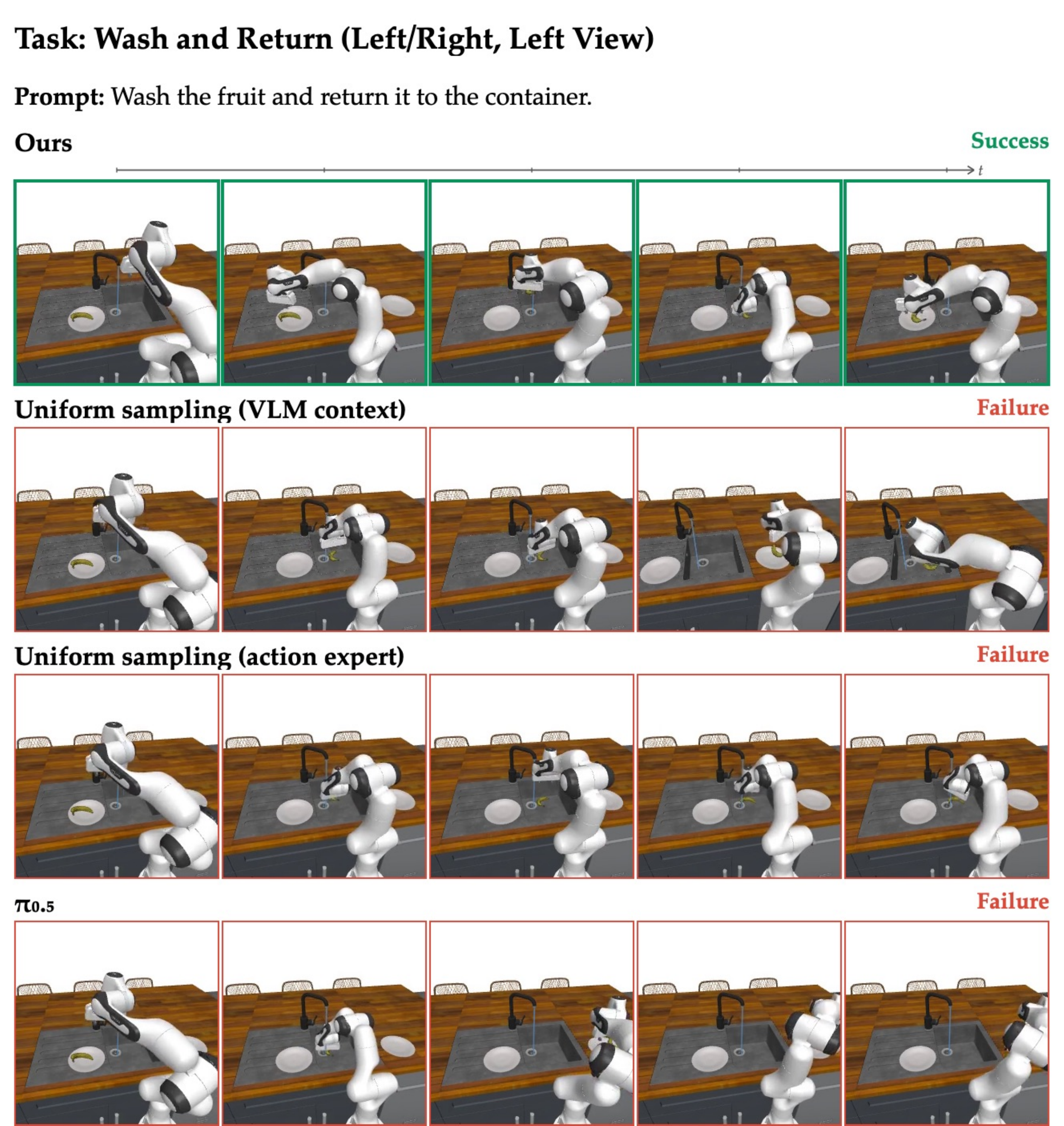}
    \caption{\textbf{RoboCasa: Wash and Return to the Original Plate.} Representative rollouts of \ours, $\pi_{0.5}$, and Uniform Sampling with VLM-context or action-expert conditioning.}
    \label{fig:robocasa_wash_and_return}
\end{figure}

\paragraph{Wash and Return to the Original Plate.}
In Fig.~\ref{fig:robocasa_wash_and_return}, only \ours returns the banana
to its original plate after washing. Both Uniform Sampling baselines fail
to pick the banana back up from the sink, while $\pi_{0.5}$ returns it to
the wrong plate. This example distinguishes a retrieval failure from an
incorrect choice of return location.

\begin{figure}[tbp]
    \centering
    \includegraphics[width=\linewidth]{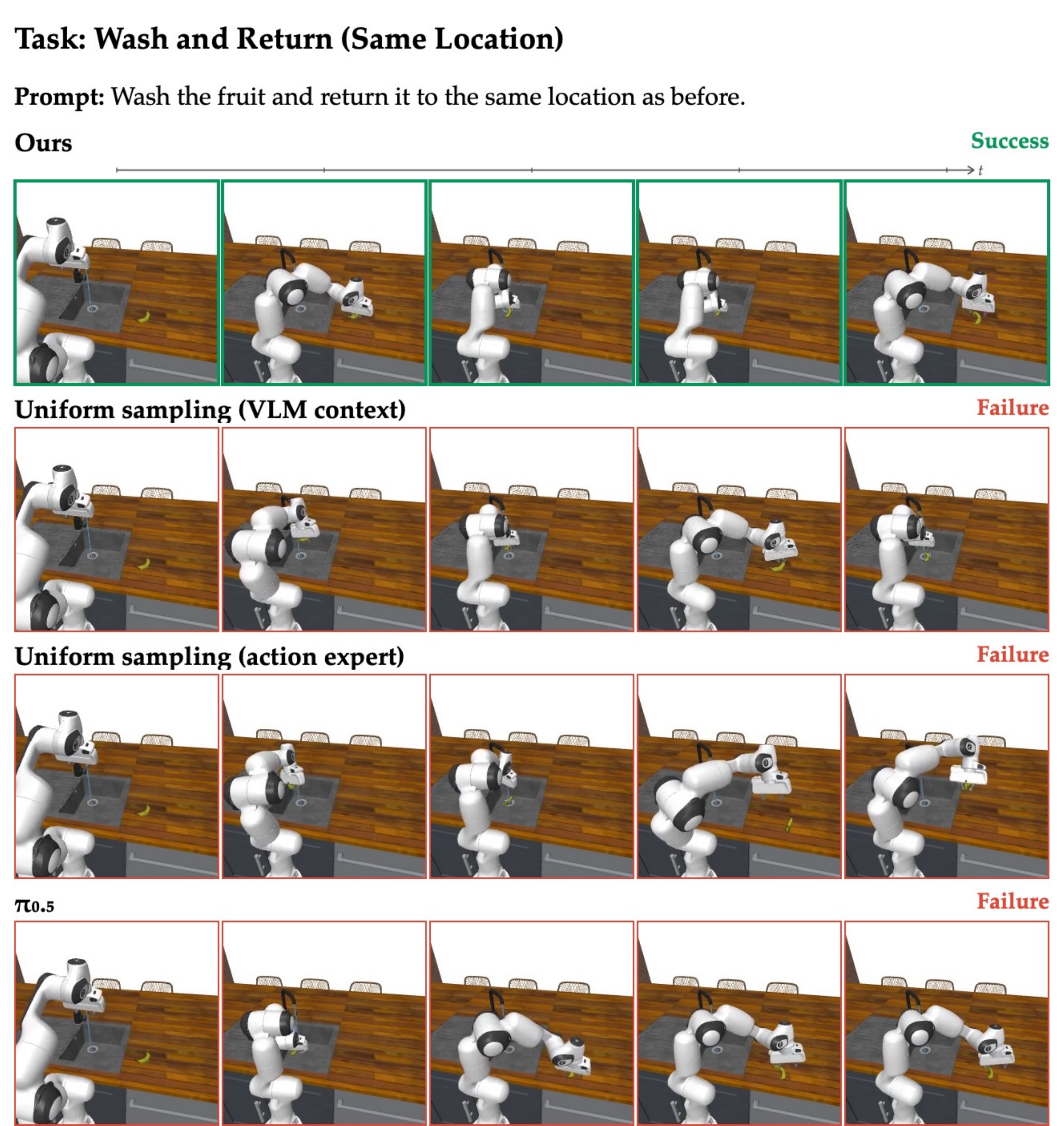}
    \caption{\textbf{RoboCasa: Wash and Return to the Same Location.} Representative rollouts of \ours, $\pi_{0.5}$, and Uniform Sampling with VLM-context or action-expert conditioning.}
    \label{fig:robocasa_wash_and_return_same_location}
\end{figure}

\paragraph{Wash and Return to the Same Location.}
In Fig.~\ref{fig:robocasa_wash_and_return_same_location}, \ours returns the
banana sufficiently close to its original location to satisfy the task's
placement tolerance. The other methods either place it too far from the
original location or repeatedly pick it up without completing the return.

\begin{figure}[tbp]
    \centering
    \includegraphics[width=\linewidth]{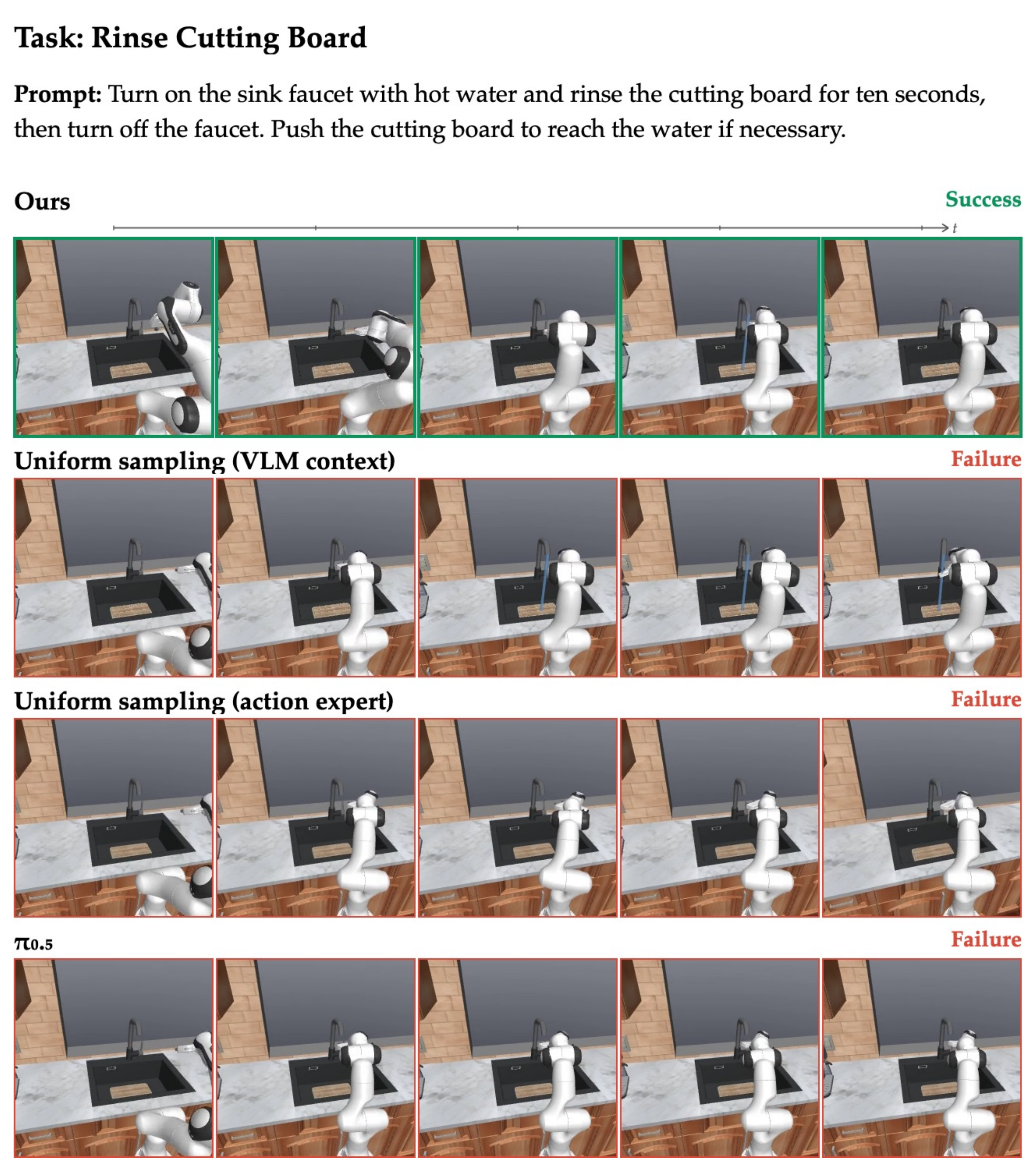}
    \caption{\textbf{RoboCasa: Rinse Cutting Board.} Representative rollouts of \ours, $\pi_{0.5}$, and Uniform Sampling with VLM-context or action-expert conditioning.}
    \label{fig:robocasa_rinse_cutting_board}
\end{figure}

\paragraph{Rinse Cutting Board.}
In Fig.~\ref{fig:robocasa_rinse_cutting_board}, \ours turns on the faucet,
keeps it running for more than 10\,s, and then turns it off. Uniform Sampling
with VLM-context conditioning turns on the faucet but fails to turn it off.
Uniform Sampling with action-expert conditioning and $\pi_{0.5}$ fail to
turn it on. Thus, the illustrated failures occur at different stages of
the required interaction sequence.

\begin{figure}[tbp]
    \centering
    \includegraphics[width=\linewidth]{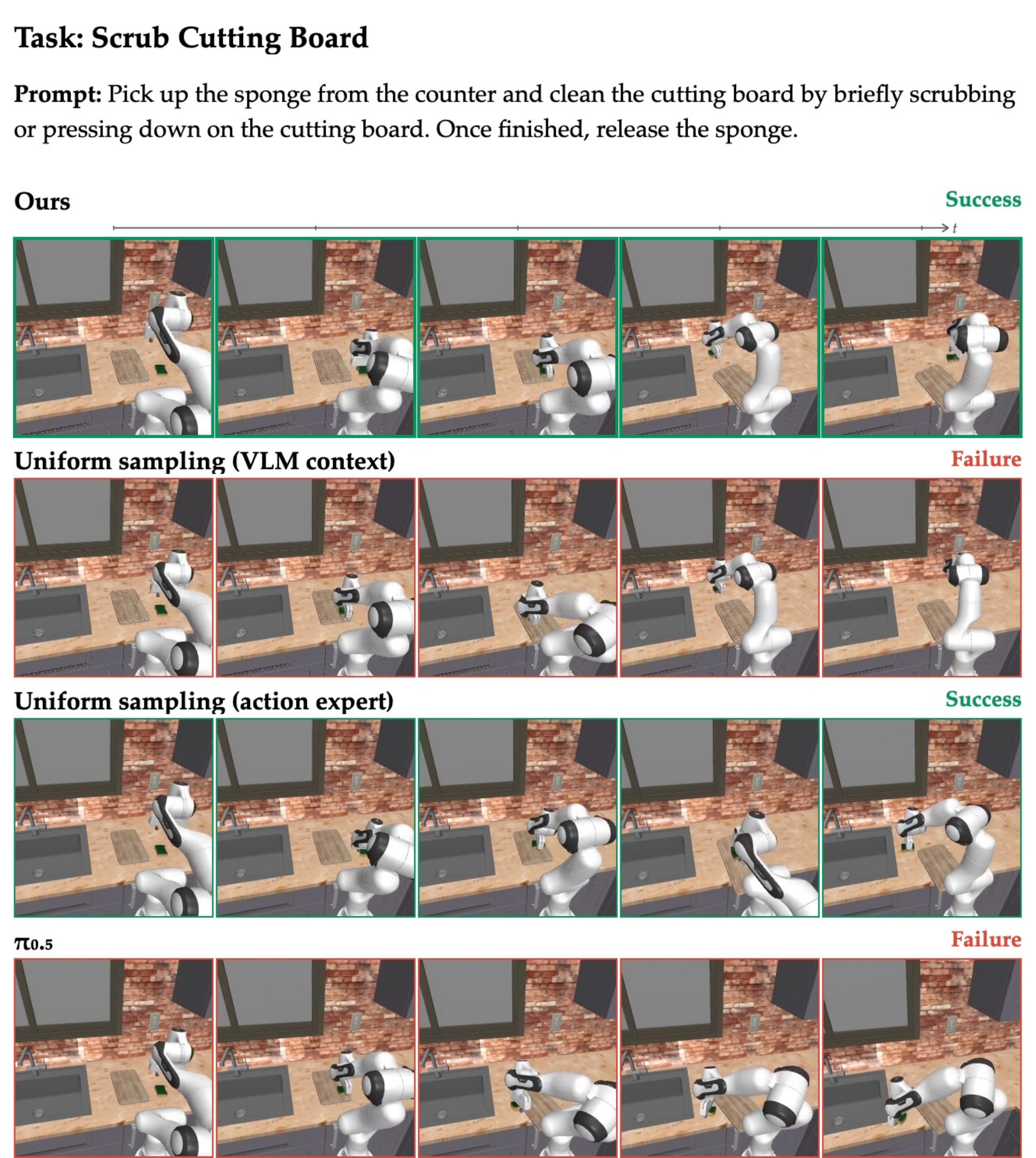}
    \caption{\textbf{RoboCasa: Scrub Cutting Board.} Representative rollouts of \ours, $\pi_{0.5}$, and Uniform Sampling with VLM-context or action-expert conditioning.}
    \label{fig:robocasa_Scrub_cutting_board}
\end{figure}

\paragraph{Scrub Cutting Board.}
As shown in Fig.~\ref{fig:robocasa_Scrub_cutting_board}, \ours and Uniform
Sampling with action-expert conditioning scrub a sufficiently large area
of the cutting board. Uniform Sampling with VLM-context conditioning and
$\pi_{0.5}$ fail to achieve the required coverage.

\begin{figure}[tbp]
    \centering
    \includegraphics[width=\linewidth]{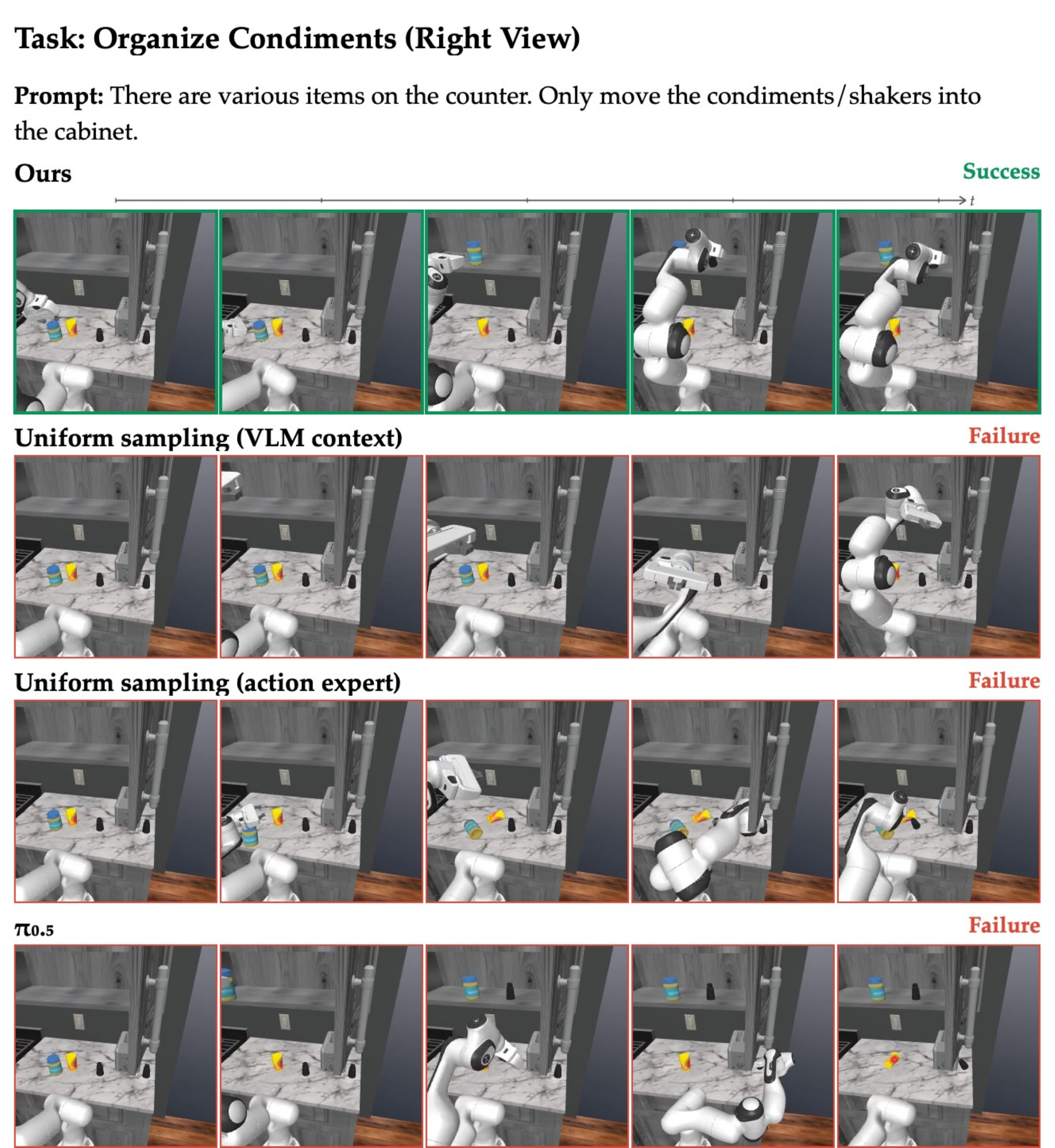}
    \caption{\textbf{RoboCasa: Organize Condiments.} Representative rollouts of \ours, $\pi_{0.5}$, and Uniform Sampling with VLM-context or action-expert conditioning.}
    \label{fig:robocasa_organize_condiments}
\end{figure}

\paragraph{Organize Condiments.}
In Fig.~\ref{fig:robocasa_organize_condiments}, \ours places all three
condiment items in the cabinet. Neither Uniform Sampling baseline places
an item inside the cabinet. $\pi_{0.5}$ places two items inside but leaves
the black item behind, failing to complete the full set of transfers.

\begin{figure}[tbp]
    \centering
    \includegraphics[width=\linewidth]{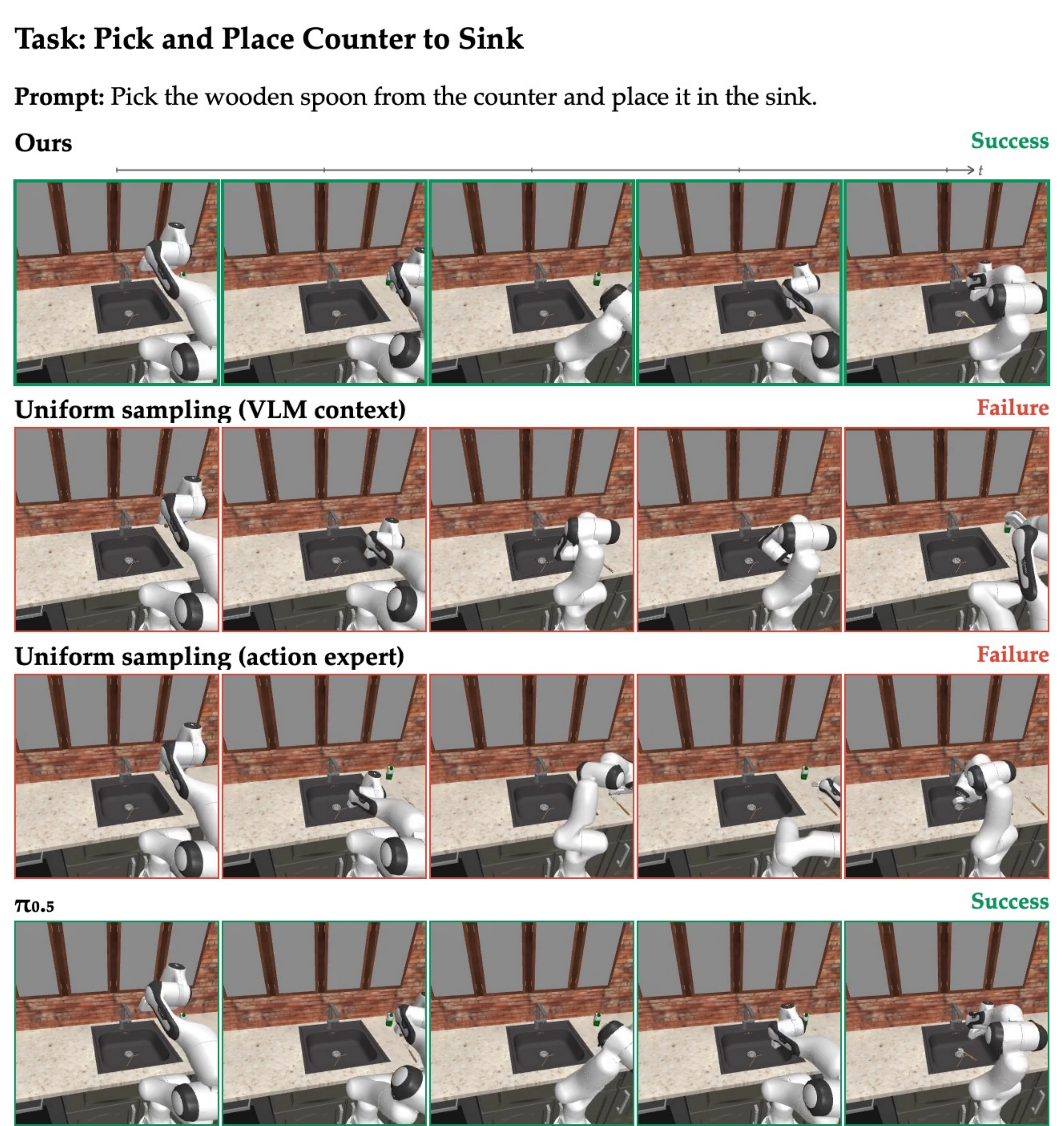}
    \caption{\textbf{RoboCasa: Pick and Place to Sink.} Representative rollouts of \ours, $\pi_{0.5}$, and Uniform Sampling with VLM-context or action-expert conditioning.}
    \label{fig:robocasa_pick_and_place_sink}
\end{figure}

\paragraph{Pick and Place to Sink.}
Both \ours and $\pi_{0.5}$ successfully pick up the spoon and place it in
the sink, while both Uniform Sampling baselines fail
(Fig.~\ref{fig:robocasa_pick_and_place_sink}). This example shows that the
qualitative differences extend to a task without an explicit memory
requirement.

\begin{figure}[tbp]
    \centering
    \includegraphics[width=\linewidth]{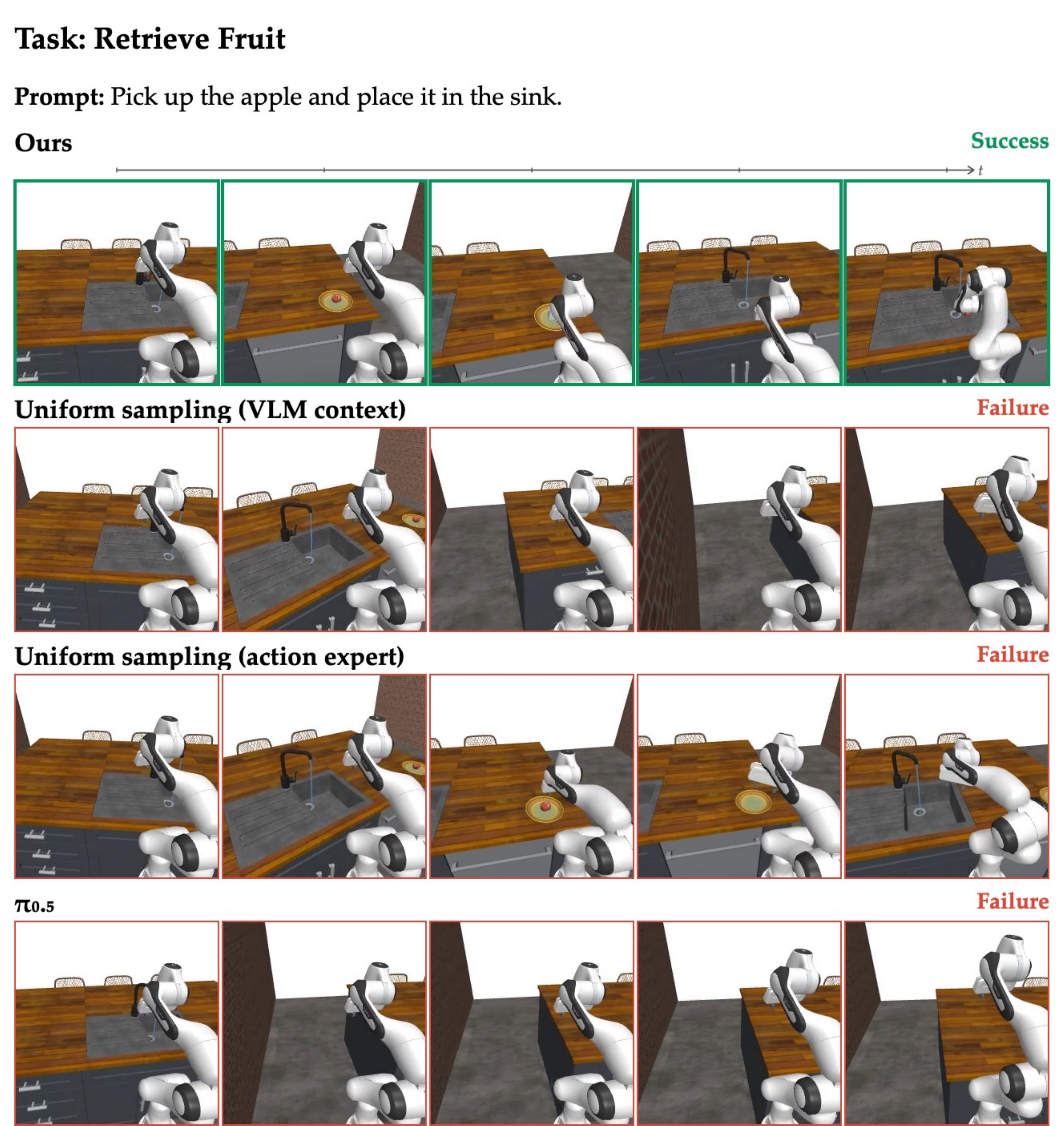}
    \caption{\textbf{RoboCasa: Retrieve Fruit.} Representative rollouts of \ours, $\pi_{0.5}$, and Uniform Sampling with VLM-context or action-expert conditioning.}
    \label{fig:robocasa_retrieve_fruit}
\end{figure}

\paragraph{Retrieve Fruit.}
In Fig.~\ref{fig:robocasa_retrieve_fruit}, only \ours retrieves the fruit
and places it in the sink. Uniform Sampling with VLM-context conditioning
and $\pi_{0.5}$ approach the wrong side of the workspace. Uniform Sampling
with action-expert conditioning approaches the correct side but drops the
fruit following an imprecise grasp. This comparison illustrates distinct
errors in selecting the retrieval location and executing the grasp.

\begin{figure}[tbp]
    \centering
    \includegraphics[width=\linewidth]{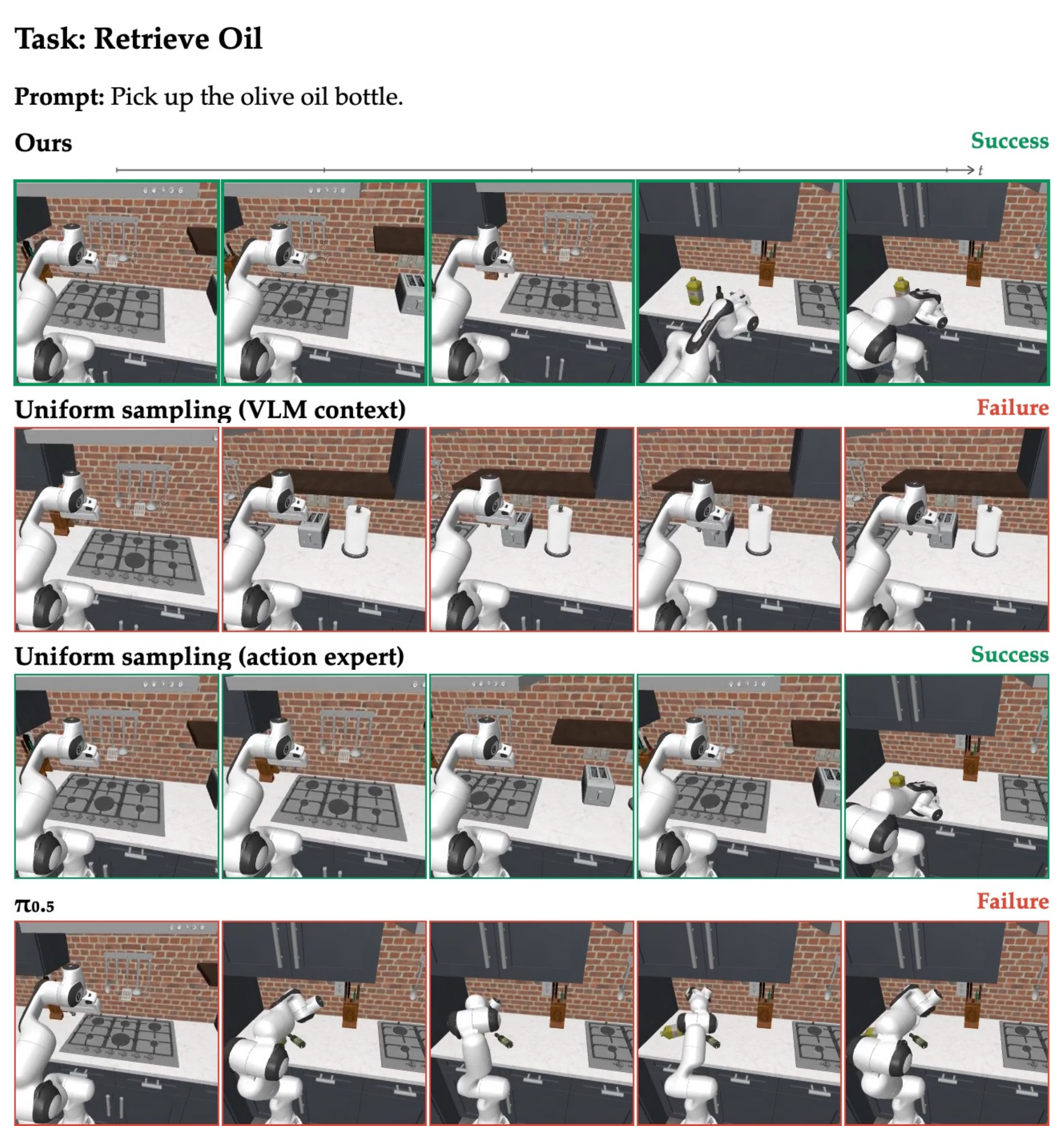}
    \caption{\textbf{RoboCasa: Retrieve Oil.} Representative rollouts of \ours, $\pi_{0.5}$, and Uniform Sampling with VLM-context or action-expert conditioning.}
    \label{fig:robocasa_retrieve_oil}
\end{figure}

\paragraph{Retrieve Oil.}
As shown in Fig.~\ref{fig:robocasa_retrieve_oil}, both \ours and Uniform
Sampling with action-expert conditioning grasp the correct olive oil
bottle. Uniform Sampling with VLM-context conditioning approaches the
wrong side of the counter, while $\pi_{0.5}$ fails to pick up the bottle.

\begin{figure}[tbp]
    \centering
    \includegraphics[width=\linewidth]{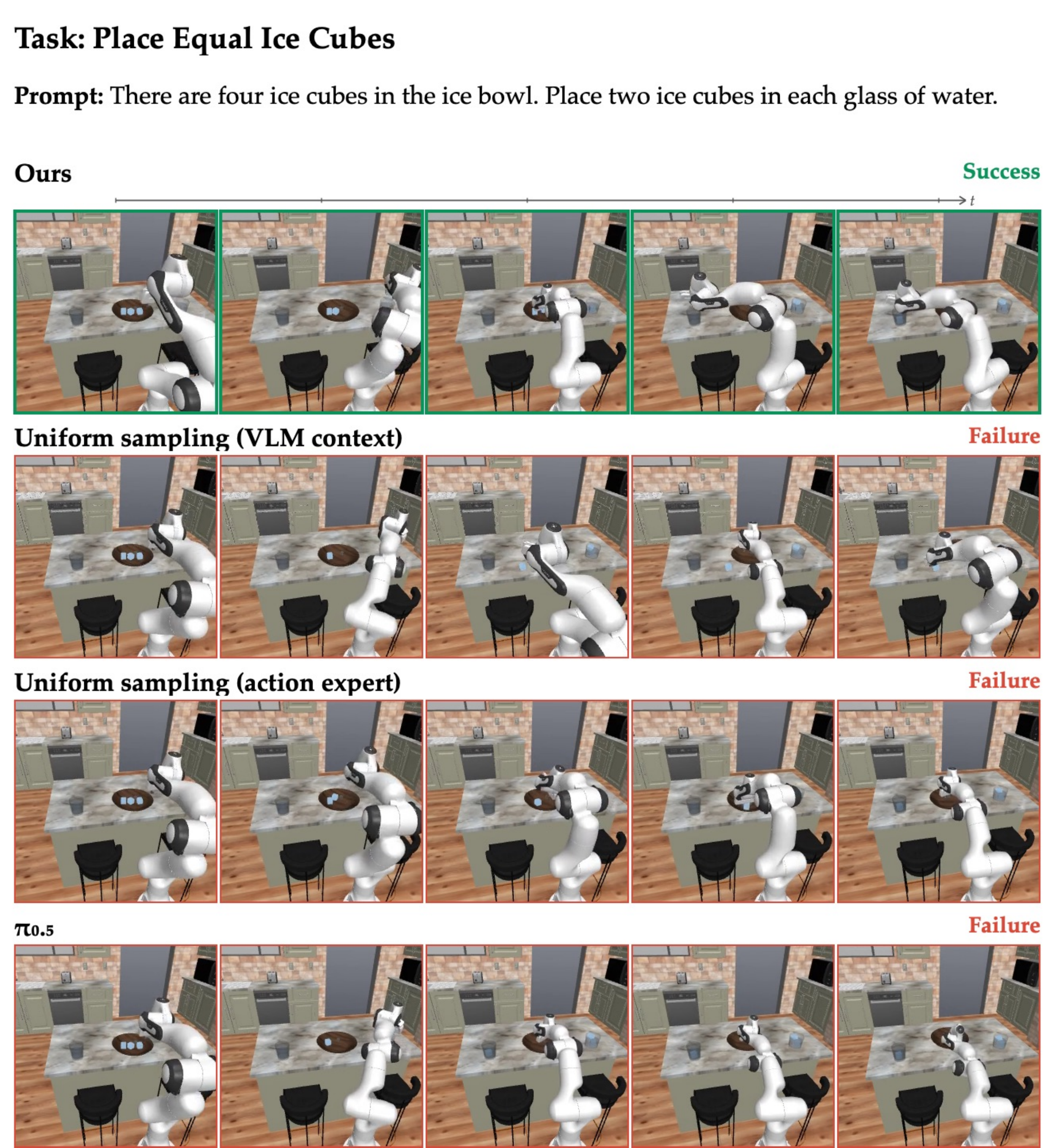}
    \caption{\textbf{RoboCasa: Place Equal Ice Cubes.} Representative rollouts of \ours, $\pi_{0.5}$, and Uniform Sampling with VLM-context or action-expert conditioning.}
    \label{fig:robocasa_place_equal_cubes}
\end{figure}

\paragraph{Place Equal Ice Cubes.}
In Fig.~\ref{fig:robocasa_place_equal_cubes}, only \ours places two ice
cubes in each cup. Each of the other three methods fails to place the
final required cube, leaving the task incomplete.

\subsection{Real-World Analysis}
\label{app:qualitative_real_world}

We compare \textsc{Imprint} with $\pi_0$ (non-memory), HAMLET, MemER, and Uniform Sampling on the three real-world tasks, and also show \textsc{Imprint} on out-of-distribution (OOD) objects (App.~\ref{app:real_world_eval}). As in simulation, the rollouts illustrate both memory errors, such as losing track of object arrangements, scanning progress, or the demonstrated folding style, and manipulation errors in otherwise correct plans. Aggregate success rates are reported in Table~\ref{tab:realworld_detailed}.

\begin{figure}[t]
    \centering
    \includegraphics[width=\linewidth]{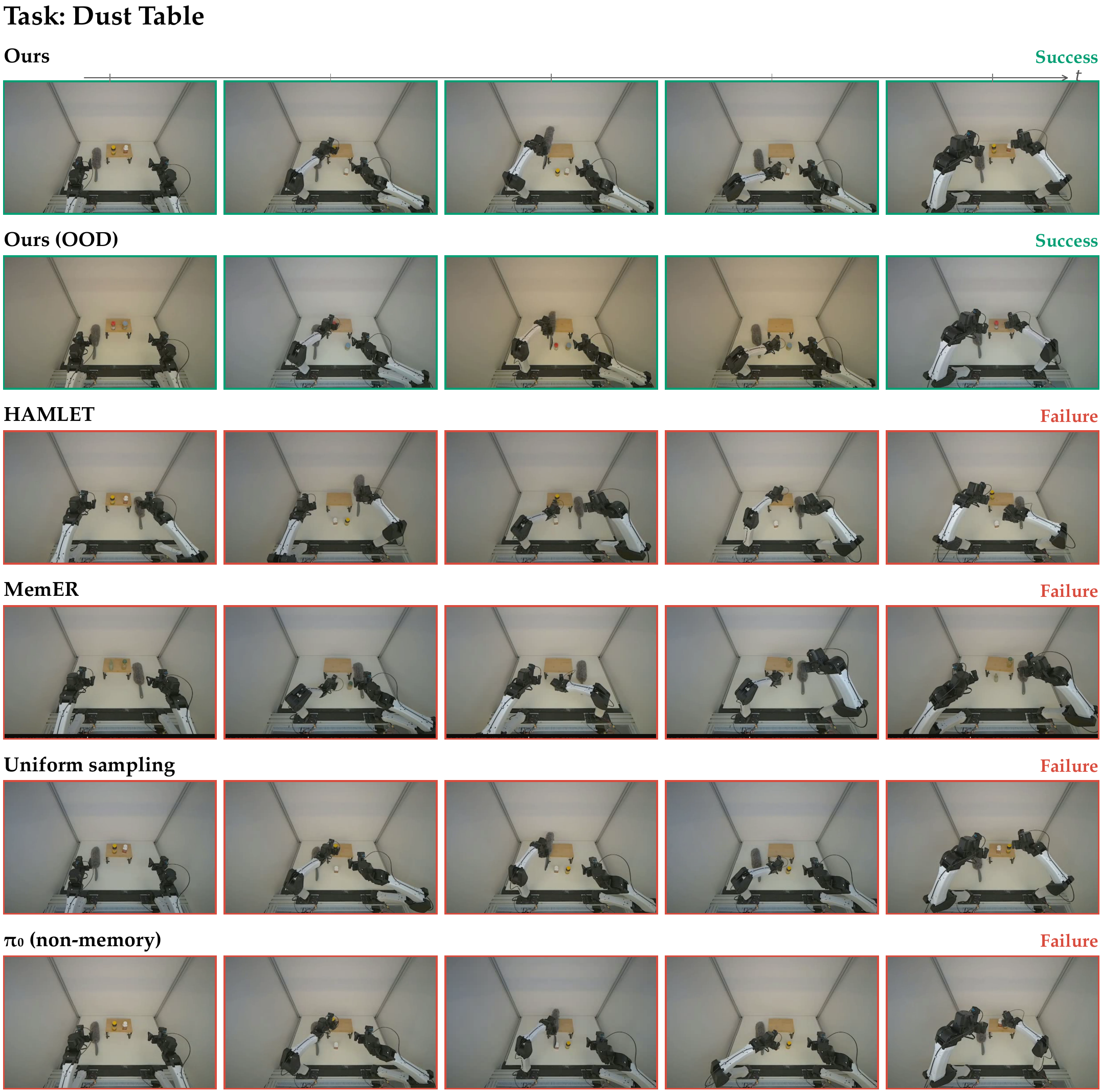}
    \caption{\textbf{Real World: Dust Table.} Representative rollouts of \textsc{Imprint} (in-distribution and OOD objects), HAMLET, MemER, Uniform Sampling, and $\pi_0$.}
    \label{fig:real_dust_table}
\end{figure}

\paragraph{Dust Table.}
As shown in Fig.~\ref{fig:real_dust_table}, \textsc{Imprint} moves both bottles off the small table, dusts it once, and returns the bottles upright in their original left/right arrangement. The same behavior holds with unseen OOD objects. HAMLET dusts the table but leaves one bottle on the workspace instead of returning it. %
Uniform Sampling and $\pi_0$ both return the bottles to the small table but fail the task: Uniform Sampling places them in the swapped arrangement, and $\pi_0$, which cannot recall the initial configuration, likewise does not restore it. %
MemER fails to pick up the last bottle showing that its high-level VLM can switch subtasks too quickly; across all trials, it succeeds in only 40\% of rollouts compared with 80\% for \textsc{Imprint} (Table~\ref{tab:realworld_detailed}).

\begin{figure}[t]
    \centering
    \includegraphics[width=\linewidth]{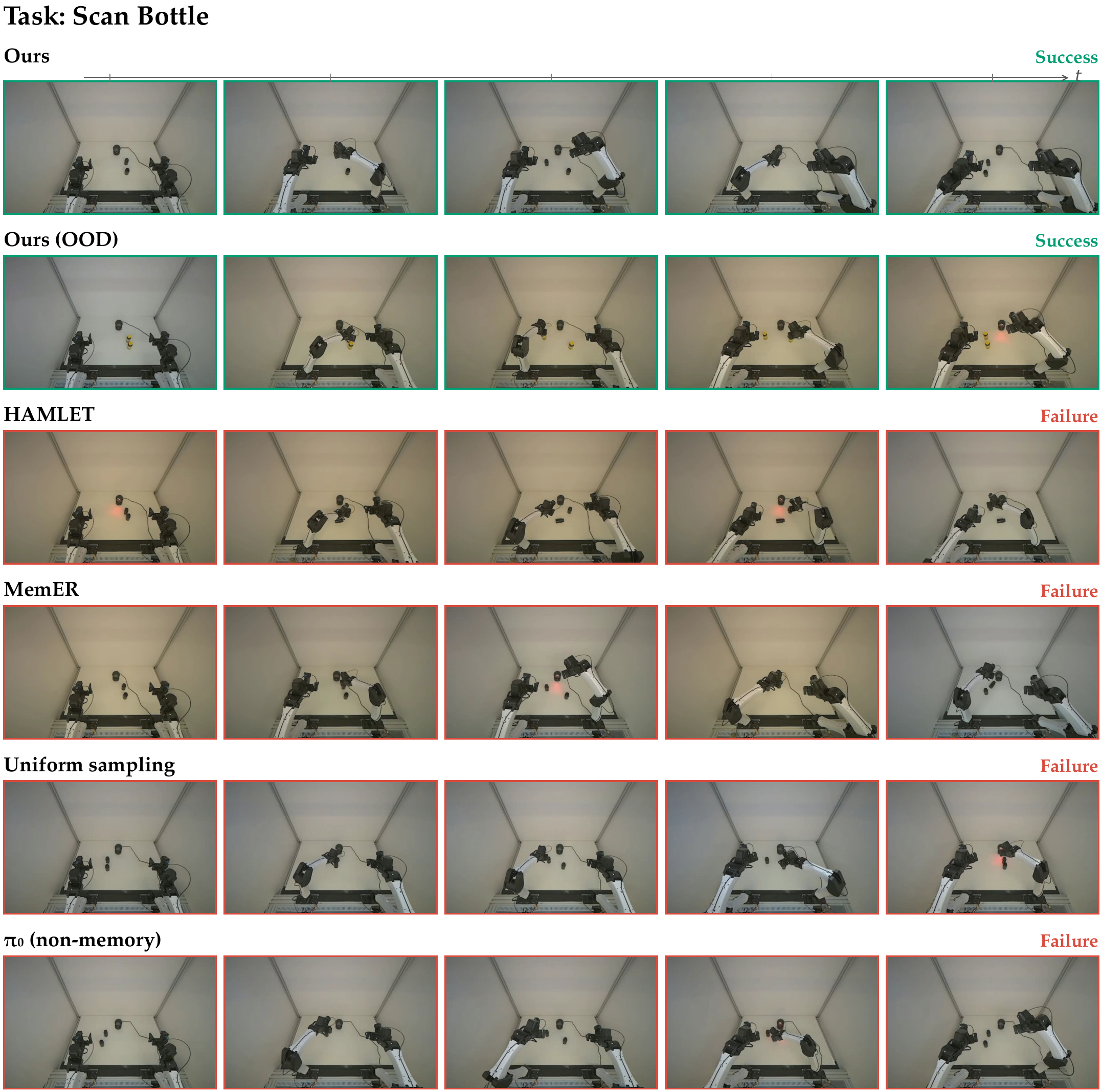}
    \caption{\textbf{Real World: Scan Bottle.} Representative rollouts of \textsc{Imprint} (in-distribution and OOD objects), HAMLET, MemER, Uniform Sampling, and $\pi_0$.}
    \label{fig:real_scan_bottle}
\end{figure}

\paragraph{Scan Bottle.}
In Fig.~\ref{fig:real_scan_bottle}, \textsc{Imprint} picks up the bottles one at a time, scans each exactly once, and places them in a row on the other side of the table. It generalizes to the OOD objects with the same sequence. HAMLET triggers the scanner more than once and does not complete the transfer, consistent with losing track of which bottles have already been scanned. %
MemER scans a bottle but then returns to it instead of proceeding to the next one, matching the subtask-prediction errors noted in App.~\ref{app:realworld_detailed}. %
Uniform Sampling and $\pi_0$ both hesitate between picking and scanning and do not complete the scan-and-transfer sequence before the rollout ends. %
Because scanning does not visibly change the bottles, the scanning status must be inferred from the robot's past actions and proprioceptive states, which \textsc{Imprint} encodes and the visual-only baselines do not.

\begin{figure}[t]
    \centering
    \includegraphics[width=\linewidth]{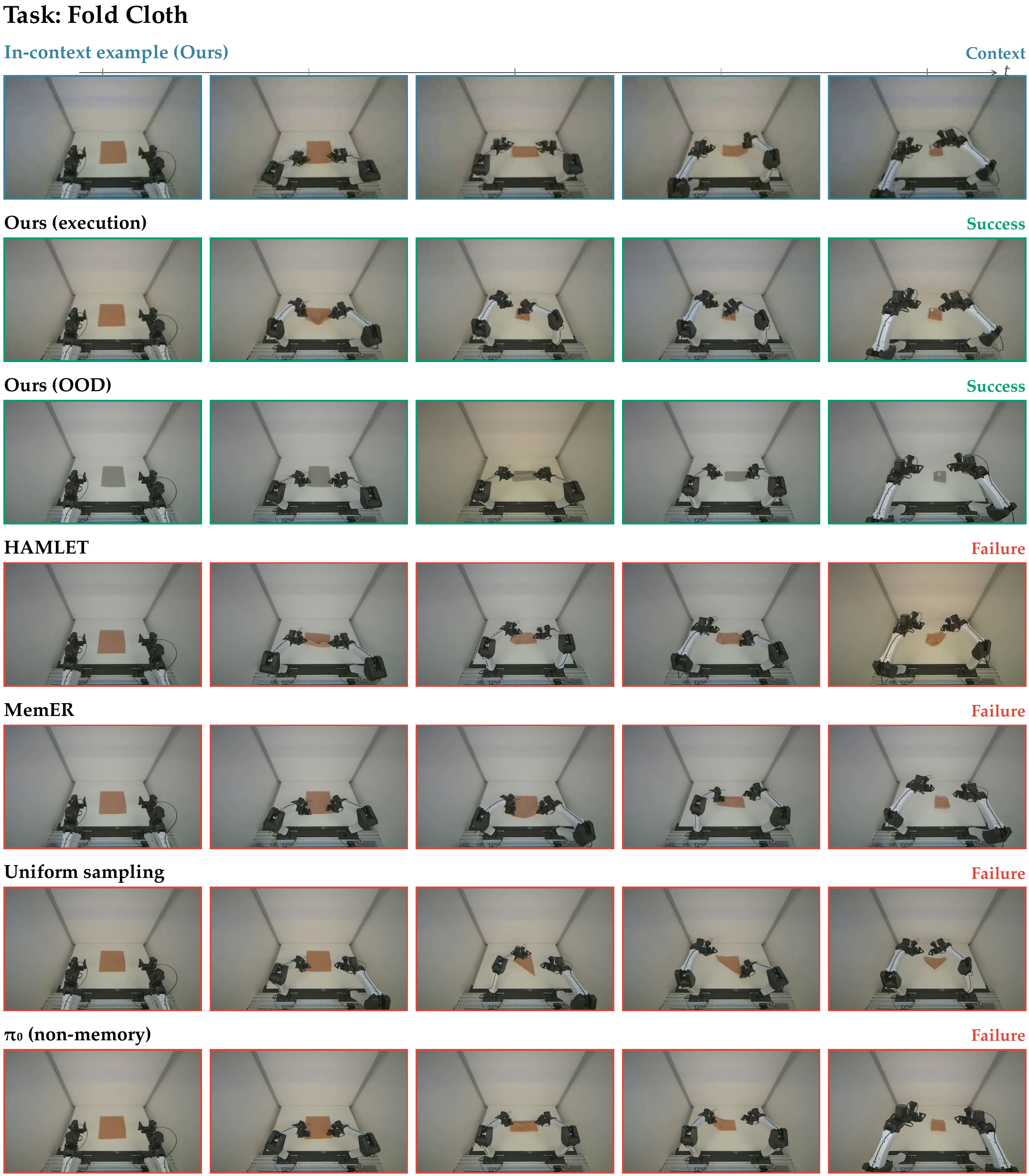}
    \caption{\textbf{Real World: Fold Cloth.} The top row shows the in-context demonstration provided to \textsc{Imprint}'s memory, followed by representative rollouts of \textsc{Imprint} (in-distribution and OOD towel), HAMLET, MemER, Uniform Sampling, and $\pi_0$.}
    \label{fig:real_fold_cloth}
\end{figure}

\paragraph{Fold Cloth.}
Fig.~\ref{fig:real_fold_cloth} first shows the in-context demonstration, which folds the towel in half and then folds it a second time. \textsc{Imprint} reproduces both folds and reaches a final shape close to the demonstrated one, including on an unseen towel color. The baselines instead fold in a style that differs from the example. Uniform Sampling folds the towel diagonally into a triangle, suggesting that it recalls that a fold is required but not which fold. %
HAMLET's and $\pi_0$'s folds do not match the demonstrated sequence, and MemER completes a fold whose final shape deviates from the example. %
Identifying the style from the example's appearance is not sufficient; reproducing it also requires the demonstrated grasp locations and folding motion, which \textsc{Imprint} retains through its action and proprioceptive streams.

\end{document}